\documentclass{article}

\PassOptionsToPackage{numbers, compress}{natbib}
\usepackage[preprint]{neurips_2026}

\usepackage[utf8]{inputenc} 
\usepackage[T1]{fontenc}    
\usepackage{hyperref}       
\usepackage{url}            
\usepackage{booktabs}       
\usepackage{amsfonts}       
\usepackage{amsmath}        
\usepackage{nicefrac}       
\usepackage{microtype}      
\usepackage{xcolor}         
\usepackage{graphicx}       
\usepackage{float}          
\usepackage{multirow}       
\usepackage{subcaption}     
\usepackage{wrapfig}        

\usepackage{xspace}
\makeatletter
\DeclareRobustCommand\onedot{\futurelet\@let@token\@onedot}
\def\@onedot{\ifx\@let@token.\else.\null\fi\xspace}
\def\vs{\emph{vs}\onedot}
\makeatother

\newcommand{\tablewidth}{0.95\linewidth}

\title{Pretrained Video Models as Differentiable Physics Simulators for Urban Wind Flows}

\author{%
Janne Perini$^{1*}$ \hspace{4em} Rafael Bischof$^{1*}$ \hspace{4em} Moab Arar$^{2}$ \hspace{4em} Ay\c{c}a Duran$^{3}$ \\ \\
\textbf{Michael A.\ Kraus}$^{4}$ \hspace{4em} \textbf{Siddhartha Mishra}$^{5}$ \hspace{4em} \textbf{Bernd Bickel}$^{1}$ \\
\\
$^{1}$Computational Design Lab, ETH Zurich, Switzerland\\
$^{2}$Tel Aviv University, Israel\\
$^{3}$Architecture and Building Systems, ETH Zurich, Switzerland\\
$^{4}$Institute of Structural Mechanics and Design, TU Darmstadt, Germany\\
$^{5}$Seminar for Applied Mathematics, ETH Zurich, Switzerland\\
$^{*}$Equal contribution. Correspondence to \href{mailto:rabischof@ethz.ch}{rabischof@ethz.ch}
}

\begin{document}

\maketitle

\begin{abstract}
  Designing urban spaces that provide pedestrian wind comfort and safety requires time-resolved Computational Fluid Dynamics (CFD) simulations, but their current computational cost makes extensive design exploration impractical. We introduce WinDiNet (Wind Diffusion Network), a pretrained video diffusion model that is repurposed as a fast, differentiable surrogate for this task. Starting from LTX-Video, a 2B-parameter latent video transformer, we fine-tune on a dataset of 13{,}000 2D incompressible CFD simulations over procedurally generated building layouts. A systematic study of training regimes, conditioning mechanisms, and VAE adaptation strategies, including a physics-informed decoder loss, identifies a configuration that outperforms purpose-built neural PDE solvers. The resulting model generates full 112-frame rollouts in under a second. As the surrogate is end-to-end differentiable, it doubles as a physics simulator for gradient-based inverse optimization: given an urban footprint layout, we optimize building positions directly through backpropagation to improve wind safety as well as pedestrian wind comfort. Experiments on single- and multi-inlet layouts show that the optimizer discovers effective layouts even under challenging multi-objective configurations, with all improvements confirmed by ground-truth CFD simulations.
\end{abstract}

\section{Introduction}
\label{sec:intro}
When urban designers and engineers plan modern, livable public spaces, wind is a key constraint. It affects pedestrian comfort, structural loads on buildings, and natural ventilation. Buildings that channel airflow can produce gusts that endanger pedestrians and stress building facades, while overly sheltered configurations create stagnant zones where heat and pollutants accumulate. In cities increasingly affected by air pollution and rising temperatures, these aspects carry real consequences for structural safety, public health, and quality of life. Exploring urban layouts that navigate this trade-off requires fast, accurate flow predictions and, ideally, guidance on how to update a given design to improve wind comfort and safety simultaneously. Computational fluid dynamics (CFD) simulations, the traditional method for this task, are computationally expensive and require tedious modeling and domain expertise in addition to taking minutes to hours at the resolution needed for a reliable comfort evaluation~\cite{MIRZAEI2021102839}. Furthermore, they are typically non-differentiable, meaning they offer no signal about which geometric changes would improve a given design.

Deep-learning surrogates mitigate these limitations by offering fast inference and differentiable predictions. While existing approaches based on convolutional and graph neural networks can predict mean velocity or wind-factor fields~\cite{mokhtar2021pedestrian,liu2023sgmsgnn}, temporally averaged outputs cannot evaluate comfort criteria defined as exceedance probabilities over threshold speeds~\cite{lawson1978wind}, nor capture transient gusts relevant to pedestrian safety. Temporal extensions based on Fourier neural operators and autoregressive transformers~\cite{qin2024localfno,chen2025fnourban} can in principle produce time-resolved predictions, but they struggle to maintain accuracy over long rollouts when trained from scratch on limited domain-specific data.

Meanwhile, video diffusion models have advanced rapidly in capabilities relevant to wind flow prediction. Recent latent diffusion transformers generate high-resolution, temporally coherent sequences with multi-scale motion and long-range spatial dependencies~\cite{hacohen2024ltx,Wan22,blattmann2023stable}. These models already encode physical priors, such as gravity, collisions, and fluid-like motion, that can be adapted to generate visually plausible physical phenomena~\cite{wiedemer2025videoreasoning,gillman2025force,zhang2025think,wang2025physctrl}. Predicting urban wind flows fits a similar framing: projected onto a horizontal plane, a velocity field evolves as a frame sequence whose channels encode physical quantities rather than pixel colors. We show that fine-tuning a pretrained video model on synthetic urban CFD data yields a differentiable surrogate accurate enough for both forward prediction and gradient-based inverse optimization of building layouts for pedestrian wind comfort, and that the physical priors transfer to quantitatively accurate simulations, not just visually plausible ones.

\begin{figure}[tbp]
    \centering
    \includegraphics[width=\linewidth]{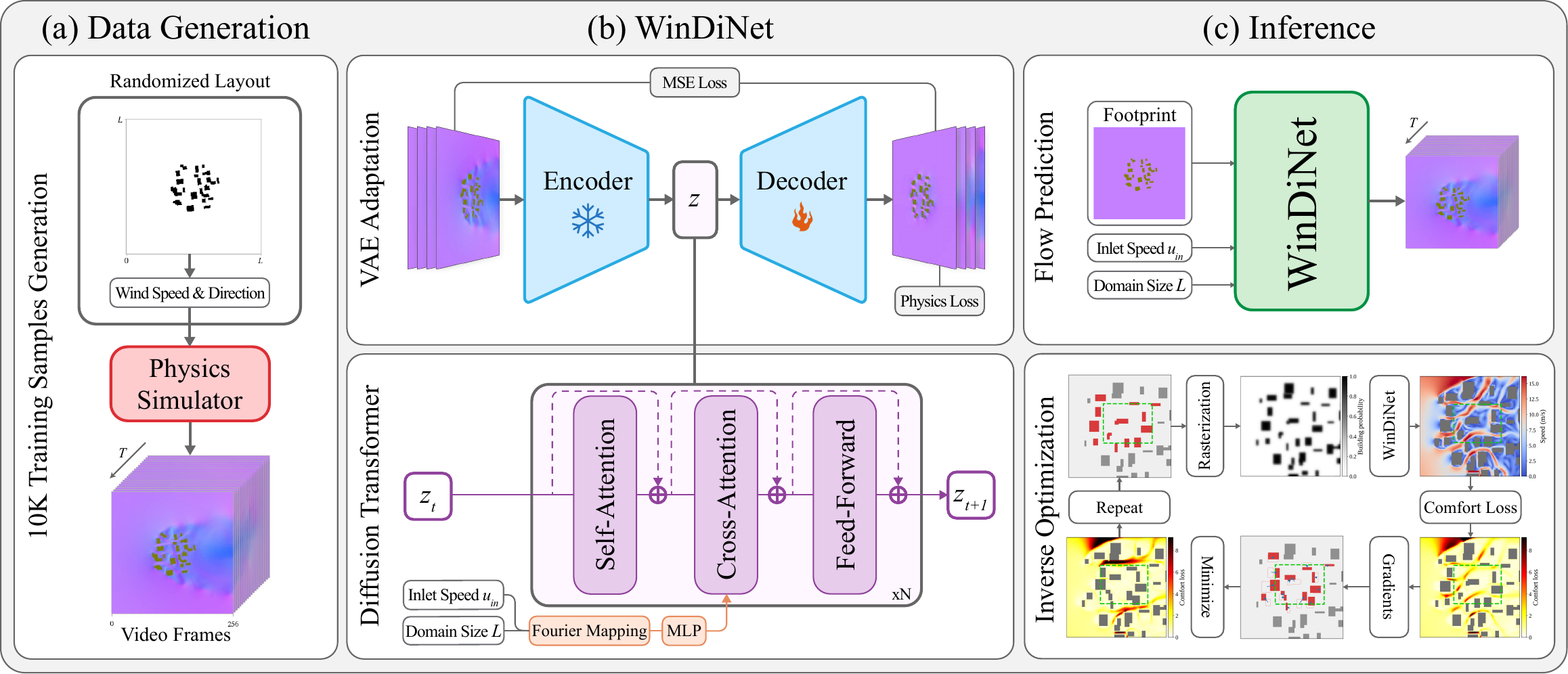}
    \caption{Overview of the proposed framework. \textbf{(a)}~Procedurally generated urban layouts are simulated with a 2D incompressible Euler solver to produce training data. \textbf{(b)}~A latent diffusion model with a physics-informed VAE is trained to generate wind field sequences conditioned on building footprint, inlet speed $u_\mathrm{in}$, and domain size $L$. \textbf{(c)}~At inference, the model generates horizontal and vertical velocity fields $(u, v)$ and enables gradient-based inverse optimization of building layouts.}
    \label{fig:pipeline}
\end{figure}

In this work, we fine-tune LTX-Video~\cite{hacohen2024ltx}, a transformer-based latent video diffusion model, on a dataset of 13{,}000 2D CFD simulations over procedurally generated urban footprint layouts and systematically study the design choices needed to obtain physically accurate predictions from a pretrained video backbone (Figure~\ref{fig:pipeline}). We compare (i) the training regime, such as Low-Rank Adaptation (LoRA)~\cite{hu2022lora} \vs full fine-tuning, (ii) the conditioning mechanism (text prompts \vs scalar-conditioned latent modulation with the text encoder removed), and (iii) the Variational Autoencoder (VAE) adaptation strategy, including color adapters, decoder fine-tuning, and physics-informed objectives. Our main contributions are:

\begin{itemize}
  \item We generate a dataset of 13{,}000 2D incompressible wind flow simulations over procedurally generated urban layouts, spanning diverse building configurations, inlet speeds, and domain sizes. 
  \item We present a systematic study of adapting a pretrained video diffusion model for physics simulation, comparing training regimes, conditioning mechanisms, and VAE adaptation strategies, including physics-informed decoder losses. 
  \item We demonstrate that the resulting differentiable surrogate enables gradient-based inverse optimization of building layouts for pedestrian wind comfort.
\end{itemize}

The code and model weights can be obtained via \href{https://github.com/rbischof/windinet}{https://github.com/rbischof/windinet}, and the dataset is shared upon request.

\section{Related Work}
\label{sec:related_work}

Pedestrian wind comfort depends on transient flow phenomena (vortex shedding, recirculation zones, shear-layer instabilities) that are computationally expensive to resolve numerically~\cite{MIRZAEI2021102839}. A large body of work has therefore pursued data-driven surrogates trained on CFD datasets. Convolutional neural networks (CNNs) and CNN--generative adversarial network (GAN) hybrids map building footprints to mean velocity or wind-factor fields~\cite{mokhtar2021pedestrian,lu2025machine,snaiki2025hierarchical,clemente2024rapid,kastner2023gan,clarke2025mlpmixer}. Graph neural networks operate on unstructured meshes at city scale~\cite{liu2023sgmsgnn,giral2025graphdiffusion}, including physics-informed variants that embed Reynolds-Averaged Navier–Stokes (RANS) residuals in the loss~\cite{shao2023pignn}. Physics-informed neural networks have also been applied to reconstruct 3D wind fields around buildings from sparse measurements~\cite{rui2023pinn}, demonstrating that embedding governing equations in the training loss can compensate for limited data. These approaches predict temporally averaged or steady-state conditions. However, the wind comfort criteria applied in practice~\cite{lawson1978wind,isyumov1975ground} are formulated as exceedance probabilities over threshold speeds, which require time-resolved velocity data to evaluate. Although Fourier Neural Operator (FNO)~\cite{li2021fourier} variants~\cite{qin2024localfno,chen2025fnourban} can predict instantaneous states and generate temporal sequences through autoregressive rollout, they struggle with generating long rollouts because errors accumulate over successive steps.

Three recent threads suggest how to close this gap. First, diffusion models have been shown to generate accurate time-dependent physical fields~\cite{du2024confil,gencfd}. Second, partial differential equation (PDE) solving has been recast as video generation~\cite{li2025videopde}. Third, foundation models pretrained on diverse equation families~\cite{herde2024poseidon,nguyen2025physix,soares2025pdefm,wiesner2025gphyt}, enabled by large-scale PDE benchmarks~\cite{takamoto2022pdebench,ohana2024well}, improve sample efficiency on downstream scientific tasks. These developments motivate formulating the unsteady urban wind flow simulation as a conditional video generation problem in which temporal coherence and long-range spatial dependencies are learned from the domain-specific data.

The video generation literature provides the architectural backbone for this idea. Early models were GAN-based~\cite{vondrick2016generating,saito2017tgan,tulyakov2018mocogan} and confined to short, low-resolution clips. Diffusion models removed this limitation, first by adding temporal attention to 3D U-Nets~\cite{ho2022videodiffusion}, then by shifting denoising into a learned latent space~\cite{blattmann2023stable}, and most recently by replacing the U-Net with transformers over patchified latent tokens~\cite{Wan22,hacohen2024ltx}. These efforts have produced models that synthesize temporally coherent video of complex scenes with long-range dependencies across both space and time. These models also encode physical priors that enable zero-shot reasoning about gravity, collisions, and fluid-like behavior~\cite{wiedemer2025videoreasoning}. Several recent methods exploit this observation: Force Prompting~\cite{gillman2025force}, DiffPhy~\cite{zhang2025think}, PhysCtrl~\cite{wang2025physctrl}, and PhysVideoGenerator~\cite{satish2026physvideogen} condition video diffusion on forces, physical cues, or trajectories to improve the plausibility of generated motion. The evaluation of prediction performance in these approaches, however, remains perceptual (human preference or FID-type scores), and none of them targets quantitative agreement with a governing PDE.

Beyond forward prediction, a growing body of work uses surrogates to accelerate the optimization of urban layouts for wind comfort. Wu and Quan~\cite{wu2024surrogatereview} survey the field and identify three dominant strategies: evolutionary methods coupled with CFD~\cite{kaseb2022cfdbased}, GAN-based surrogates paired with genetic algorithms~\cite{huang2022accelerated}, and response-surface models fitted to parametric CFD sweeps. All of these treat the surrogate as a black-box evaluator and rely on derivative-free optimizers that require many candidate evaluations per iteration. In contrast, an end-to-end differentiable surrogate can offer gradient-based layout optimization when combined with a soft rasterizer inspired by differentiable rendering~\cite{liu2019softrasterizer} to map continuous building coordinates to occupancy masks.


\newpage
\section{Methodology}
\label{sec:method}

We consider 2D wind flow over a square urban domain of side length $L$ (in meters). The built environment is described by a binary building footprint $B \in \{0,1\}^{H \times W}$, where $B_{x,y} = 1$ indicates a solid structure and $B_{x,y} = 0$ indicates open space. The corresponding fluid mask is $F = 1 - B$. Without loss of generality, wind enters from the left boundary at a prescribed inlet speed $u_\mathrm{in}$ (in m/s) and exits through the right. Other wind directions are obtained by rotating the building footprint before inference and rotating the predicted field back. The flow evolves over time from the uniform inlet condition to a quasi-steady state shaped by the building geometry.

We cast wind field prediction as a conditional video generation task. Given $(B, u_\mathrm{in}, L)$, the goal is to generate the velocity field sequence $\{\mathbf{w}_t\}_{t=1}^{T}$, where each $\mathbf{w}_t = (u_t, v_t) \in \mathbb{R}^{H \times W \times 2}$ contains the horizontal and vertical velocity components at time step $t$.

\begin{figure}[tbp]
    \centering
    \includegraphics[width=\linewidth]{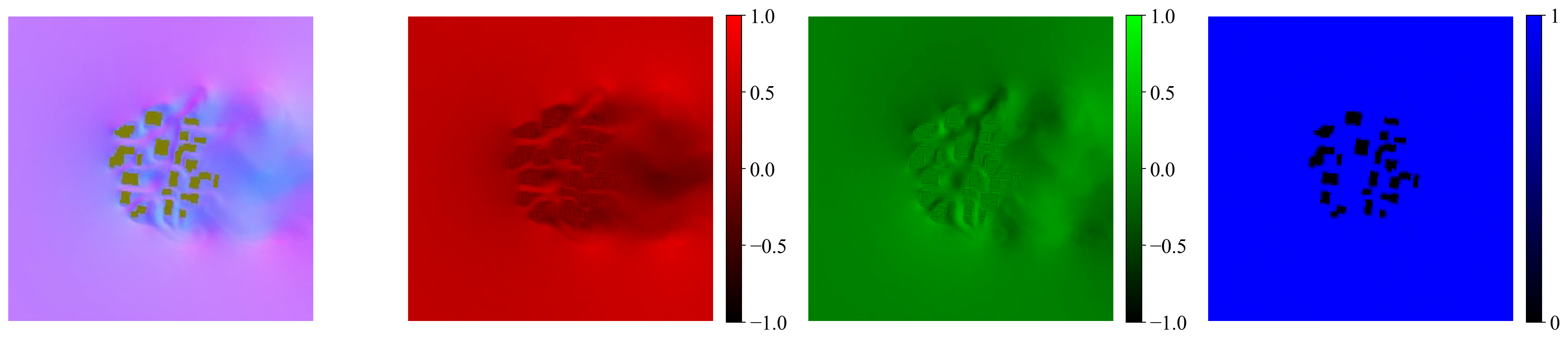}
    \caption{
    Channel decomposition of a single simulation frame. From left to right: encoded RGB composite, red channel encoding horizontal velocity $u$, green channel encoding vertical velocity $v$, and blue channel encoding the fluid mask ($1$: fluid, $0$: building). Velocity values are linearly rescaled to $[-1, 1]$ using the dataset-wide maximum speed $u_{\max}$.
    }
    \label{fig:rgb_encoding}
\end{figure}

\label{sec:encoding}
To leverage a pretrained RGB video model, we encode the physical velocity fields as pixel values (Figure~\ref{fig:rgb_encoding}). Each frame maps the two velocity components $(u, v)$ to the red and green channels of an RGB image by linearly rescaling with a dataset-wide maximum speed $u_{\max}$ so that values lie in $[-1, 1]$. The blue channel encodes the fluid mask $F$.

A directional conditioning image is prepended as frame $t{=}0$: the red channel is set to $u_\mathrm{in}/u_{\max}$ everywhere in fluid cells and zero inside buildings, the green channel is zero, and the blue channel carries the fluid mask. This conditioning signals both the geometry and the inlet condition to the model.

While this RGB encoding is convenient for training and inference, the per-channel differences are visually subtle. For all result figures, we therefore convert to wind speed magnitude $\|\mathbf{w}\| = \sqrt{u^2 + v^2}$ and apply a \texttt{coolwarm} colormap to better highlight flow structures.

\subsection{Base Model}
\label{sec:base_model}
We build upon LTX-Video~\cite{hacohen2024ltx}, a transformer-based latent video diffusion model that generates all frames jointly in a single denoising pass, rather than autoregressively one frame at a time. Specifically, we use the 2B-parameter text-and-image-to-video (TI2V) variant, which comprises three components: (i)~a causal 3D VAE that compresses video into a compact latent representation, (ii)~a T5-based text encoder for conditioning, and (iii)~a diffusion transformer (DiT)~\cite{peebles2023dit} that performs denoising in latent space by flow matching. We consider both full fine-tuning and LoRA fine-tuning for the DiT in our experiments.

\subsection{VAE Adaptation}
\label{sec:vae_adaptation}
The pretrained LTX-Video VAE was designed for natural RGB video. When applied directly to our wind-field encoding, it introduces reconstruction artifacts due to the domain gap between photorealistic content and pseudo-colored velocity fields (cf.\ Figure~\ref{fig:vae_reconstructions}). Modifying the encoder would shift the latent distribution and corrupt the DiT's pre-trained denoising dynamics. We therefore keep the encoder frozen and instead learn a color mapping around the frozen VAE, or fine-tune only the decoder~\cite{valevski2024gamengen}. Both strategies are applied in a dedicated stage before training the diffusion model.

\textit{Color adapter.} Our mapping from physical quantities ($u$, $v$, building footprint) to RGB channels is ultimately arbitrary (cf.\ appendix~\ref{sec:appendix_channel_ablation}
 for an ablation).
Rather than hand-picking a mapping, we can let the network learn one: a shallow Multilayer Perceptron (MLP) (3\,$\to$\,32\,$\to$\,3 with SiLU activation and tanh output) before the encoder transforms the wind-field encoding into a new, learned color space, and an equivalent MLP after the decoder maps back to physical channels. Both adapters are trained end-to-end with a frozen VAE.

\textit{Decoder fine-tuning.} A different approach is to fine-tune the decoder weights while keeping only the encoder frozen~\cite{valevski2024gamengen}. The reconstruction loss can optionally be augmented with physics-based regularizers: a \emph{divergence penalty} enforcing incompressibility ($\nabla \cdot \mathbf{w} = 0$), a \emph{no-penetration penalty} at building walls, and distance-weighted Mean Squared Error (MSE) that upweights fluid cells near building boundaries where velocity gradients are steepest (full formulations in the appendix~\ref{sec:appendix_physics_loss}
). Physics-informed losses are rarely combined with latent diffusion models because they operate in pixel space, and every training step would require decoding the full output and backpropagating through the decoder. Our preliminary decoder fine-tuning removes this obstacle, since these physics-informed losses are applied in a short stage before the diffusion model is trained. Approximately 5k optimization steps suffice for a significant improvement in reconstruction quality. Table~\ref{tab:variants} summarizes the resulting VAE configurations.

\begin{table}[t]
  \centering
  \caption{VAE adaptation variants. Color Adapter learns a nonlinear channel transformation around the frozen VAE. Dec.\ FT fine-tunes the decoder weights. Dec.\ FT Physics adds physics-informed losses during decoder fine-tuning.}
  \label{tab:variants}
  \small
  \begin{tabular}{@{}lccc@{}}
    \toprule
    \textbf{Variant} & \textbf{Color adaptation} & \textbf{Decoder} & \textbf{Physics losses} \\
    \midrule
    Base                & --         & frozen     & --         \\
    Color Adapter       & \checkmark & frozen     & --         \\
    Dec.\ FT      & --         & fine-tuned & --         \\
    Dec.\ FT Physics  & --         & fine-tuned & \checkmark \\
    \bottomrule
  \end{tabular}
  \vspace{-1em}
\end{table}

\subsection{Conditioning Strategies}
\label{sec:conditioning}
The original LTX-Video model is conditioned on text prompts via cross-attention. While text can describe simulation parameters (e.g., \textit{``inlet speed 18 m/s, domain size 1300 m''}), natural language is an imprecise interface for continuous physical quantities. We therefore introduce an alternative \emph{scalar conditioning} mechanism.

A learnable embedding module maps $u_\mathrm{in}$ and $L$ directly into the transformer's conditioning space, bypassing the text encoder entirely. Each scalar is normalized to $[0,1]$, encoded via Fourier features~\cite{tancik2020fourier} with log-spaced frequencies, and projected by a small MLP into embedding tokens that replace the text encoder output in the transformer's cross-attention.

\section{Experiments}
\label{sec:experiments}

\subsection{Dataset}
\label{sec:dataset}

\begin{figure}[t]
    \centering
    \includegraphics[width=\linewidth]{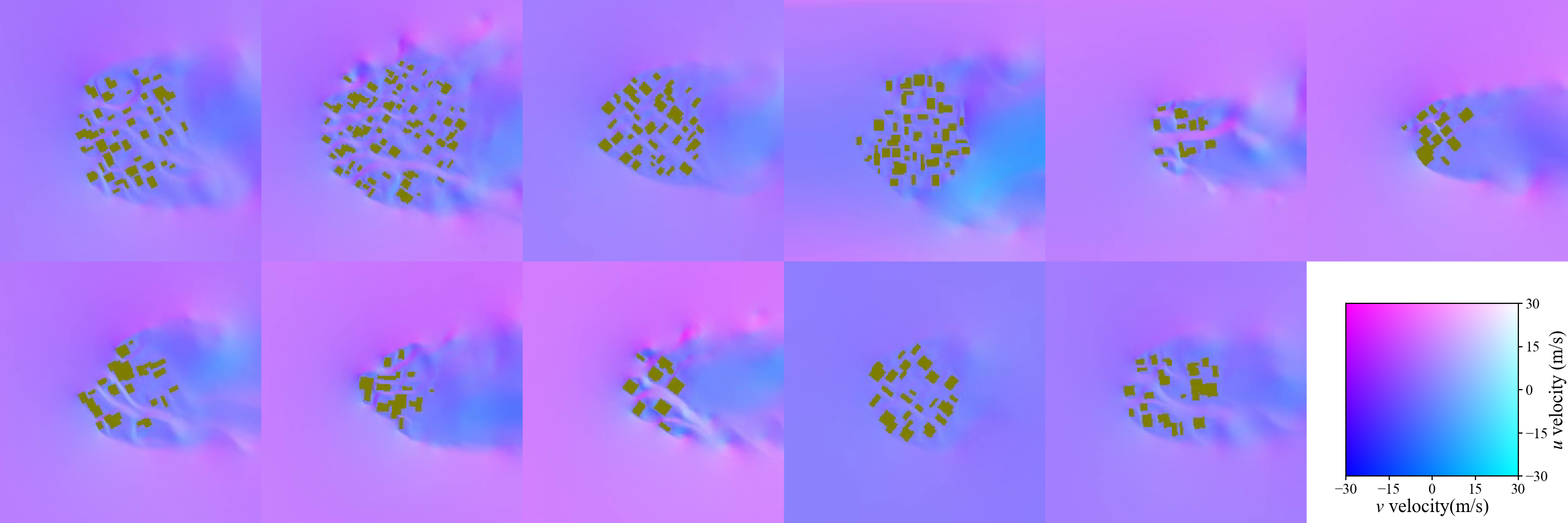}
    \caption{
    Representative samples from the training dataset. Each tile shows the RGB-encoded velocity field at frame 100 for a distinct simulation.
    }
    \label{fig:synthetic_dataset}
\end{figure}

We generate a dataset of 2D urban wind simulations using an incompressible Euler solver and procedural building footprint generator~\cite{folk2024learning}. The incompressibility assumption is standard for pedestrian comfort and building aerodynamics studies, where typical wind speeds keep Mach numbers well below compressible regimes~\cite{letzel2008}. Building footprints consist of randomly placed rectangular blocks ($10$ to $50$\,m, min. 10\,m alley width) within a circular city region whose diameter is sampled from $\{300, 400, \ldots, 800\}$\,m, with block counts scaling proportionally with area. A 300\,m buffer surrounds each city to allow flow development and reduce boundary effects. Together, the city region and buffer zone thus form the full domain with side length $L$ in $[900, 1400]\,\text{m}$. The inlet wind speeds $u_\mathrm{in}$ are sampled from $[0.1, 20]\,\text{m/s}$, covering the range of conditions relevant to pedestrian comfort and safety assessments~\cite{EN1991-1-4,CoLWind2019}. Finally, inlet wind direction is sampled uniformly from $[0, 360]^\circ$, and each simulation is subsequently rotated so that wind flows left-to-right. This canonicalization removes wind direction as a degree of freedom, leaving only $u_\mathrm{in}$, domain size $L$, and the domain geometry as varying parameters. Each simulation covers 112\,s of physical time and is stored as $T{=}112$ velocity snapshots at $256{\times}256$ resolution plus an initial conditioning frame (Sec.~\ref{sec:encoding}). The dataset comprises 10{,}000 training, 1{,}000 validation, and 2{,}000 test simulations. Randomly selected samples are shown in Figure~\ref{fig:synthetic_dataset}.

\subsection{Training}
\label{sec:training}

Training uses AdamW~\cite{loshchilov2019adamw} ($\text{lr}{=}10^{-4}$, cosine schedule), batch size 64 for 2{,}000 steps. The model variants trained with full fine-tuning used AdamW ($\text{lr}{=}10^{-5}$, cosine schedule) and batch size 64 for 10{,}000 steps (156 steps/epoch, ${\sim}$64 epochs) for DiT transformer training. The variants with VAE decoder fine-tuning trained the decoder separately prior to DiT fine-tuning using the same configurations.

We evaluate generated wind fields against the ground truth using five metrics, computed on fluid pixels only: \textit{MAE} (m/s), \textit{MRE} (\%), \textit{VRMSE} (variance-normalized RMSE; primary ranking metric), \textit{spectral divergence} (temporal frequency fidelity), and \textit{Wasserstein-1 distance} $W_1$ (speed distribution accuracy). The full metric definitions are provided in the appendix~\ref{sec:appendix_metrics}.

\subsection{Main Results}
\label{sec:results}

\begin{table*}[h]
  \centering
  \caption{Results of baseline models and all WinDiNet variants on the test set. All metrics computed on fluid pixels only.}
  \label{tab:results}
  \vspace{0.5ex}
  \footnotesize
  \begin{tabular*}{\textwidth}{@{\extracolsep{\fill}}lllccccc}
    \toprule
    & \textbf{Model} & \textbf{Cond.}
      & \textbf{VRMSE $\downarrow$} & \textbf{MAE $\downarrow$}
      & \textbf{MRE $\downarrow$}
      & \textbf{Spectral $\downarrow$} & $\mathbf{W_1}\downarrow$ \\
    \cmidrule{1-8}
    \multirow{6}{*}{%
      \begin{tabular}{@{}c@{}}
        Baselines
      \end{tabular}
    }
    &U-Net~\cite{ronneberger2015unet}          &   \multirow{6}{*}{Scalar}   & 1.467 & 4.73 & 15.46 & 2.74 & 7.06 \\
    &Poseidon~\cite{herde2024poseidon}       &      & 0.781 & 2.17 & 7.05  & 2.40 & 2.79 \\
    &AFNO~\cite{guibas2022efficient}           &      & 0.618 & 1.44 & 4.61 & 1.71 & 1.31 \\
    &FNO~\cite{li2021fourier}            &      & 0.610 & 1.42 & 4.53 & 1.68 & 1.32 \\
    &OFormer~\cite{li2023transformer}        &      & 0.585 & 1.36 & 4.34 & \textbf{1.52} & 1.22 \\
    &RNO~\cite{liuschiaffini2023tipping}            &      & 0.563 & 1.19 & 3.91 & 1.72 & 1.14 \\
    \cmidrule[\heavyrulewidth]{1-8}
    \multirow{6}{*}{%
      \begin{tabular}{@{}c@{}}
        WinDiNet \\
        (ours)
      \end{tabular}
    }
     & LoRA  & \multirow{2}{*}{Text}      & 0.746 & 1.56 & 5.20 & 1.84 & 1.29 \\
     & Base           &         & 0.702 & 1.36 & 4.53 & 1.74 & 1.08 \\
    \cmidrule{2-8}
    & Base          &     \multirow{4}{*}{Scalar}      & 0.616 & 1.12 & 3.75 & 1.61 & 0.95 \\
    & Color Adapter  &          & 0.577 & 1.11 & 3.72 & 1.63 & 0.91 \\
    & Dec.\ FT   &        & 0.531 & 1.02 & 3.37 & 1.55 & \textbf{0.83} \\
    & Dec.\ FT Physics  &     & \textbf{0.520} & \textbf{1.01} & \textbf{3.33} & 1.54 & \textbf{0.83} \\
    \bottomrule
  \end{tabular*}
\end{table*}

Table~\ref{tab:results} compares WinDiNet against six neural operator baselines on the full inference pipeline, where the diffusion model generates latent sequences that are decoded into velocity fields. The baselines fall into three performance tiers: autoregressive models (OFormer, RNO) perform best, followed by one-shot predictors (AFNO, FNO), with frame-to-frame models that must be rolled out autoregressively at inference (U-Net, Poseidon) trailing behind. RNO achieves the lowest baseline VRMSE at 0.563.

Full fine-tuning of the DiT transformer reduces VRMSE by 5.9\% over LoRA (rank 512); therefore, all subsequent variants use full fine-tuning. Replacing text conditioning with scalar embeddings yields a further 12\% reduction, bringing the scalar base model (VRMSE\,=\,0.616) into the range of the stronger baselines despite operating through a lossy VAE bottleneck that none of the baselines require. Text conditioning tokenizes continuous physical quantities as strings, a representation that the pretrained text encoder was not designed to handle. The scalar embedding injects simulation parameters directly into the transformer's conditioning pathway, which improves both accuracy and efficiency.

\begin{figure}[!htbp]
  \centering
  \includegraphics[width=\linewidth]{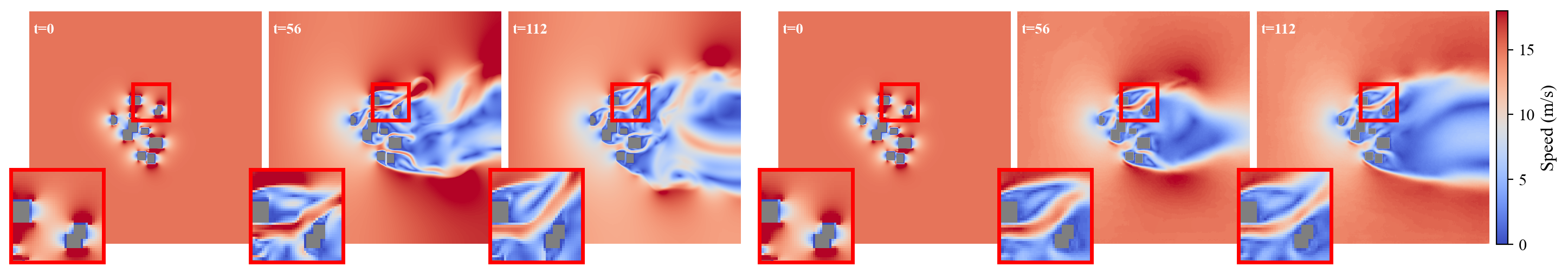}
  \caption{Wind speed magnitude predicted by Dec.\ FT Physics for a procedurally generated urban layout at $15\,\mathrm{m/s}$ inlet velocity. Ground truth (left) and model prediction (right) at timesteps $t\!=\!0$, $56$, and $112$.}
  \label{fig:physics_example}
\end{figure}

With VAE adaptation, the color adapter brings WinDiNet close to the best baselines, and decoder fine-tuning pushes it beyond them. Dec.\ FT Physics (Fig.~\ref{fig:physics_example}) outperforms the best baseline (RNO) by 7.6\% in VRMSE and 15\% in MAE. These are the lowest scores across all metrics except spectral divergence, where OFormer retains a marginal lead (1.52 vs.\ 1.54).

To isolate the contribution of the pretrained video prior, we retrain the model under the same compute budget while removing pretraining. Training the DiT from scratch on top of the fine-tuned VAE raises VRMSE from 0.52 to 0.57, and training both the DiT and the VAE from scratch raises it further to 0.92. Pretraining therefore accounts for a large share of the model's accuracy, consistent with what has been reported for physical reasoning~\cite{wiedemer2025videoreasoning}.

\subsection{VAE Adaptation}
\label{sec:vae_adaptation_results}

\begin{figure*}[h]
  \centering
  \begin{subfigure}[b]{0.235\textwidth}
    \centering
    \includegraphics[width=\textwidth]{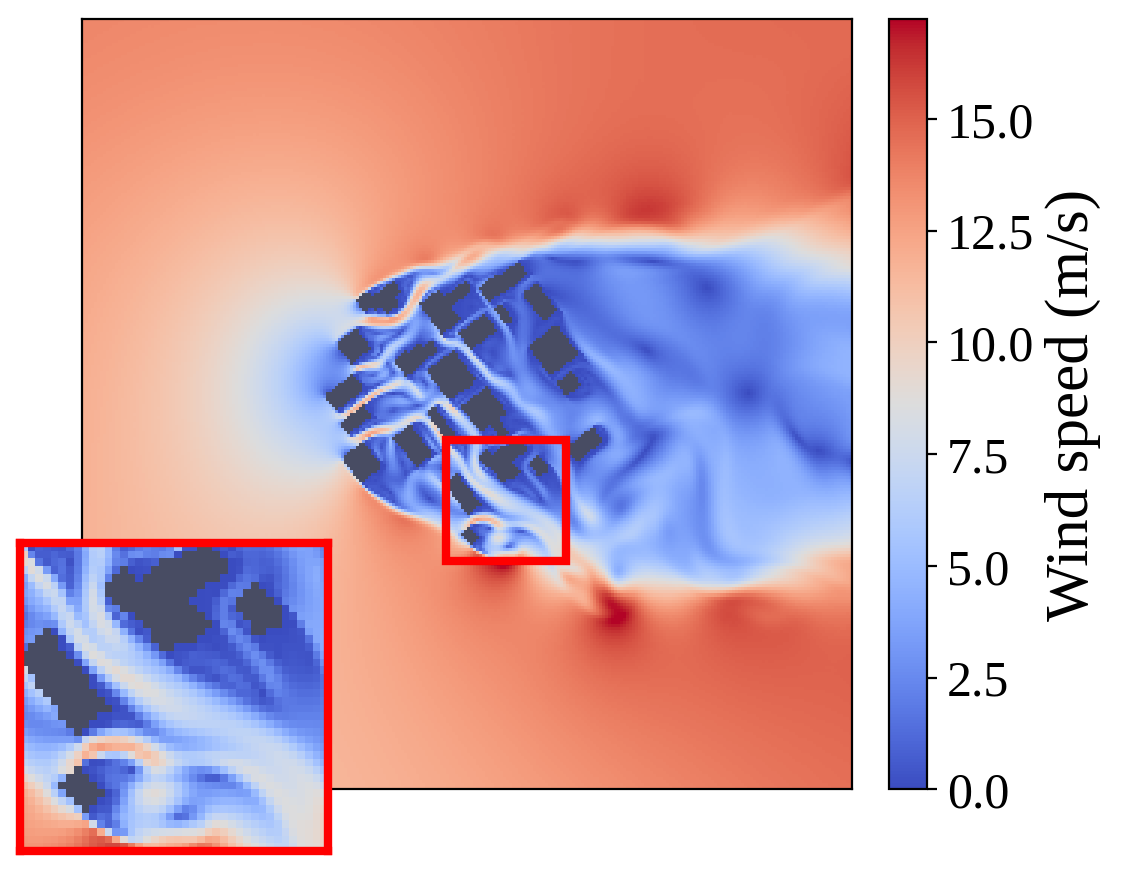}
    \caption{Ground truth}
    \label{fig:vae_gt}
  \end{subfigure}\hfill
  \begin{subfigure}[b]{0.235\textwidth}
    \centering
    \includegraphics[width=\textwidth]{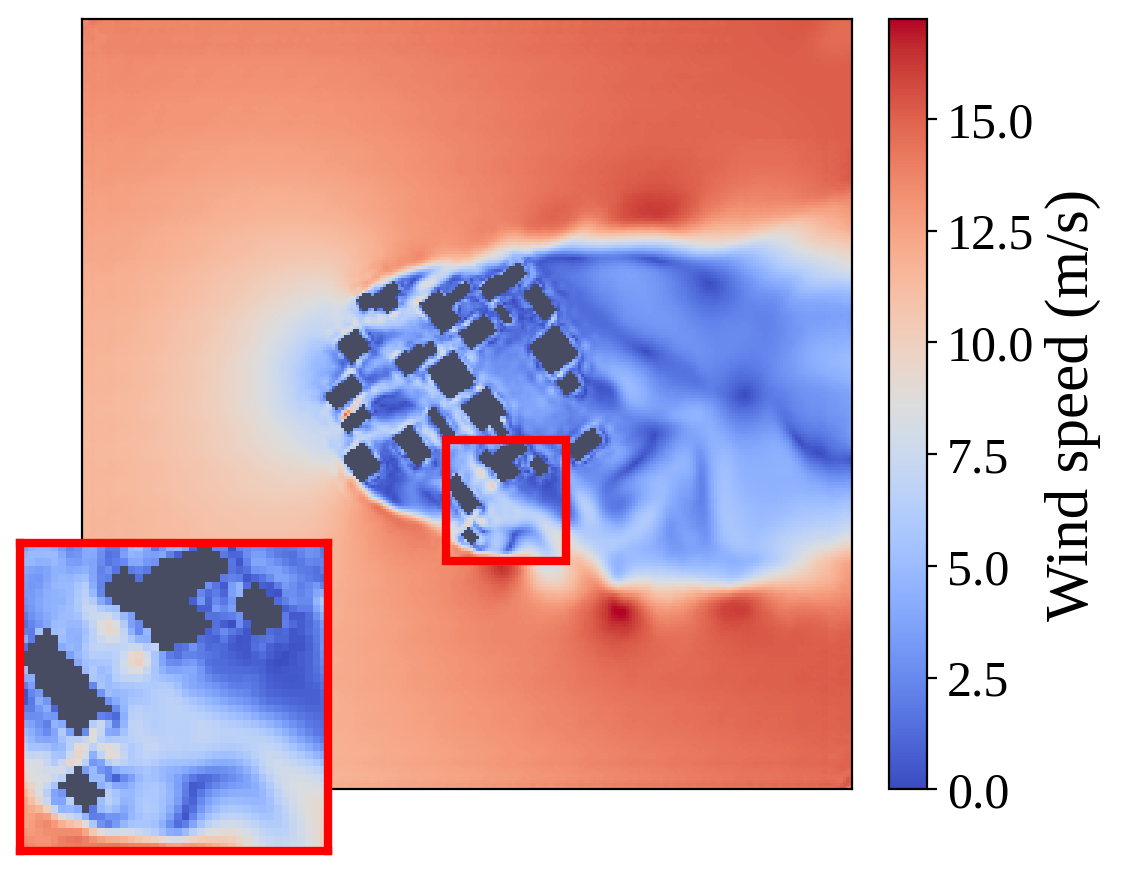}
    \caption{Base}
    \label{fig:vae_default}
  \end{subfigure}\hfill
  \begin{subfigure}[b]{0.235\textwidth}
    \centering
    \includegraphics[width=\textwidth]{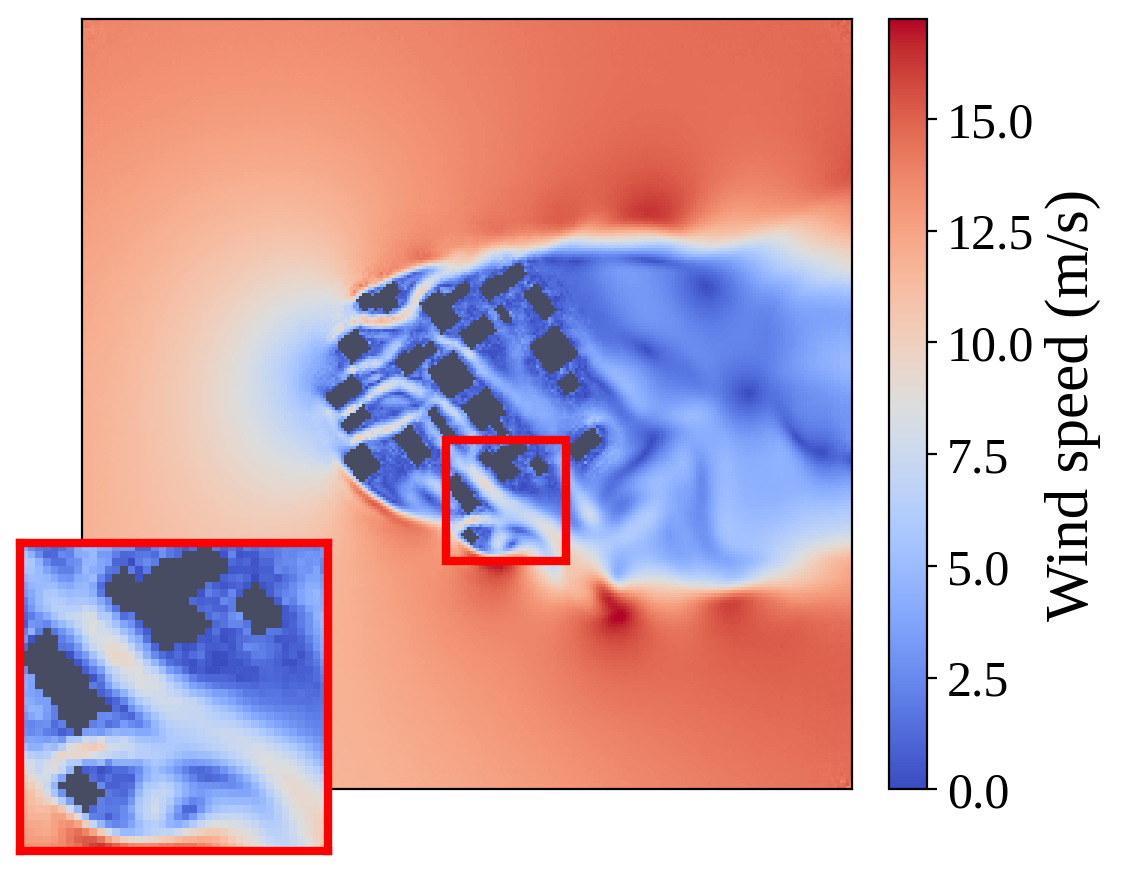}
    \caption{Dec.\ FT}
    \label{fig:vae_mse}
  \end{subfigure}\hfill
  \begin{subfigure}[b]{0.235\textwidth}
    \centering
    \includegraphics[width=\textwidth]{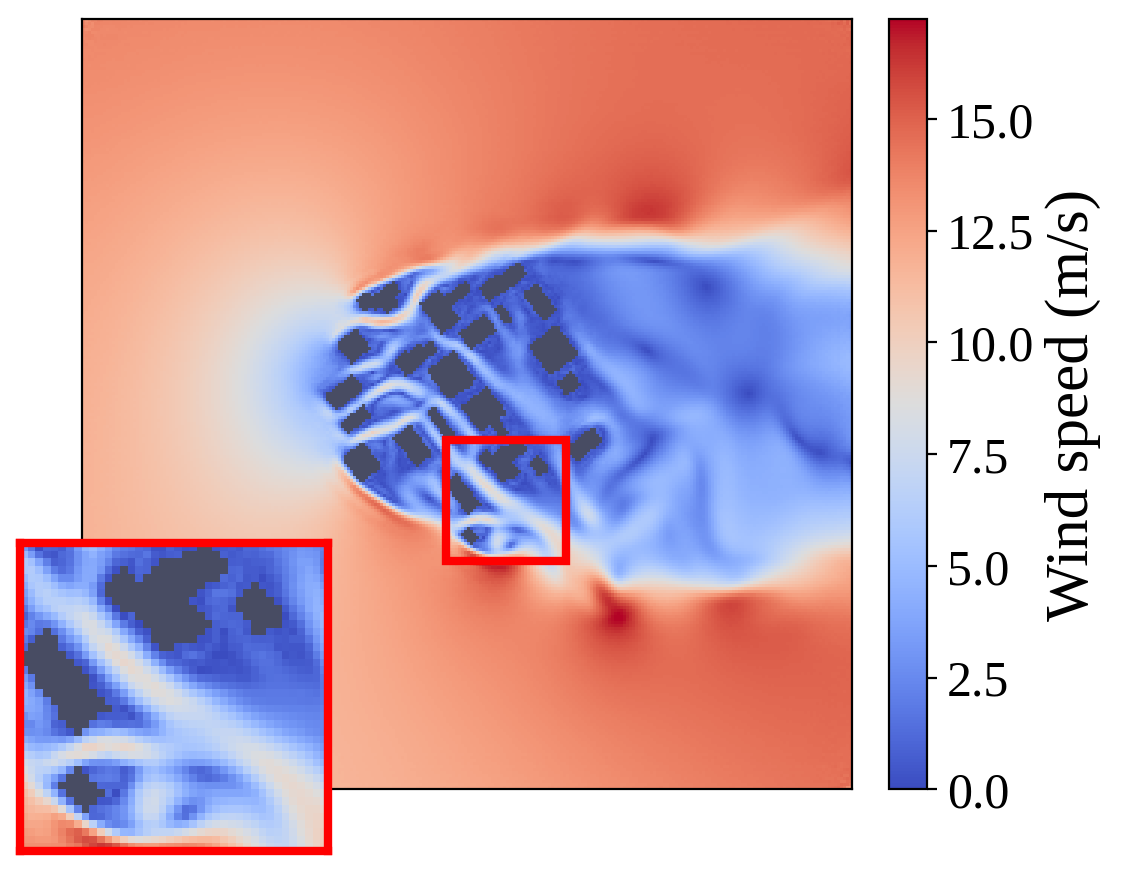}
    \caption{Dec.\ FT Physics}
    \label{fig:vae_physics}
  \end{subfigure}
  \caption{%
    VAE reconstruction quality at $t{=}90$ for a sample from the test set.
  }
  \label{fig:vae_reconstructions}
\end{figure*}

\begin{table}[tb]
  \centering
  \caption{VAE reconstruction quality (encode $\to$ decode) on the test set. Adapted variants trained for 5{,}000 steps. Base uses the pretrained VAE without modification.}
  \label{tab:vae_comparison}
  \vspace{0.5ex}
  \small
  \begin{tabular*}{\tablewidth}{@{\extracolsep{\fill}}lccccc}
    \toprule
    \textbf{Config} & \textbf{VRMSE$\downarrow$} & \textbf{MAE$\downarrow$} & \textbf{MRE$\downarrow$} & \textbf{Spectral$\downarrow$} & $\mathbf{W_1}\downarrow$ \\
    \midrule
    Base              & 0.188 & $1.63{\times}10^{-2}$ & 8.9\% & 0.781 & 0.0157 \\
    Color Adapter        & 0.157 & $1.60{\times}10^{-2}$ & 8.4\% & 0.704 & 0.0167 \\
    Dec.\ FT       & 0.084 & $5.97{\times}10^{-3}$ & 3.2\% & 0.461 & 0.0054 \\
    Dec.\ FT Physics   & \textbf{0.070} & $\mathbf{4.95{\times}10^{-3}}$ & \textbf{2.6\%} & \textbf{0.396} & \textbf{0.0045} \\
    \bottomrule
  \end{tabular*}
\end{table}

VAE adaptation is performed as a separate stage before diffusion model training. To evaluate its effect in isolation, we encode ground-truth velocity fields into the latent space and decode them back without involving the diffusion model (Table~\ref{tab:vae_comparison}). Every adaptation method improves reconstruction fidelity over the frozen baseline, with the best decoder variant reducing VRMSE by over 62\%. 

\begin{wrapfigure}{r}{0.4\linewidth}
  \vspace{-8pt}
  \centering
  \begin{subfigure}[b]{0.48\linewidth}
    \centering
    \includegraphics[width=\textwidth]{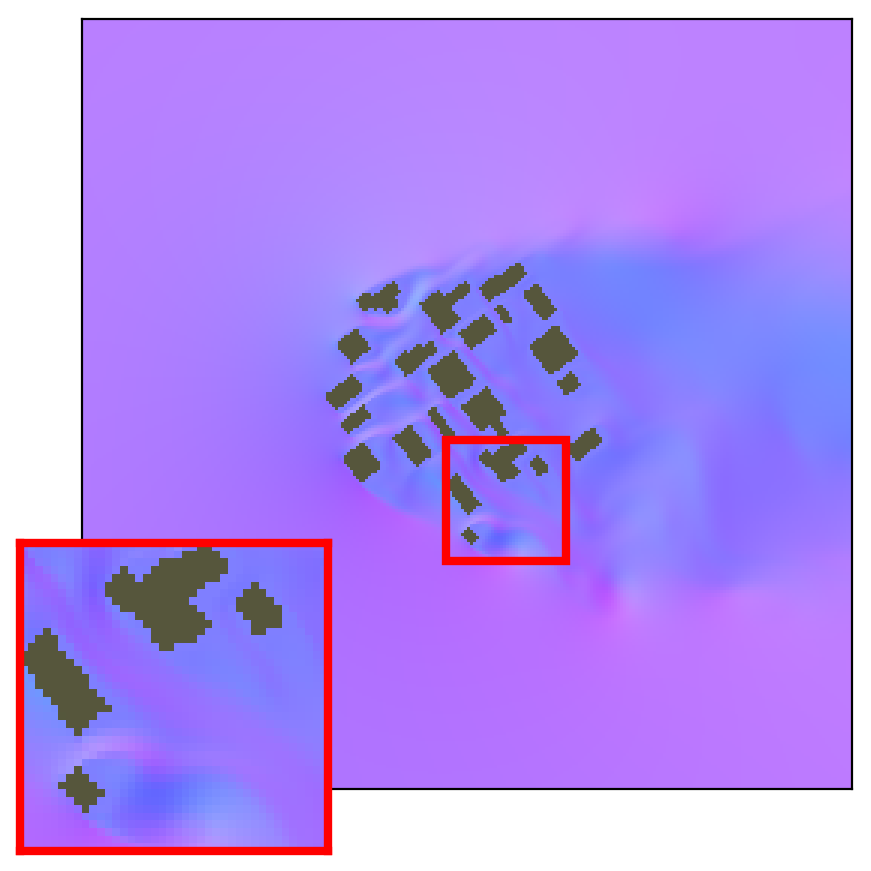}
    \caption{Input RGB}
    \label{fig:gt_rgb}
  \end{subfigure}\hfill
  \begin{subfigure}[b]{0.48\linewidth}
    \centering
    \includegraphics[width=\textwidth]{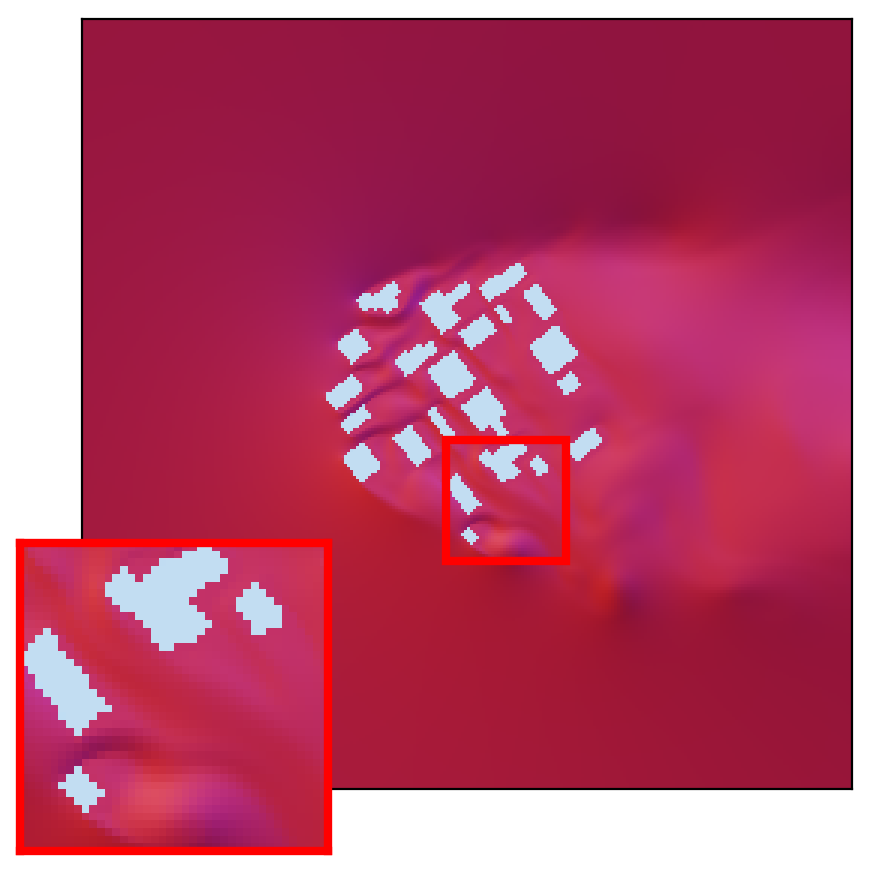}
    \caption{Color adapter}
    \label{fig:channeltransf_adapted}
  \end{subfigure}
  \caption{%
    Learned channel transformation by the color adapter.
  }
  \label{fig:channel_transform}
  \vspace{-10pt}
\end{wrapfigure}

The color adapter learns a nonlinear channel mapping that transforms the velocity field into a visually distinct palette (Fig.~\ref{fig:channel_transform}), primarily increasing contrast between buildings and fluid regions rather than sharpening fine-scale flow structures such as vortices. This may explain why the color adapter slightly degrades speed distribution fidelity, despite improving spatial reconstruction metrics (cf.\ Table~\ref{tab:vae_comparison}).
Given the color adapter's limited impact on distributional fidelity, we instead fine-tune the decoder. This proves substantially more effective, reducing the reconstruction VRMSE by more than 55\% relative to the base configuration and making the color adapter redundant. Fine-tuning allows the decoder to learn the domain-shifted distribution directly, rather than requiring latents to conform to pretrained expectations. As shown in Fig.~\ref{fig:vae_reconstructions}, this produces significantly sharper vortex boundaries. Adding a physics-informed loss provides a complementary objective that penalizes aspects of the flow that pixel-level losses do not capture, such as near-wall gradients. These physics-informed gains are primarily reflected in the quantitative metrics rather than in qualitative reconstructions.

The ranking of VAE adaptation strategies in the isolated reconstruction experiment (Table~\ref{tab:vae_comparison}) transfers directly to the full inference pipeline (Table~\ref{tab:results}), where each decoder variant produces velocity fields from the same diffusion-generated latents. VAE reconstruction quality is a reliable predictor of simulation accuracy: decoder-level improvements propagate through the generative process and are not degraded by stochastic sampling. Since the underlying diffusion model weights are shared across all VAE variants, the gap between the Scalar Base and Scalar Dec.\ FT Physics (15.6\% reduction in VRMSE) can be attributed to decoder adaptation.

\begin{wrapfigure}{r}{0.4\linewidth}
  \vspace{-15pt}
  \centering
  \includegraphics[width=\linewidth]{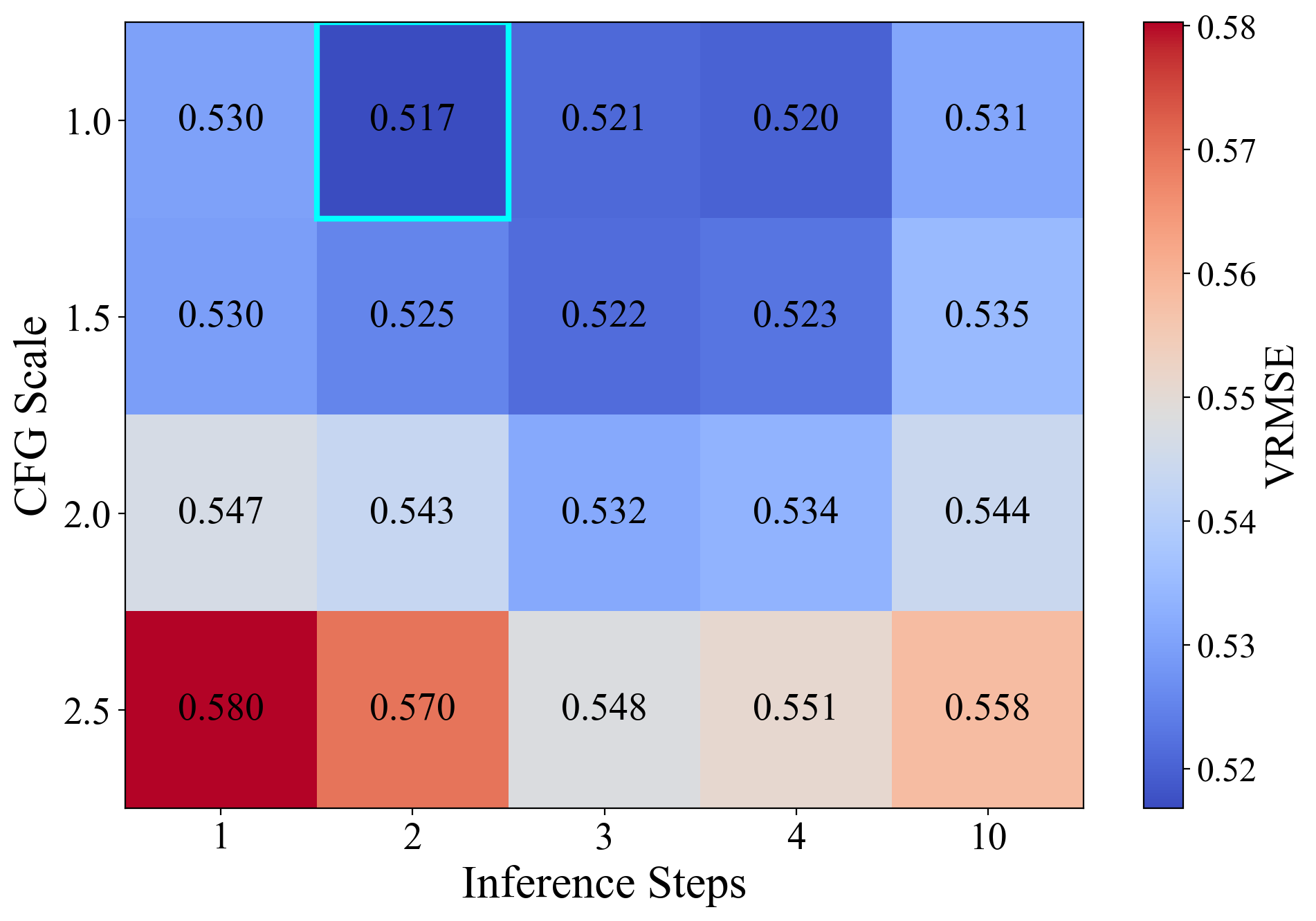}
  \caption{%
    Grid search over CFG scale and number of denoising steps on the validation set (Scalar conditioning, Dec.\ FT Physics).
  }
  \label{fig:parameter_sweep}
  \vspace{-10pt}
\end{wrapfigure}

\subsection{Inference Configuration}
\label{sec:inference_config}

We select the number of denoising steps and the classifier-free guidance (CFG) scale via grid search on the validation set (Fig.~\ref{fig:parameter_sweep}). The best configuration uses guidance scale 1.0 and 2 steps. At 2 denoising steps on a single NVIDIA H200 GPU, the model generates a full $(T{+}1)$-frame velocity field sequence at $256{\times}256$ resolution in approximately 0.32\,s in FP16, roughly three orders of magnitude faster than the incompressible Euler solver used to generate the training data. At this setting inference fits within 24\,GB of VRAM.

Two factors explain why so few steps are sufficient. First, LTX-Video uses rectified flow~\cite{hacohen2024ltx}, which typically requires fewer integration steps than conventional DDPM schedulers. 
Second, 2D urban wind fields are structurally simpler than natural video: there are no occlusions, no texture variety, and the dynamics are governed by a single PDE. Regarding CFG, natural video generation typically uses CFG scales of 3 or higher~\cite{hacohen2024ltx}. We found that higher CFG values produced visually sharper vortices and building boundaries, but did not improve quantitative metrics. Because our conditioning signal encodes physical quantities rather than semantic descriptions, scaling its contribution distorts the learned mapping from simulation parameters to velocity fields. We conclude that CFG should be disabled in conditioned diffusion models for physics simulation.

Additional ablations on temporal extrapolation, domain size variation, inlet speed generalization, and RGB channel assignment are provided in the appendix (Appendix~\ref{app:generalization}).

\newpage
\begin{figure}[h]
  \centering
  \begin{subfigure}[b]{0.165\textwidth}
    \centering
    \includegraphics[width=\textwidth]{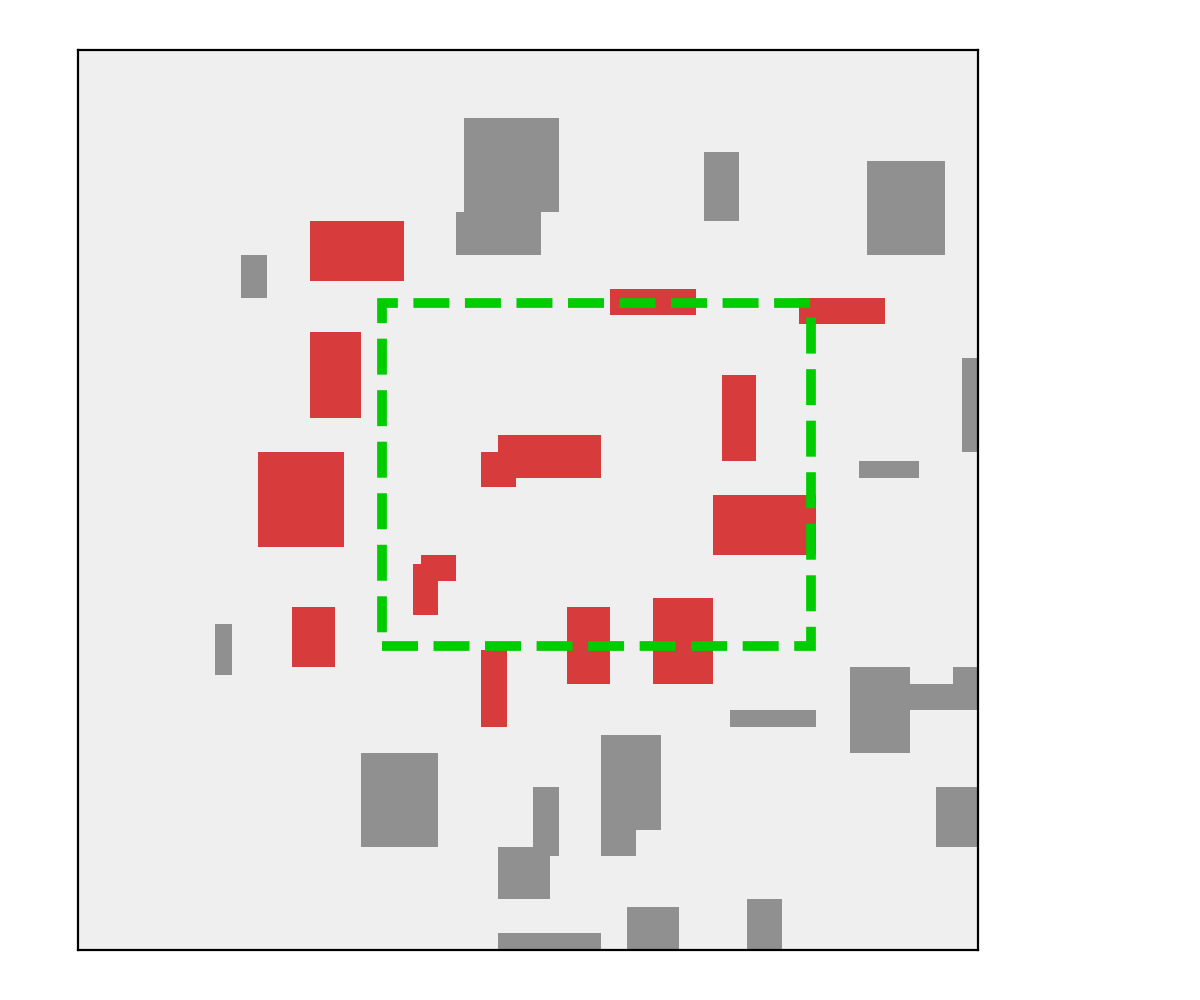}
    \caption{}
  \end{subfigure}\hfill
  \begin{subfigure}[b]{0.165\textwidth}
    \centering
    \includegraphics[width=\textwidth]{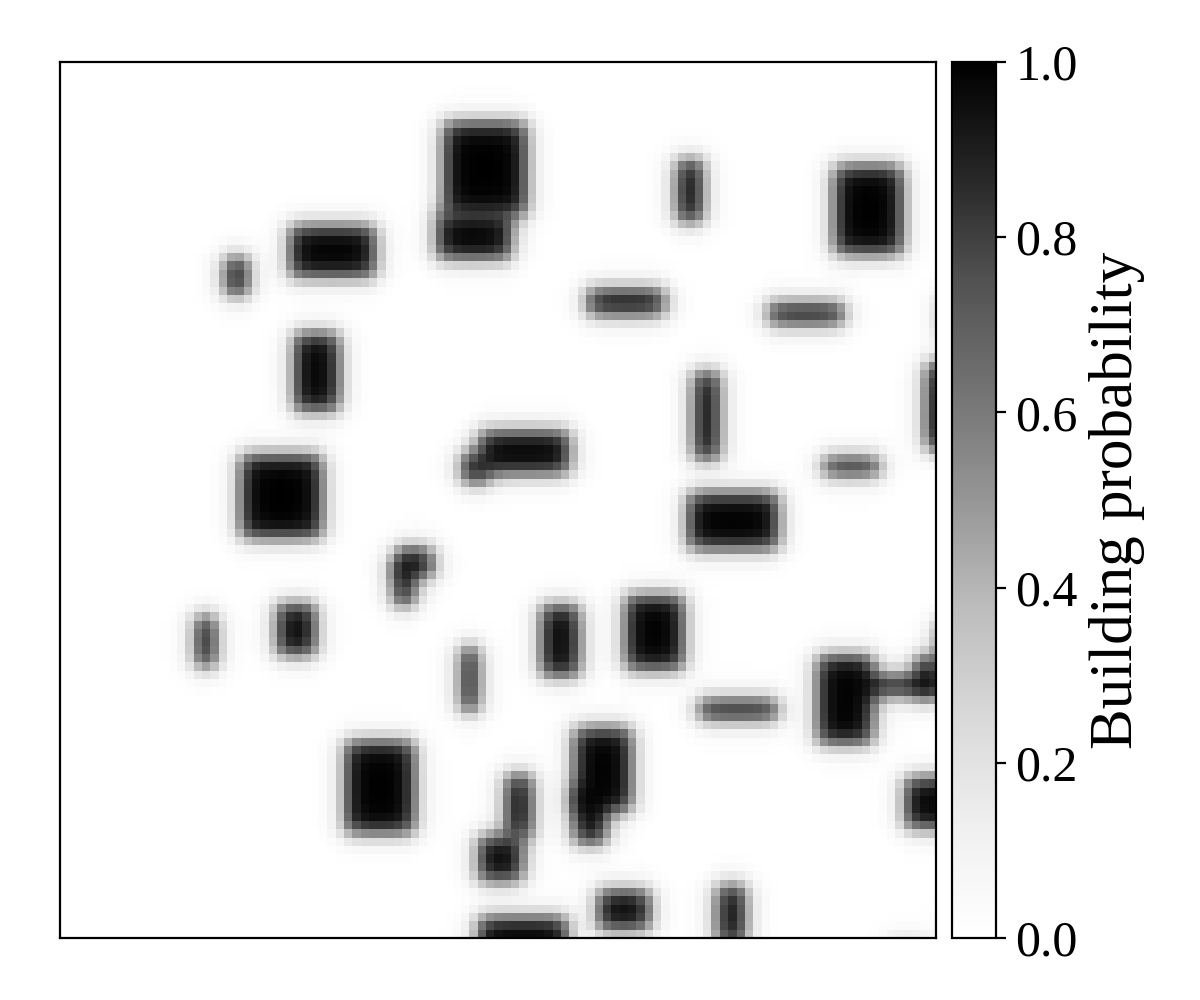}
    \caption{}
  \end{subfigure}\hfill
  \begin{subfigure}[b]{0.165\textwidth}
    \centering
    \includegraphics[width=\textwidth]{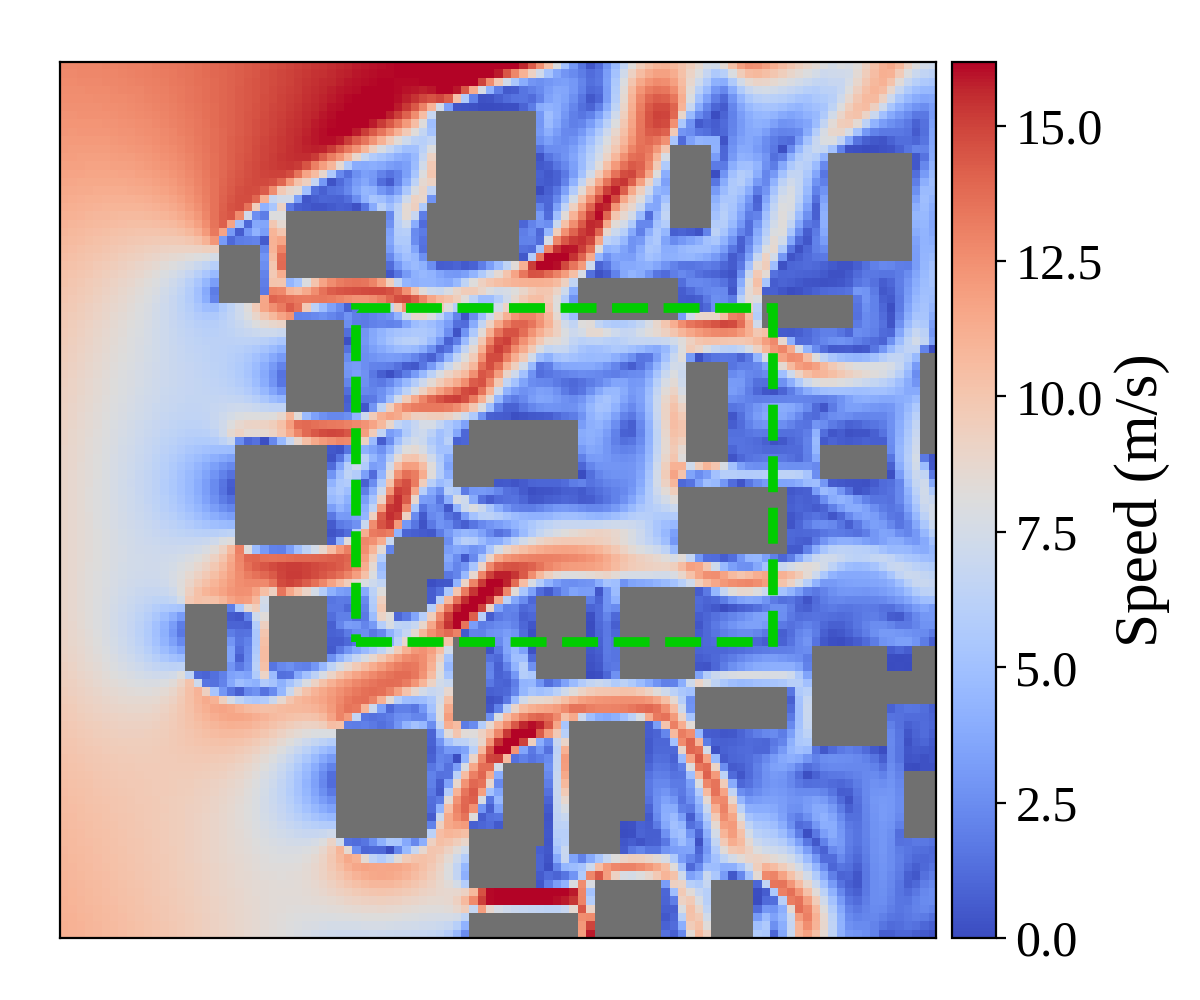}
    \caption{}
  \end{subfigure}\hfill
  \begin{subfigure}[b]{0.165\textwidth}
    \centering
    \includegraphics[width=\textwidth]{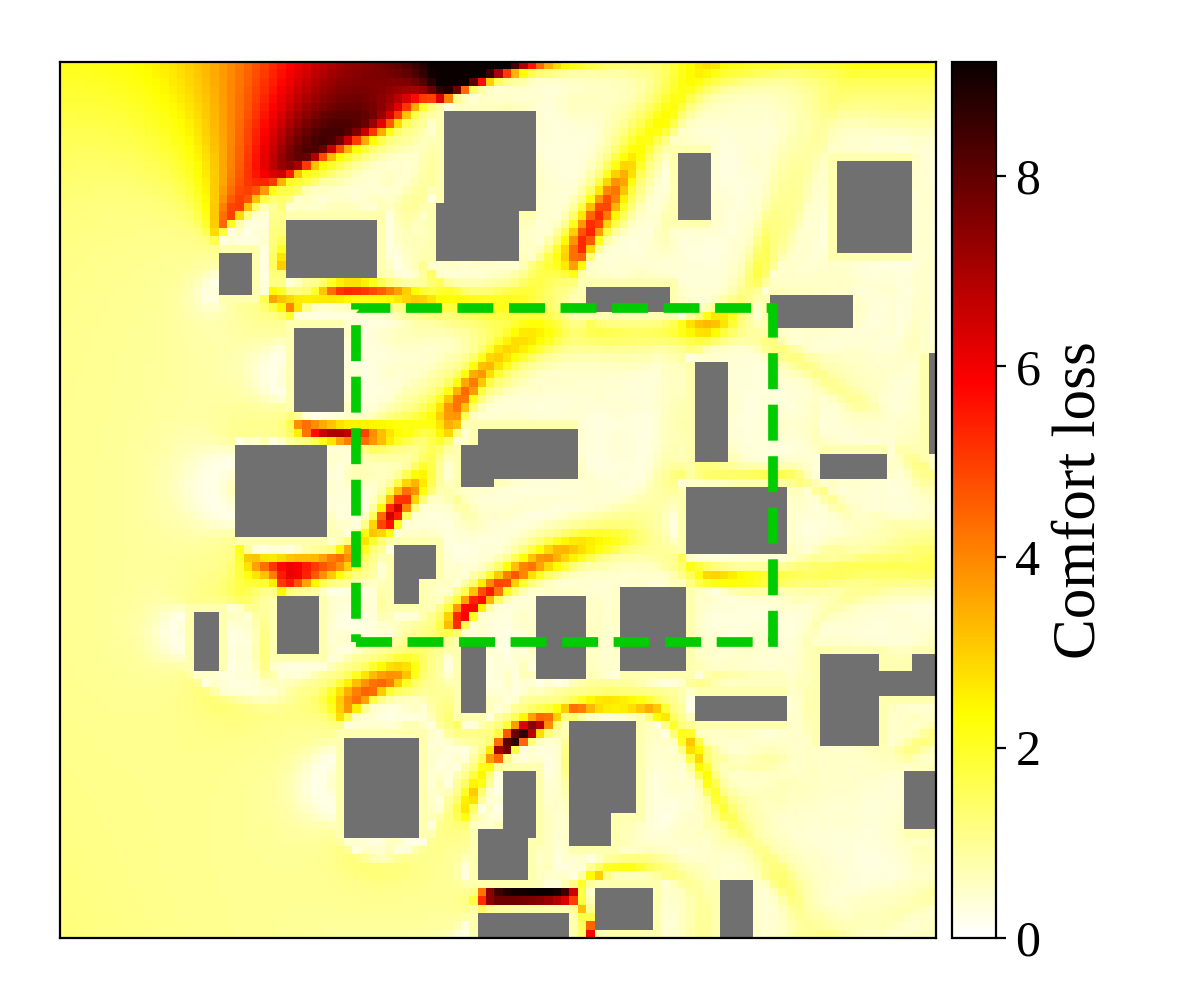}
    \caption{}
  \end{subfigure}\hfill
  \begin{subfigure}[b]{0.165\textwidth}
    \centering
    \includegraphics[width=\textwidth]{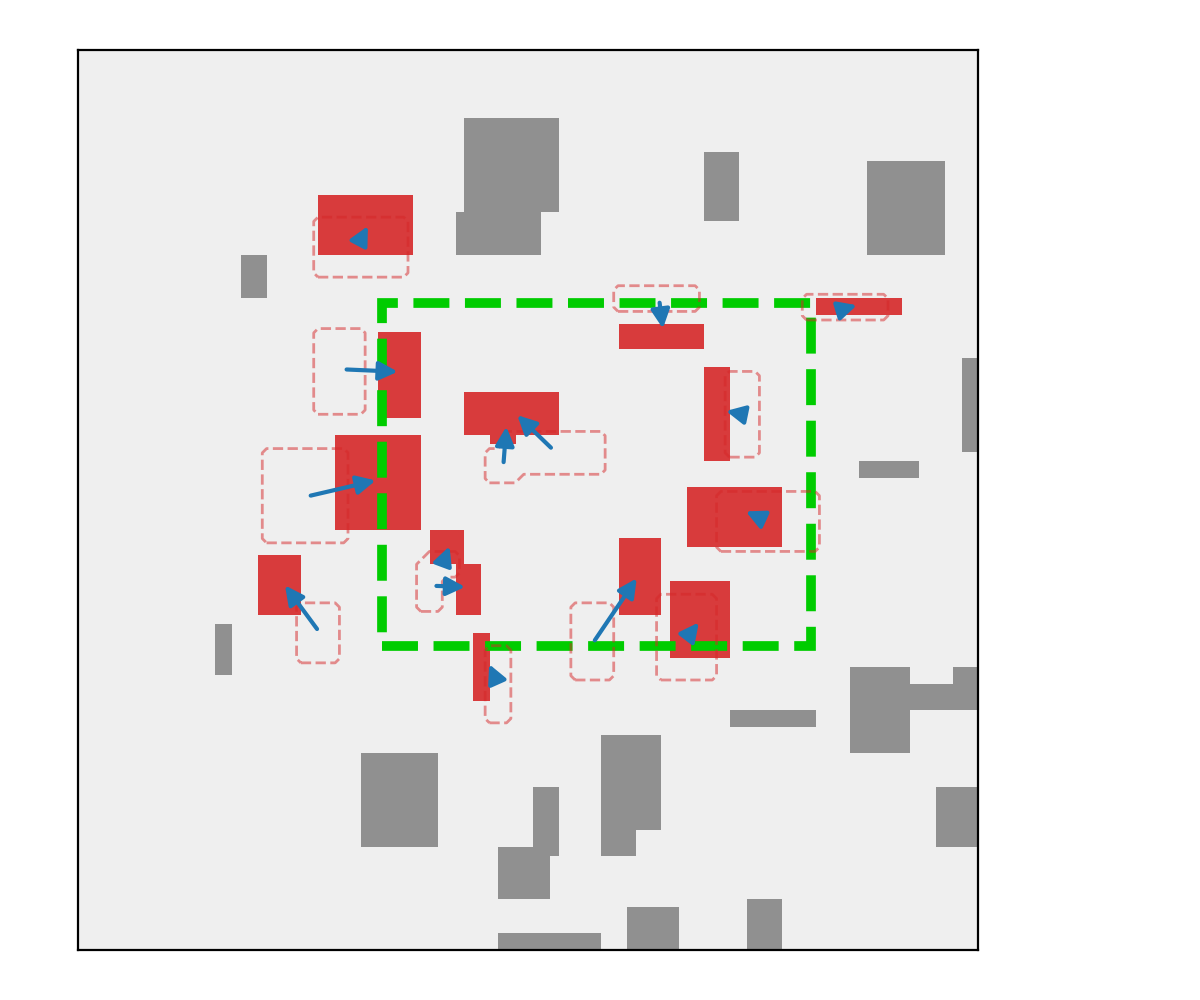}
    \caption{}
  \end{subfigure}\hfill
  \begin{subfigure}[b]{0.165\textwidth}
    \centering
    \includegraphics[width=\textwidth]{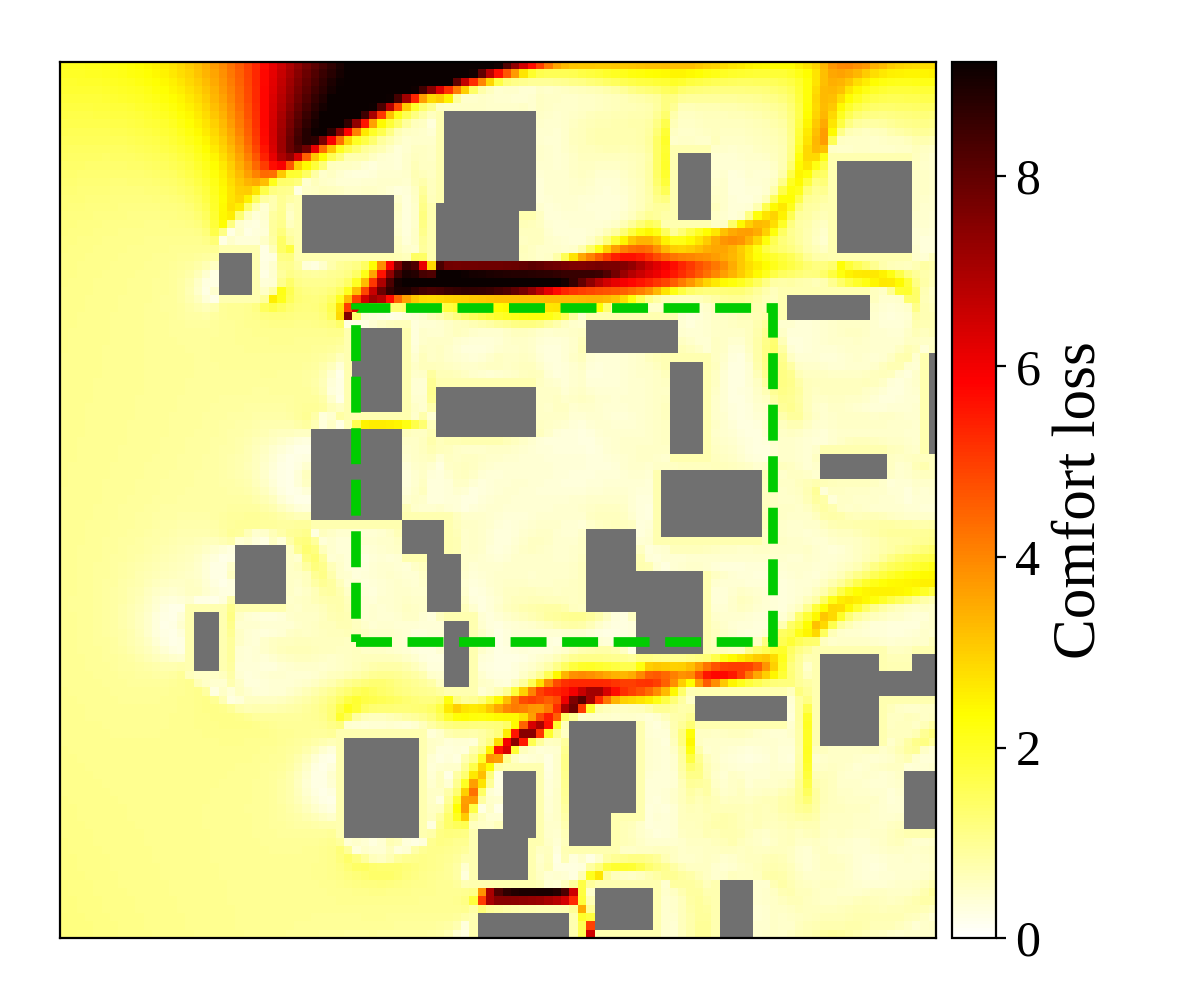}
    \caption{}
  \end{subfigure}
  \caption{%
    Inverse optimization pipeline, zoomed in for visibility.
    (a)~Movable buildings (red) and objective region (green rectangle) are defined.
    (b)~A differentiable rasterizer produces a soft (blurred) occupancy mask.
    (c)~WinDiNet predicts the wind field.
    (d)~Comfort loss penalizes speeds outside the target band.
    (e)~Gradient descent updates building positions.
    (f)~Optimized layout concentrates speeds within the comfort band.
  }
  \label{fig:inverse_opt_pipeline}
\end{figure}

\section{Inverse Optimization of Building Layouts}
\label{sec:inverse_opt}

Encouraged by the surrogate's accuracy and its ability to produce full velocity fields in under a second, we investigate whether it can serve as a differentiable physics simulator for inverse optimization of urban building layouts. Given an existing urban layout, we optimize building positions to minimize wind comfort violations in a target region, replacing high-fidelity CFD with the frozen surrogate during gradient computation. The coordinates of building footprint centers pass through a differentiable rasterizer into the frozen surrogate, which predicts the wind field for the current building layout. 

A composite loss then penalizes wind speeds outside the desired comfort range (Fig.~\ref{fig:inverse_opt_pipeline}). The loss takes the form
\begin{equation}
  \mathcal{L} = \lambda_\mathrm{d}\,e_\mathrm{danger} + \lambda_\mathrm{c}\,e_\mathrm{comfort} + \lambda_\mathrm{s}\,e_\mathrm{stag},
\end{equation}
where $e_\mathrm{danger}$, $e_\mathrm{comfort}$, and $e_\mathrm{stag}$ measure the fraction of wind speeds above 15\,m/s, above 5\,m/s, and below 1\,m/s, respectively, in the objective region. The danger term receives ten times the weight of the others ($\lambda_\mathrm{d}=10$, $\lambda_\mathrm{c}=\lambda_\mathrm{s}=1$).

We explore two design modes. In \emph{rigid} mode, each building translates as a unit with a fixed footprint geometry. However, relocating an entire building may not always be necessary. Often, small modifications to a building's footprint geometry can satisfy wind comfort requirements. To emulate this, \emph{morph} mode subdivides each building into $2\!\times\!2$ sub-blocks that move independently, subject to a cohesion loss that prevents them from drifting apart and breaking the building into disconnected fragments. Optimization runs for 200 Adam~\cite{kingma2015adam} steps in both modes. Full details of the optimization, differentiable rasterizer, and regularization terms are provided in the appendix~\ref{app:inverse_opt}.

\begin{table}[h]
\centering
\caption{Pedestrian-level wind speed distribution (\%) before and after single-inlet layout optimization in rigid and morph mode.}
\label{tab:wind_zones}
\begin{tabular}{lrrrr}
\toprule
 & \quad$<1$\,m/s & \quad$1$--$5$\,m/s & \quad$5$--$15$\,m/s & \quad$>15$\,m/s \\
\midrule
Initial &  6.8 & 43.6 & 47.1 & 2.6 \\
Rigid   & 23.7 & 63.5 & 12.6 & 0.2 \\
Morph   & 19.3 & 62.0 & 18.3 & 0.4 \\
\bottomrule
\end{tabular}
\end{table}

\begin{figure}[ht]
  \centering
  \begin{subfigure}[b]{0.30\textwidth}
    \centering
    \includegraphics[width=\textwidth]{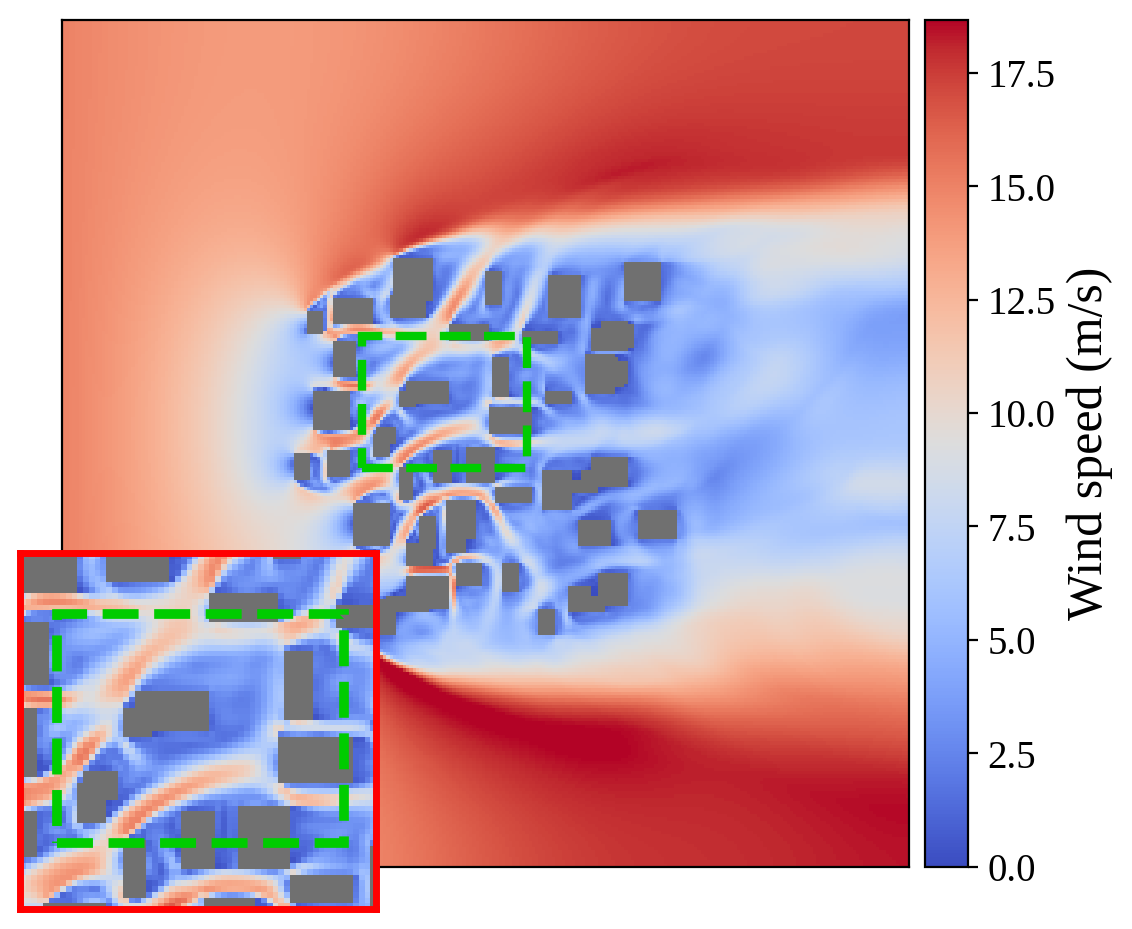}
    \captionsetup{margin={0pt,10pt}}
    \caption{Initial layout}
    \label{fig:rigid_initial}
  \end{subfigure}
  \hfill
  \begin{subfigure}[b]{0.30\textwidth}
    \centering
    \includegraphics[width=\textwidth]{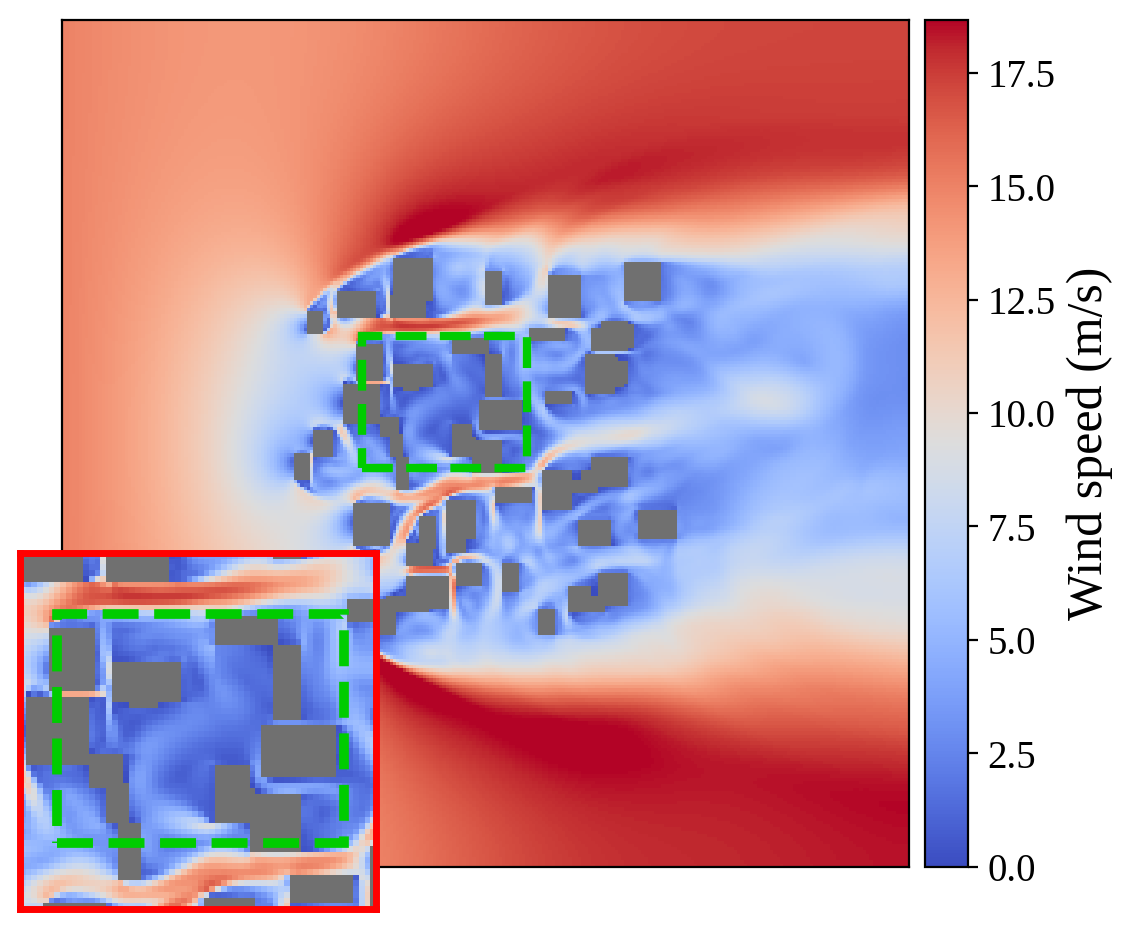}
    \captionsetup{margin={0pt,10pt}}
    \caption{Optimized layout}
    \label{fig:rigid_final}
  \end{subfigure}
  \hfill
  \begin{subfigure}[b]{0.355\textwidth}
    \centering
    \includegraphics[width=\textwidth]{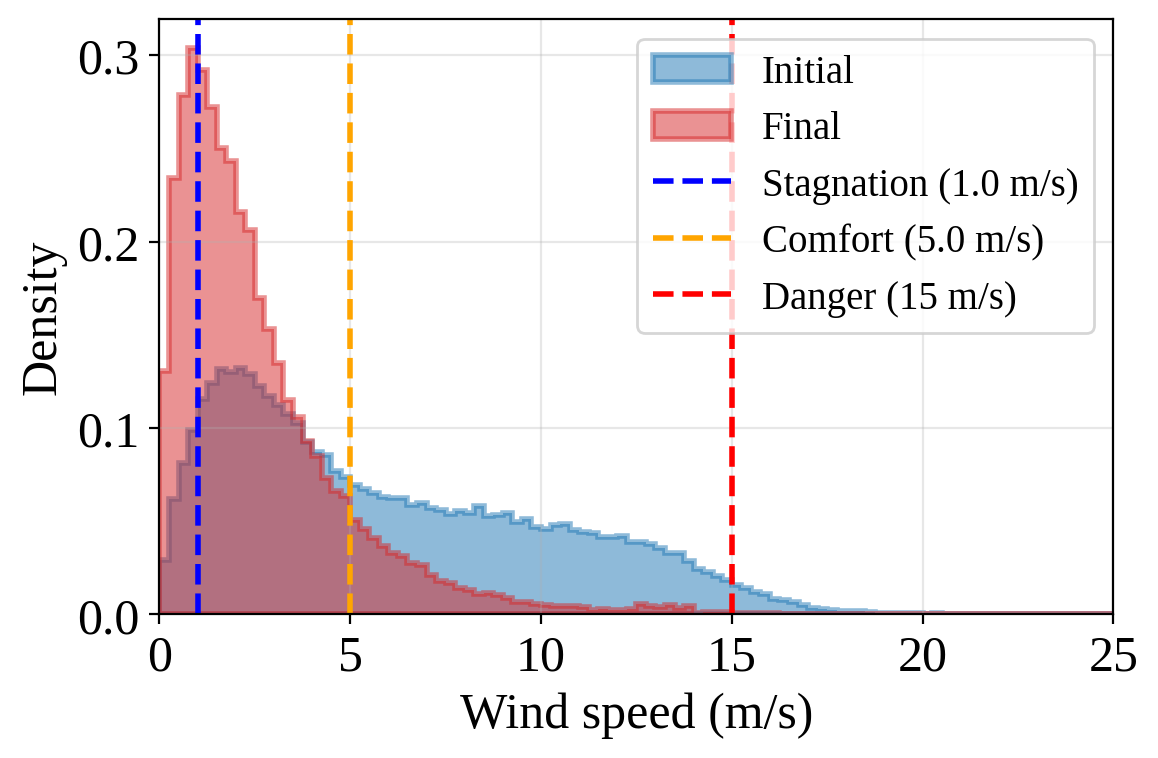}
    \caption{Speed distribution}
    \label{fig:rigid_dist}
  \end{subfigure}
  \caption{%
    Single-inlet \textit{rigid} optimization (15\,m/s, left to right). Buildings translate to shelter the objective region (green rectangle), targeting wind speeds within the 1--5\,m/s comfort band.
  }
  \label{fig:inverse_opt_rigid}
\end{figure}

Figure~\ref{fig:inverse_opt_rigid} shows results for a single inlet boundary condition (left-to-right wind at 15\,m/s) in \textit{rigid} mode.
The buildings translate to form a windbreak upstream of the objective region, deflecting the incoming flow and reducing through flow.
Table~\ref{tab:wind_zones} quantifies the improvement: dangerous wind speeds above 15\,m/s drop from 2.6\% to 0.2\%, and the fraction above 5\,m/s falls from 49.7\% to 12.8\%.

\begin{figure*}[ht]
  \centering
  \begin{subfigure}[b]{0.30\textwidth}
    \centering
    \includegraphics[width=\textwidth]{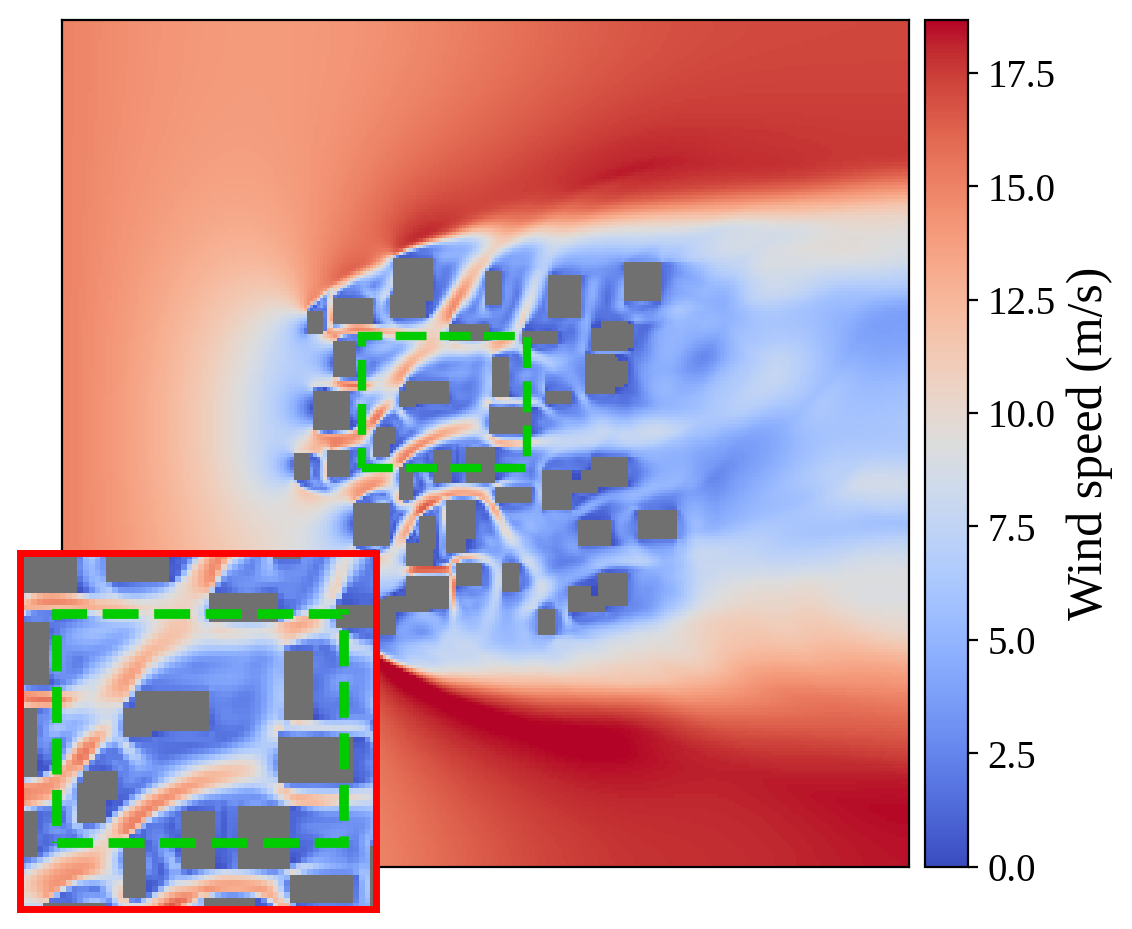}
    \captionsetup{margin={0pt,10pt}}
    \caption{Initial layout}
    \label{fig:morph_initial}
  \end{subfigure}
  \hfill
  \begin{subfigure}[b]{0.30\textwidth}
    \centering
    \includegraphics[width=\textwidth]{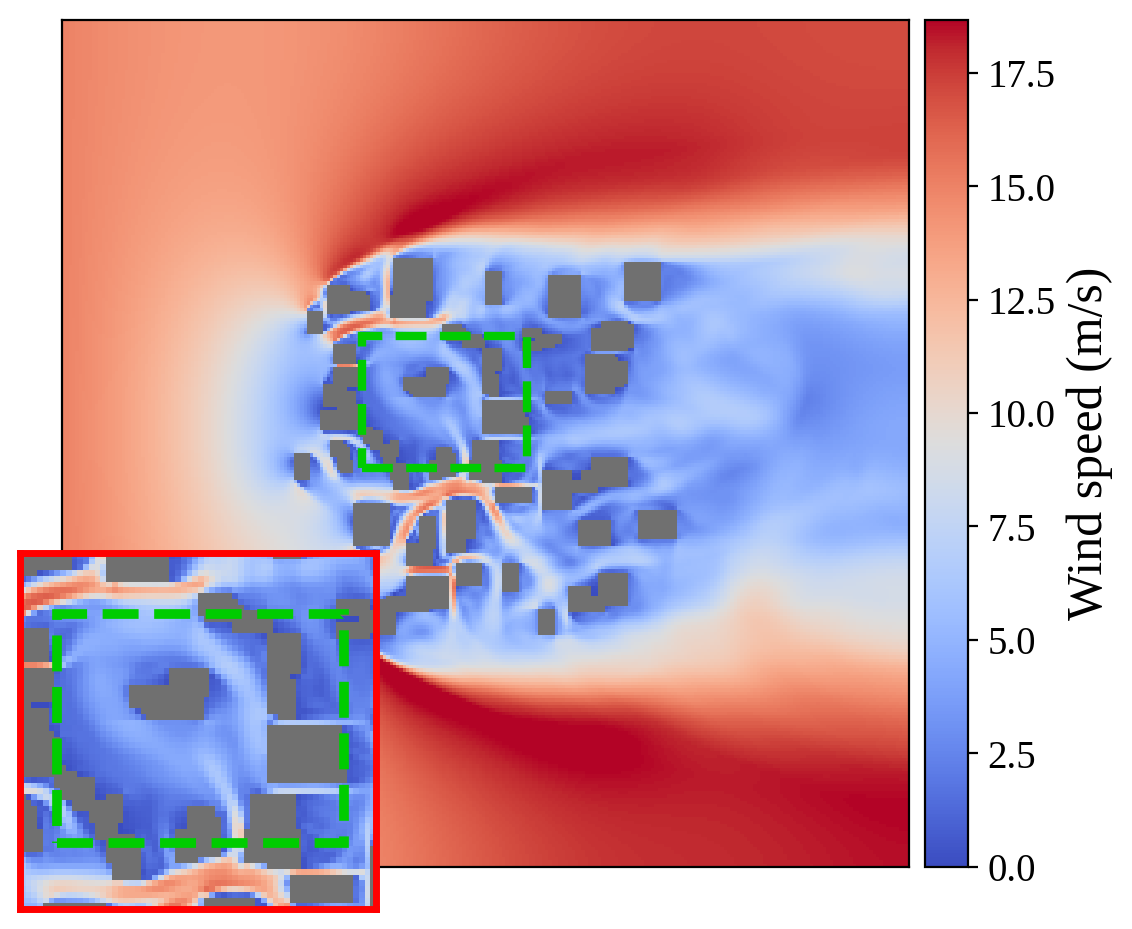}
    \captionsetup{margin={0pt,10pt}}
    \caption{Optimized layout}
    \label{fig:morph_final}
  \end{subfigure}
  \hfill
  \begin{subfigure}[b]{0.355\textwidth}
    \centering
    \includegraphics[width=\textwidth]{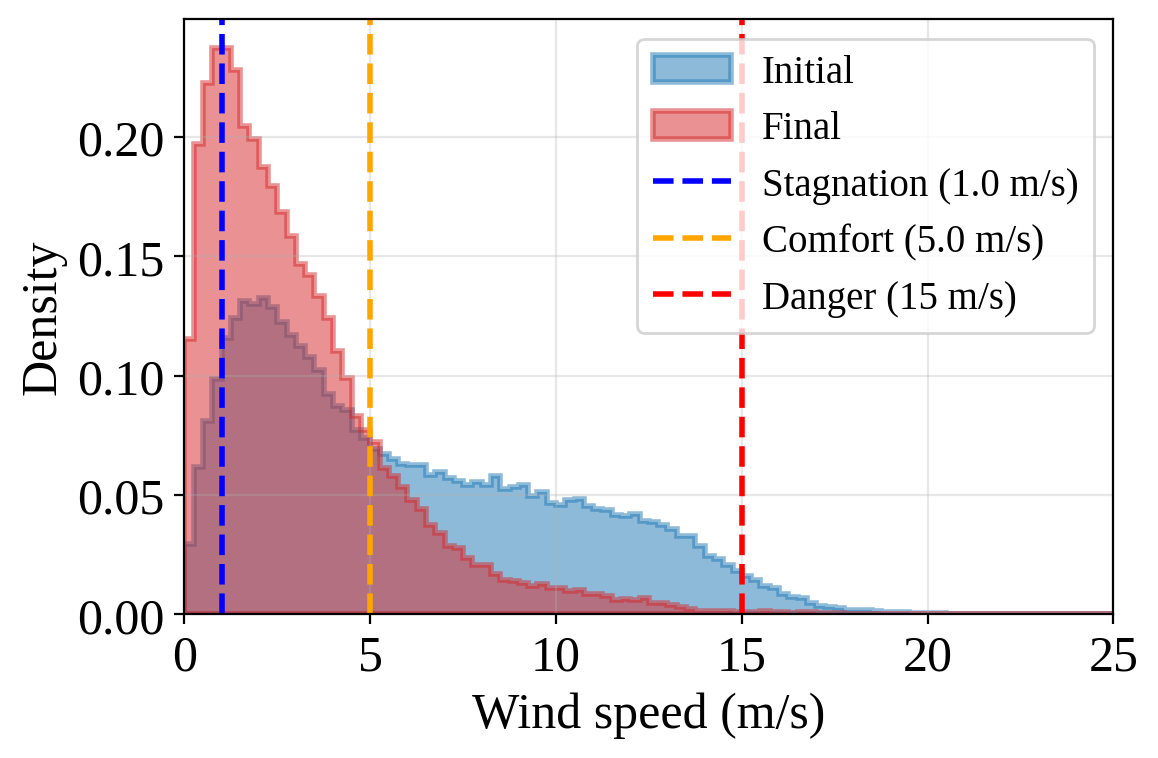}
    \caption{Speed distribution}
    \label{fig:morph_dist}
  \end{subfigure}
  \caption{%
    Single-inlet \textit{morph} optimization (15\,m/s, left to right). Buildings are subdivided into independently movable sub-blocks that can deform the overall shape, targeting the 1--5\,m/s band.
  }
  \label{fig:inverse_opt_morph}
\end{figure*}

The \emph{morph} mode exploits its additional degrees of freedom to reshape building geometry rather than simply translating it (Fig.~\ref{fig:inverse_opt_morph}). The resulting layout produces a different trade-off than \textit{rigid} mode. The fraction of speeds above 5\,m/s is somewhat higher (18.7\% vs.\ 12.8\%), but the stagnation fraction is substantially lower (19.3\% vs.\ 23.7\%). The \textit{morph} optimizer appears to leave controlled openings in the shelter that maintain airflow through the objective region, channeling wind into a circular motion rather than blocking it entirely.

Both modes trade dangerous and uncomfortable winds for stagnation, as the density plots in Figs.~\ref{fig:rigid_dist} and~\ref{fig:morph_dist} show. Most of the probability mass sits just above the 1\,m/s threshold. This is inherent to the composite loss, since reducing high speeds inevitably concentrates mass at the low end. If stagnation is a primary concern for a given site, designers can increase $\lambda_\mathrm{s}$ to penalize it more aggressively.

\begin{figure}[t]
  \centering
  \begin{subfigure}[b]{0.30\textwidth}
    \centering
    \includegraphics[width=\textwidth]{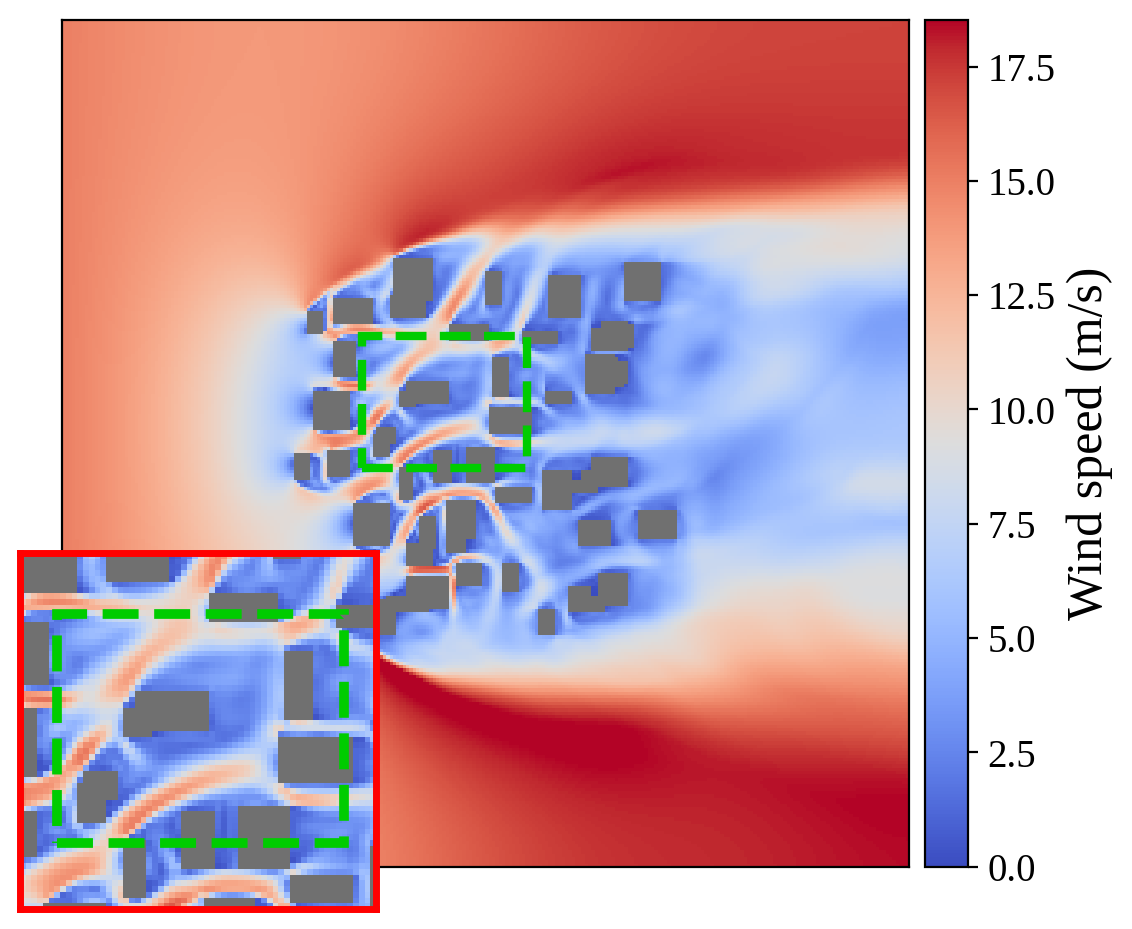}
    \captionsetup{margin={0pt,10pt}}
    \caption{Left inlet: initial}
  \end{subfigure}\hfill
  \begin{subfigure}[b]{0.30\textwidth}
    \centering
    \includegraphics[width=\textwidth]{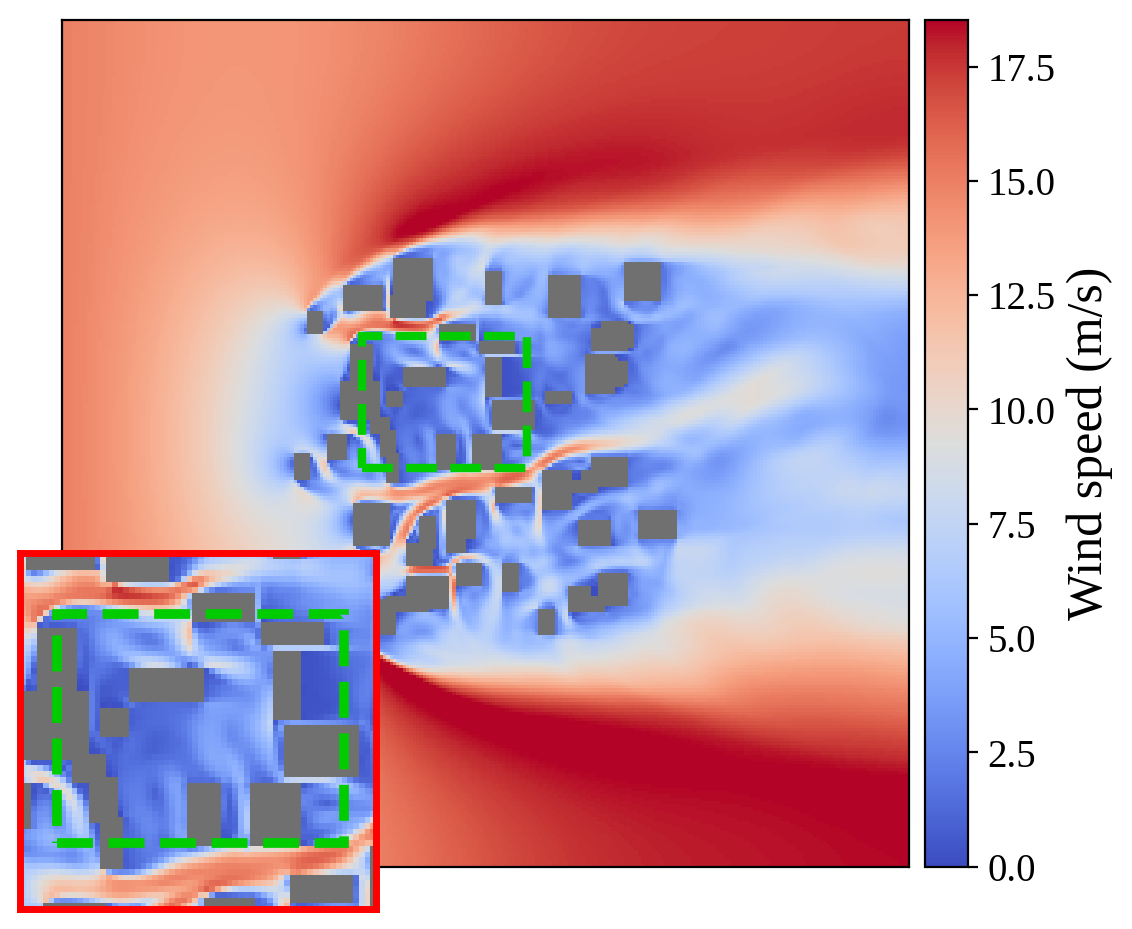}
    \captionsetup{margin={0pt,10pt}}
    \caption{Left inlet: optimized}
  \end{subfigure}\hfill
  \begin{subfigure}[b]{0.355\textwidth}
    \centering
    \includegraphics[width=\textwidth]{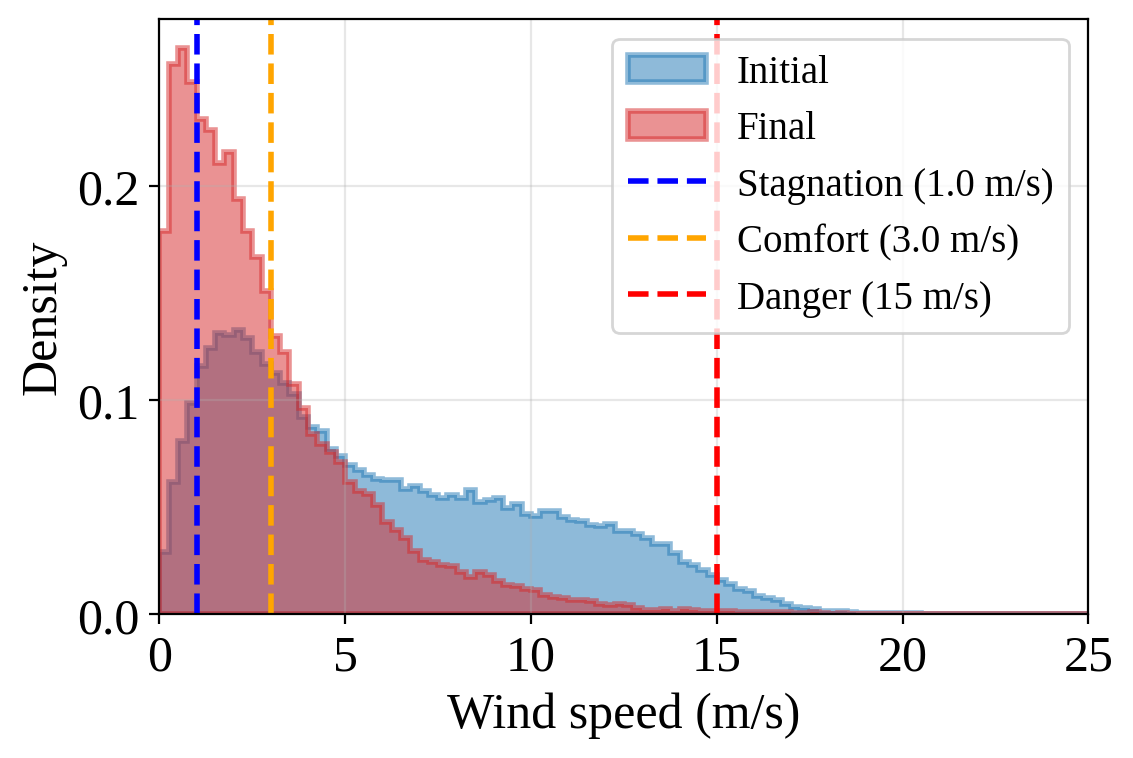}
    \caption{Left inlet: distribution}
    \label{fig:multi_inlet_rigid_left_density}
  \end{subfigure}
  \\[0.4em]
  \begin{subfigure}[b]{0.30\textwidth}
    \centering
    \includegraphics[width=\textwidth]{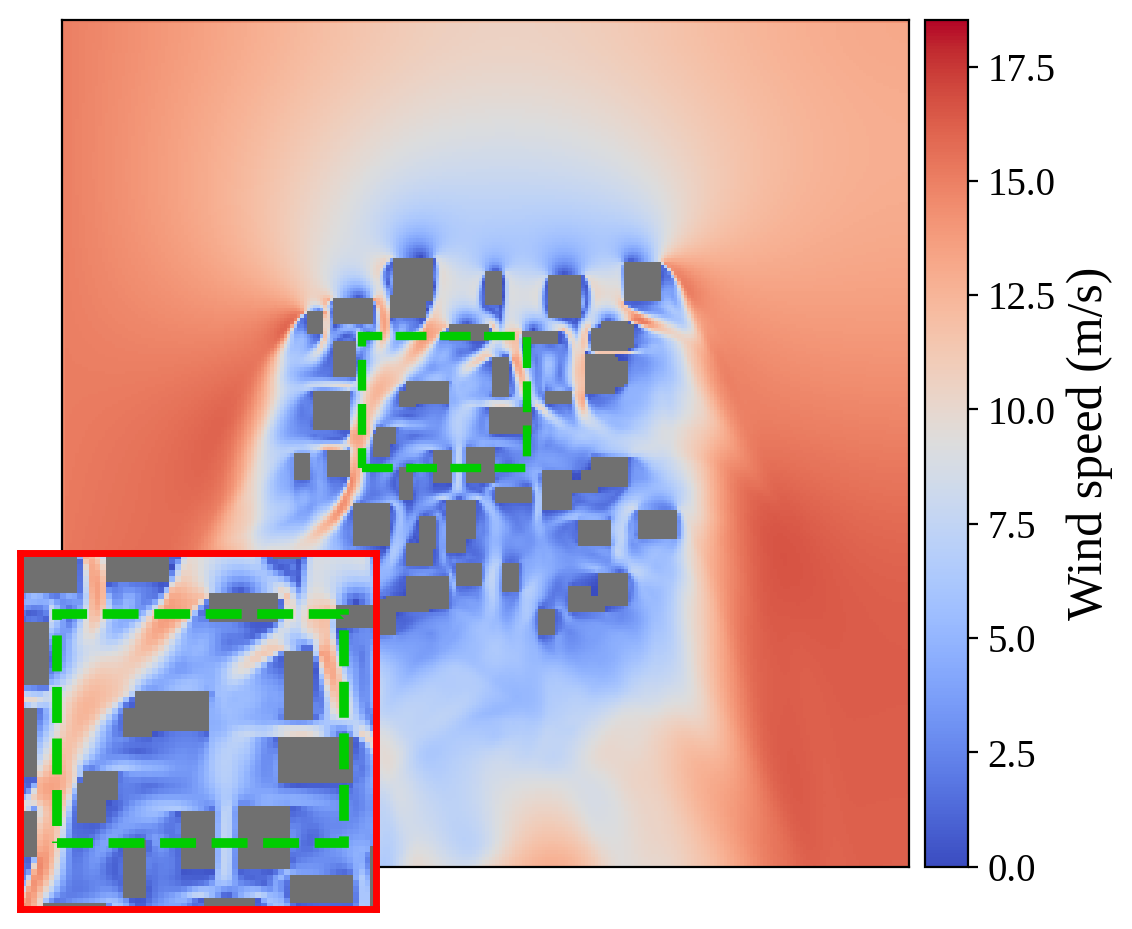}
    \captionsetup{margin={0pt,10pt}}
    \caption{Top inlet: initial}
  \end{subfigure}\hfill
  \begin{subfigure}[b]{0.30\textwidth}
    \centering
    \includegraphics[width=\textwidth]{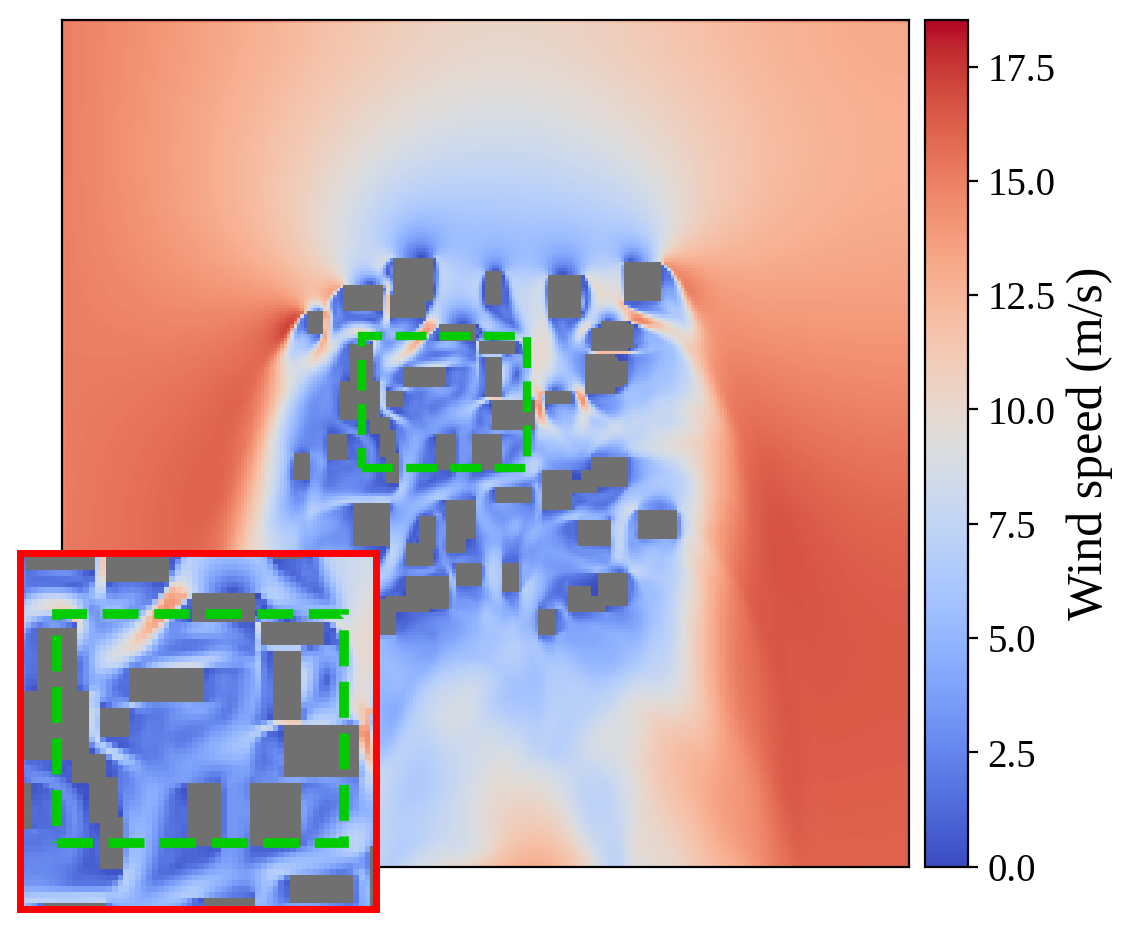}
    \captionsetup{margin={0pt,10pt}}
    \caption{Top inlet: optimized}
  \end{subfigure}\hfill
  \begin{subfigure}[b]{0.355\textwidth}
    \centering
    \includegraphics[width=\textwidth]{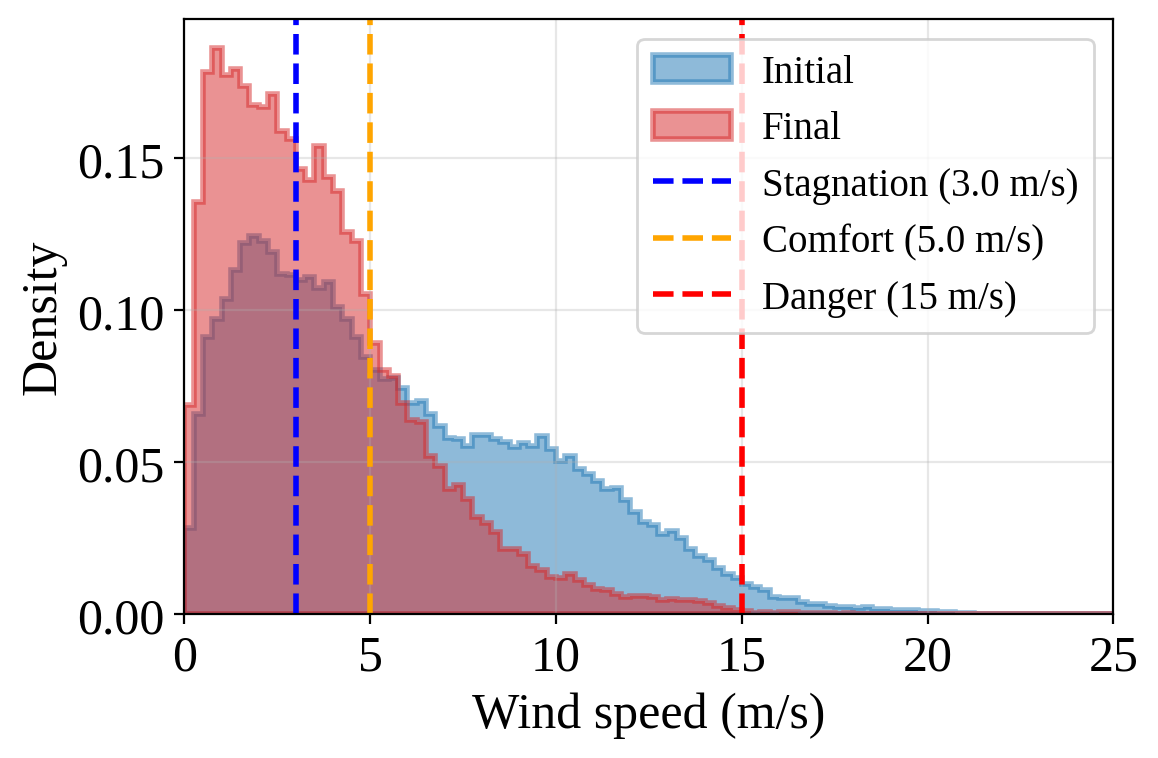}
    \caption{Top inlet: distribution}
    \label{fig:multi_inlet_rigid_top_density}
  \end{subfigure}
  \caption{%
    Multi-inlet \textit{rigid} optimization with left inlet $(15,0)$\,m/s (target comfort band 1--3\,m/s) and top inlet $(0,15)$\,m/s (target comfort band 3--5\,m/s). A single layout is co-optimized for both inlets and evaluated under each direction separately.
  }
  \label{fig:multi_inlet_rigid}
\end{figure}

Urban areas often experience wind from various directions throughout the year, and comfort requirements may differ between seasons and climate zones. In winter, shelter from cold wind is desirable (low target speeds), whereas in summer, some airflow is desired for cooling (moderate target speeds), provided it does not introduce dangerous gusts. Figure~\ref{fig:multi_inlet_rigid} demonstrates a climate-adaptive design scenario with two conflicting wind directions, a left inlet at 15\,m/s targeting a comfort band of 1--3\,m/s and a top inlet at 15\,m/s targeting 3--5\,m/s. A single layout must satisfy both objectives simultaneously. The wind speed distribution for left inlet (Fig.~\ref{fig:multi_inlet_rigid_left_density}) concentrates mass toward the 1--3\,m/s band, while the distribution for top inlet (Fig.~\ref{fig:multi_inlet_rigid_top_density}) shifts mass into the 3--5\,m/s range. The corresponding \textit{morph} and \textit{rigid} results are shown in the appendix (Fig.~\ref{fig:multi_inlet_results}).

\section{Conclusion}

This work demonstrates that a video diffusion model pretrained on natural video can be fine-tuned into a quantitatively accurate surrogate for urban wind simulation, and that it can outperform neural operators designed specifically for PDE solving. The resulting model generates 112-frame rollouts in under a second, roughly three orders of magnitude faster than the CFD solver. The combination of speed and inherent differentiability enables a design workflow that is difficult to achieve with conventional CFD: candidate building layouts can be assessed in real time, and gradients obtained by backpropagation through the predicted flow field can be used to adjust geometry toward improved wind comfort. Our inverse optimization experiments show that this works for single and multiple inlet directions, with both \textit{rigid} translation and continuous \textit{morph}ing of buildings. The current sub-block parametrization could easily be extended to richer representations, such as per-face offsets or spline-based footprints, which would allow for realistic shape exploration. Furthermore, the objective itself could incorporate additional differentiable constraints, such as minimum passage widths or pedestrian-flow norms. A complementary direction is to pair our gradient-based optimizer with generative inverse-design methods that impose a learned prior over candidate solutions, such as CinDM~\cite{wu2024compositional} and DiffPhyCon~\cite{wei2024diffphycon}, which could further improve the discovered layouts.

The current formulation operates in 2D. An extension to 3D is relatively straightforward: instead of binary occupancy masks, building heights can be encoded as continuous values in the conditioning frame. The pretrained video models have substantial unused capacity, as shown by the fact that only two denoising steps are enough for our current flow fields. We expect the bottleneck to be the data, because generating 3D urban CFD at comparable diversity and resolution would require substantially more computational resources.

We selected LTX-Video for its open weights, active community, and focus on efficient generation. We expect our findings to transfer to other non-autoregressive diffusion models such as Wan~2.2~\cite{Wan22}. An interesting experiment would be to test autoregressive video models such as MAGI-1~\cite{sand2025magi}, in particular, whether error accumulation across sequential generation steps would pose problems for inverse optimization, similar to what we observe for OFormer in our surrogate comparison (cf.\ appendix, Table~\ref{tab:surrogate_comparison}).

\begin{ack}
We thank Cornelia Kalender for guidance on evaluation metrics, Kai Marti for the real-world building footprints, and Spencer Folk for access to the fluid-solver codebase.
This work was supported as part of the Swiss AI Initiative by a grant from the Swiss National Supercomputing Centre (CSCS) under project ID a0120 on Alps.
\end{ack}

\newpage

\bibliographystyle{plainnat}
\bibliography{main}

@article{loshchilov2019adamw,
  title={Decoupled weight decay regularization},
  author={Loshchilov, Ilya and Hutter, Frank},
  journal={arXiv preprint arXiv:1711.05101},
  year={2017}
}

@article{kingma2015adam,
  title={Adam: A method for stochastic optimization},
  author={Kingma, Diederik P and Ba, Jimmy},
  journal={arXiv preprint arXiv:1412.6980},
  year={2014}
}

@article{lu2025machine,
  title={Machine learning predicts pedestrian wind flow from urban morphology and prevailing wind direction},
  author={Lu, Jiachen and Li, Wei and Hobeichi, Sanaa and Azad, Shakir Aymam and Nazarian, Negin},
  journal={Environmental Research Letters},
  volume={20},
  number={5},
  pages={054006},
  year={2025},
  publisher={IOP Publishing}
}

@article{snaiki2025hierarchical,
  title={A hierarchical deep learning model for predicting pedestrian-level urban winds},
  author={Snaiki, Reda and Lu, Jiachen and Li, Shaopeng and Nazarian, Negin},
  journal={Building and Environment},
  pages={114354},
  year={2026},
  publisher={Elsevier}
}

@inproceedings{mokhtar2021pedestrian,
  title={Pedestrian wind factor estimation in complex urban environments},
  author={Mokhtar, Sarah and Beveridge, Matt and Cao, Yumeng and Drori, Iddo},
  booktitle={Asian conference on machine learning},
  pages={486--501},
  year={2021},
  organization={PMLR}
}

@article{li2021fourier,
  title={Fourier neural operator for parametric partial differential equations},
  author={Li, Zongyi and Kovachki, Nikola and Azizzadenesheli, Kamyar and Liu, Burigede and Bhattacharya, Kaushik and Stuart, Andrew and Anandkumar, Anima},
  journal={arXiv preprint arXiv:2010.08895},
  year={2020}
}

@article{folk2024learning,
  author={Folk, Spencer and Melton, John and Margolis, Benjamin W. L. and Yim, Mark and Kumar, Vijay},
  journal={IEEE Robotics and Automation Letters},
  title={Learning Local Urban Wind Flow Fields From Range Sensing},
  year={2024},
  volume={9},
  number={9},
  pages={7413-7420},
  doi={10.1109/LRA.2024.3426209}
}

@article{lawson1978wind,
  title={The wind content of the built environment},
  author={Lawson, T. V.},
  journal={Journal of Wind Engineering and Industrial Aerodynamics},
  volume={3},
  number={2--3},
  pages={93--105},
  year={1978}
}

@inproceedings{isyumov1975ground,
  title={The ground level wind environment in built-up areas},
  author={Isyumov, N},
  booktitle={Proc. of the 4th Int. Conf. on Wind Effects on Buildings and Structures, London, 1975},
  year={1975}
}

@article{blattmann2023stable,
  title={Stable video diffusion: Scaling latent video diffusion models to large datasets},
  author={Blattmann, Andreas and Dockhorn, Tim and Kulal, Sumith and Mendelevitch, Daniel and Kilian, Maciej and Lorenz, Dominik and Levi, Yam and English, Zion and Voleti, Vikram and Letts, Adam and others},
  journal={arXiv preprint arXiv:2311.15127},
  year={2023}
}

@article{gillman2025force,
  title={Force prompting: Video generation models can learn and generalize physics-based control signals},
  author={Gillman, Nate and Herrmann, Charles and Freeman, Michael and Aggarwal, Daksh and Luo, Evan and Sun, Deqing and Sun, Chen},
  journal={arXiv preprint arXiv:2505.19386},
  year={2025}
}

@article{zhang2025think,
  title={Think Before You Diffuse: Infusing Physical Rules into Video Diffusion},
  author={Zhang, Ke and Xiao, Cihan and Xu, Jiacong and Mei, Yiqun and Patel, Vishal M},
  journal={arXiv preprint arXiv:2505.21653},
  year={2025}
}

@article{wiedemer2025videoreasoning,
  title={Video models are zero-shot learners and reasoners},
  author={Wiedemer, Thadd{\"a}us and Li, Yuxuan and Vicol, Paul and Gu, Shixiang Shane and Matarese, Nick and Swersky, Kevin and Kim, Been and Jaini, Priyank and Geirhos, Robert},
  journal={arXiv preprint arXiv:2509.20328},
  year={2025}
}

@article{Wan22,
  title={Wan: Open and advanced large-scale video generative models},
  author={Wan, Team and Wang, Ang and Ai, Baole and Wen, Bin and Mao, Chaojie and Xie, Chen-Wei and Chen, Di and Yu, Feiwu and Zhao, Haiming and Yang, Jianxiao and others},
  journal={arXiv preprint arXiv:2503.20314},
  year={2025}
}

@article{du2024confil,
  title={Conditional neural field latent diffusion model for generating spatiotemporal turbulence},
  author={Du, Pan and Parikh, Meet Hemant and Fan, Xiantao and Liu, Xin-Yang and Wang, Jian-Xun},
  journal={Nature Communications},
  volume={15},
  number={1},
  pages={10416},
  year={2024},
  publisher={Nature Publishing Group UK London}
}

@inproceedings{saito2017tgan,
  title={Temporal generative adversarial nets with singular value clipping},
  author={Saito, Masaki and Matsumoto, Eiichi and Saito, Shunta},
  booktitle={Proceedings of the IEEE international conference on computer vision},
  pages={2830--2839},
  year={2017}
}

@article{vondrick2016generating,
  title={Generating videos with scene dynamics},
  author={Vondrick, Carl and Pirsiavash, Hamed and Torralba, Antonio},
  journal={Advances in neural information processing systems},
  volume={29},
  year={2016}
}

@inproceedings{tulyakov2018mocogan,
  title={Mocogan: Decomposing motion and content for video generation},
  author={Tulyakov, Sergey and Liu, Ming-Yu and Yang, Xiaodong and Kautz, Jan},
  booktitle={Proceedings of the IEEE conference on computer vision and pattern recognition},
  pages={1526--1535},
  year={2018}
}

@article{ho2022videodiffusion,
  title={Video diffusion models},
  author={Ho, Jonathan and Salimans, Tim and Gritsenko, Alexey and Chan, William and Norouzi, Mohammad and Fleet, David J},
  journal={Advances in neural information processing systems},
  volume={35},
  pages={8633--8646},
  year={2022}
}

@article{hacohen2024ltx,
  title={Ltx-video: Realtime video latent diffusion},
  author={HaCohen, Yoav and Chiprut, Nisan and Brazowski, Benny and Shalem, Daniel and Moshe, Dudu and Richardson, Eitan and Levin, Eran and Shiran, Guy and Zabari, Nir and Gordon, Ori and others},
  journal={arXiv preprint arXiv:2501.00103},
  year={2024}
}

@article{valevski2024gamengen,
  title={Diffusion models are real-time game engines},
  author={Valevski, Dani and Leviathan, Yaniv and Arar, Moab and Fruchter, Shlomi},
  journal={arXiv preprint arXiv:2408.14837},
  year={2024}
}

@article{wang2025physctrl,
  title={Physctrl: Generative physics for controllable and physics-grounded video generation},
  author={Wang, Chen and Chen, Chuhao and Huang, Yiming and Dou, Zhiyang and Liu, Yuan and Gu, Jiatao and Liu, Lingjie},
  journal={arXiv preprint arXiv:2509.20358},
  year={2025}
}

@article{herde2024poseidon,
  title={Poseidon: Efficient foundation models for pdes},
  author={Herde, Maximilian and Raoni{\'c}, Bogdan and Rohner, Tobias and K{\"a}ppeli, Roger and Molinaro, Roberto and De Bezenac, Emmanuel and Mishra, Siddhartha},
  journal={Advances in Neural Information Processing Systems},
  volume={37},
  pages={72525--72624},
  year={2024}
}

@article{nguyen2025physix,
  title={Physix: A foundation model for physics simulations},
  author={Nguyen, Tung and Koneru, Arsh and Li, Shufan and Grover, Aditya},
  journal={arXiv preprint arXiv:2506.17774},
  year={2025}
}

@article{takamoto2022pdebench,
  title={Pdebench: An extensive benchmark for scientific machine learning},
  author={Takamoto, Makoto and Praditia, Timothy and Leiteritz, Raphael and MacKinlay, Daniel and Alesiani, Francesco and Pfl{\"u}ger, Dirk and Niepert, Mathias},
  journal={Advances in neural information processing systems},
  volume={35},
  pages={1596--1611},
  year={2022}
}

@article{ohana2024well,
  title={The well: a large-scale collection of diverse physics simulations for machine learning},
  author={Ohana, Ruben and McCabe, Michael and Meyer, Lucas and Morel, Rudy and Agocs, Fruzsina J and Beneitez, Miguel and Berger, Marsha and Burkhart, Blakesley and Dalziel, Stuart B and Fielding, Drummond B and others},
  journal={Advances in Neural Information Processing Systems},
  volume={37},
  pages={44989--45037},
  year={2024}
}

@article{li2025videopde,
  title={Videopde: Unified generative pde solving via video inpainting diffusion models},
  author={Li, Edward and Wang, Zichen and Huang, Jiahe and Park, Jeong Joon},
  journal={arXiv preprint arXiv:2506.13754},
  year={2025}
}

@article{liu2023sgmsgnn,
  title={Accurate and Efficient Urban Wind Prediction at City-Scale with Memory-Scalable Graph Neural Network},
  author={Liu, Zhijian and Zhang, Siqi and Shao, Xuqiang and Wu, Zhaohui},
  journal={Sustainable Cities and Society},
  year={2023}
}

@article{shao2023pignn,
  title={PIGNN-CFD: A physics-informed graph neural network for rapid predicting urban wind field defined on unstructured mesh},
  author={Shao, Xuqiang and Liu, Zhijian and Zhang, Siqi and Zhao, Zijia and Hu, Chenxing},
  journal={Building and Environment},
  volume={232},
  pages={110056},
  year={2023},
  publisher={Elsevier}
}

@article{qin2024localfno,
  title={Modeling multivariable high-resolution 3D urban microclimate using localized Fourier neural operator},
  author={Qin, Shaoxiang and Zhan, Dongxue and Geng, Dingyang and Peng, Wenhui and Tian, Geng and Shi, Yurong and Gao, Naiping and Liu, Xue and Wang, Liangzhu Leon},
  journal={Building and Environment},
  volume={273},
  pages={112668},
  year={2025},
  publisher={Elsevier}
}

@article{chen2025fnourban,
  title={Generalization of urban wind environment using fourier neural operator across different wind directions and cities},
  author={Chen, Cheng and Tian, Geng and Qin, Shaoxiang and Yang, Senwen and Geng, Dingyang and Zhan, Dongxue and Yang, Jinqiu and Vidal, David and Wang, Liangzhu Leon},
  journal={arXiv preprint arXiv:2501.05499},
  year={2025}
}

@article{giral2025graphdiffusion,
  title={Generative Urban Flow Modeling: From Geometry to Airflow with Graph Diffusion},
  author={Giral, Francisco and Manzano, {\'A}lvaro and G{\'o}mez, Ignacio and Koumoutsakos, Petros and Clainche, Soledad Le},
  journal={arXiv preprint arXiv:2512.14725},
  year={2025}
}

@article{soares2025pdefm,
  title={Towards a Foundation Model for Partial Differential Equations Across Physics Domains},
  author={Soares, Eduardo and Brazil, Emilio Vital and Shirasuna, Victor and de Carvalho, Breno WSR and Malossi, Cristiano},
  journal={arXiv preprint arXiv:2511.21861},
  year={2025}
}

@article{wiesner2025gphyt,
  title={Towards a physics foundation model},
  author={Wiesner, Florian and Wessling, Matthias and Baek, Stephen},
  journal={arXiv preprint arXiv:2509.13805},
  year={2025}
}

@ARTICLE{satish2026physvideogen,
       author = {{Nilol Kundur Satish}, Siddarth and {Jaiswal}, Devesh and {Chen}, Hongyu and {Bakshi}, Abhishek},
        title = "{PhysVideoGenerator: Towards Physically Aware Video Generation via Latent Physics Guidance}",
      journal = {arXiv e-prints},
         year = 2026,
        month = jan,
          eid = {arXiv:2601.03665},
        pages = {arXiv:2601.03665},
          doi = {10.48550/arXiv.2601.03665},
archivePrefix = {arXiv},
       eprint = {2601.03665},
 primaryClass = {cs.CV},
       adsurl = {https://ui.adsabs.harvard.edu/abs/2026arXiv260103665N}
}

@inproceedings{guibas2022efficient,
  title={Efficient token mixing for transformers via adaptive fourier neural operators},
  author={Guibas, John and Mardani, Morteza and Li, Zongyi and Tao, Andrew and Anandkumar, Anima and Catanzaro, Bryan},
  booktitle={International conference on learning representations},
  year={2021}
}

@article{li2023transformer,
  title={Transformer for partial differential equations' operator learning},
  author={Li, Zijie and Meidani, Kazem and Farimani, Amir Barati},
  journal={arXiv preprint arXiv:2205.13671},
  year={2022}
}

@article{liuschiaffini2023tipping,
  title={Tipping point forecasting in non-stationary dynamics on function spaces},
  author={Liu-Schiaffini, Miguel and Singer, Clare E and Kovachki, Nikola and Schneider, Tapio and Azizzadenesheli, Kamyar and Anandkumar, Anima},
  journal={arXiv preprint arXiv:2308.08794},
  year={2023}
}

@misc{physicsnemo,
  title={{NVIDIA PhysicsNeMo}: An Open-Source Framework for Physics-{ML} Model Building and Training},
  author={{NVIDIA Corporation}},
  year={2025},
  url={https://github.com/NVIDIA/physicsnemo}
}

@inproceedings{perez2018film,
  title={Film: Visual reasoning with a general conditioning layer},
  author={Perez, Ethan and Strub, Florian and De Vries, Harm and Dumoulin, Vincent and Courville, Aaron},
  booktitle={Proceedings of the AAAI conference on artificial intelligence},
  volume={32},
  year={2018}
}

@inproceedings{ronneberger2015unet,
  title={U-net: Convolutional networks for biomedical image segmentation},
  author={Ronneberger, Olaf and Fischer, Philipp and Brox, Thomas},
  booktitle={International Conference on Medical image computing and computer-assisted intervention},
  pages={234--241},
  year={2015},
  organization={Springer}
}

@article{tancik2020fourier,
  title={Fourier features let networks learn high frequency functions in low dimensional domains},
  author={Tancik, Matthew and Srinivasan, Pratul and Mildenhall, Ben and Fridovich-Keil, Sara and Raghavan, Nithin and Singhal, Utkarsh and Ramamoorthi, Ravi and Barron, Jonathan and Ng, Ren},
  journal={Advances in neural information processing systems},
  volume={33},
  pages={7537--7547},
  year={2020}
}

@article{kastner2023gan,
  title={A GAN-based surrogate model for instantaneous urban wind flow prediction},
  author={Kastner, Patrick and Dogan, Timur},
  journal={Building and Environment},
  volume={242},
  pages={110384},
  year={2023},
  publisher={Elsevier}
}

@article{clemente2024rapid,
  title={Rapid pedestrian-level wind field prediction for early-stage design using Pareto-optimized convolutional neural networks},
  author={Clemente, Alfredo Vicente and Giljarhus, Knut Erik Teigen and Oggiano, Luca and Ruocco, Massimiliano},
  journal={Computer-Aided Civil and Infrastructure Engineering},
  volume={39},
  number={18},
  pages={2826--2839},
  year={2024},
  publisher={Wiley Online Library}
}

@article{clarke2025mlpmixer,
  title={Deep learning for urban wind prediction: An MLP-Mixer approach with 3D encoding},
  author={Clarke, Adam and Giljarhus, Knut Erik Teigen and Oggiano, Luca and Saddington, Alistair and Depuru-Mohan, Karthik},
  journal={Building and Environment},
  pages={113495},
  year={2025},
  publisher={Elsevier}
}

@article{rui2023pinn,
  title={Reconstruction of 3D flow field around a building model in wind tunnel: a novel physics-informed neural network framework adopting dynamic prioritization self-adaptive loss balance strategy},
  author={Rui, En-Ze and Chen, Zheng-Wei and Ni, Yi-Qing and Yuan, Lei and Zeng, Guang-Zhi},
  journal={Engineering Applications of Computational Fluid Mechanics},
  volume={17},
  number={1},
  pages={2238849},
  year={2023},
  publisher={Taylor \& Francis}
}

@article{wu2024surrogatereview,
  title={A review of surrogate-assisted design optimization for improving urban wind environment},
  author={Wu, Yihan and Quan, Steven Jige},
  journal={Building and Environment},
  volume={253},
  pages={111157},
  year={2024},
  publisher={Elsevier}
}

@article{kaseb2022cfdbased,
  title={Towards CFD-based optimization of urban wind conditions: Comparison of Genetic algorithm, Particle Swarm Optimization, and a hybrid algorithm},
  author={Kaseb, Zeynab and Rahbar, Morteza},
  journal={Sustainable Cities and Society},
  volume={77},
  pages={103565},
  year={2022},
  publisher={Elsevier}
}

@article{huang2022accelerated,
  title={Accelerated environmental performance-driven urban design with generative adversarial network},
  author={Huang, Chenyu and Zhang, Gengjia and Yao, Jiawei and Wang, Xiaoxin and Calautit, John Kaiser and Zhao, Cairong and An, Na and Peng, Xi},
  journal={Building and Environment},
  volume={224},
  pages={109575},
  year={2022},
  publisher={Elsevier}
}

@inproceedings{liu2019softrasterizer,
  title={Soft rasterizer: A differentiable renderer for image-based 3d reasoning},
  author={Liu, Shichen and Li, Tianye and Chen, Weikai and Li, Hao},
  booktitle={Proceedings of the IEEE/CVF international conference on computer vision},
  pages={7708--7717},
  year={2019}
}

@inproceedings{
    hu2022lora,
    title={Lo{RA}: Low-Rank Adaptation of Large Language Models},
    author={Edward J Hu and Yelong Shen and Phillip Wallis and Zeyuan Allen-Zhu and Yuanzhi Li and Shean Wang and Lu Wang and Weizhu Chen},
    booktitle={International Conference on Learning Representations},
    year={2022},
}

@inproceedings{peebles2023dit,
  title={Scalable diffusion models with transformers},
  author={Peebles, William and Xie, Saining},
  booktitle={Proceedings of the IEEE/CVF international conference on computer vision},
  pages={4195--4205},
  year={2023}
}

@article{sand2025magi,
  title={Magi-1: Autoregressive video generation at scale},
  author={Teng, Hansi and Jia, Hongyu and Sun, Lei and Li, Lingzhi and Li, Maolin and Tang, Mingqiu and Han, Shuai and Zhang, Tianning and Zhang, WQ and Luo, Weifeng and others},
  journal={arXiv preprint arXiv:2505.13211},
  year={2025}
}

@article{gencfd,
  title={Generative ai for fast and accurate statistical computation of fluids},
  author={Molinaro, Roberto and Lanthaler, Samuel and Raoni{\'c}, Bogdan and Rohner, Tobias and Armegioiu, Victor and Simonis, Stephan and Grund, Dana and Ramic, Yannick and Wan, Zhong Yi and Sha, Fei and others},
  journal={arXiv preprint arXiv:2409.18359},
  year={2024}
}

@misc{letzel2008,
  title={High resolution large-eddy simulation of turbulent flow around buildings},
  author={Letzel, Marcus Oliver},
  year={2007},
  publisher={Hannover: Gottfried Wilhelm Leibniz Universit{\"a}t Hannover}
}

@misc{simscalePWC,
  author       = {{SimScale GmbH}},
  title        = {Pedestrian Wind Comfort Analysis -- Online Documentation},
  year         = {2025},
  howpublished = {\url{https://www.simscale.com/docs/analysis-types/pedestrian-wind-comfort-analysis/}, last accessed 2026-06-26},
}

@misc{CoLWind2019,
  author       = {{City of London Corporation}},
  title        = {Wind Microclimate Guidelines for Developments in the City of London},
  year         = {2019},
  howpublished = {\url{https://www.cityoflondon.gov.uk/assets/Services-Environment/wind-microclimate-guidelines.pdf}, last accessed 2026-06-26},
}

@misc{EN1991-1-4,
  author       = {{CEN}},
  title        = {Eurocode 1: Actions on structures -- Part 1-4: General actions -- Wind actions},
  howpublished = {Standard},
  number       = {EN 1991-1-4:2005+A1:2010},
  year         = {2010},
  key          = {EN1991-1-4},
}

@article{MIRZAEI2021102839,
  title={CFD modeling of micro and urban climates: Problems to be solved in the new decade},
  author={Mirzaei, Parham A},
  journal={Sustainable Cities and Society},
  volume={69},
  pages={102839},
  year={2021},
  publisher={Elsevier}
}

@inproceedings{
wu2024compositional,
title={Compositional Generative Inverse Design},
author={Tailin Wu and Takashi Maruyama and Long Wei and Tao Zhang and Yilun Du and Gianluca Iaccarino and Jure Leskovec},
booktitle={The Twelfth International Conference on Learning Representations},
year={2024},
url={https://openreview.net/forum?id=wmX0CqFSd7},
note={last accessed 2026-06-26}
}

@article{wei2024diffphycon,
  title={Diffphycon: a generative approach to control complex physical systems},
  author={Wei, Long and Hu, Peiyan and Feng, Ruiqi and Feng, Haodong and Du, Yixuan and Zhang, Tao and Wang, Rui and Wang, Yue and Ma, Zhi-Ming and Wu, Tailin},
  journal={Advances in Neural Information Processing Systems},
  volume={37},
  pages={4090--4147},
  year={2024}
}

\newpage

\appendix
\section{Dataset Details}
\label{sec:appendix_dataset}

\subsection{Incompressible CFD Setup}

We model the atmospheric boundary-layer flow as an incompressible fluid, which is appropriate for typical urban wind speeds where Mach numbers remain well below compressible regimes~\cite{letzel2008}. The numerical solver advances the incompressible Euler equations on a fixed 2D horizontal domain representing the building canopy layer, neglecting vertical acceleration but resolving horizontal flow separation and channeling between obstacles~\cite{letzel2008}.

\subsection{Boundary Conditions and Reference Wind Speed}

At the inlet we prescribe a uniform horizontal wind speed of $[0.1, 20]\,\text{m/s}$, consistent with the upper range of wind conditions used in pedestrian-level comfort and safety studies and aligned with external aerodynamics examples in SimScale's pedestrian wind comfort documentation~\cite{simscalePWC,CoLWind2019}. This reference speed is chosen to approximate strong but realistic urban wind events below typical design storm conditions in EN~1991-1-4 and the London City Wind Microclimate Guidelines~\cite{EN1991-1-4,CoLWind2019}.

\subsection{Velocity Statistics}

\begin{figure}[H]
    \centering
    \includegraphics[width=0.5\linewidth]{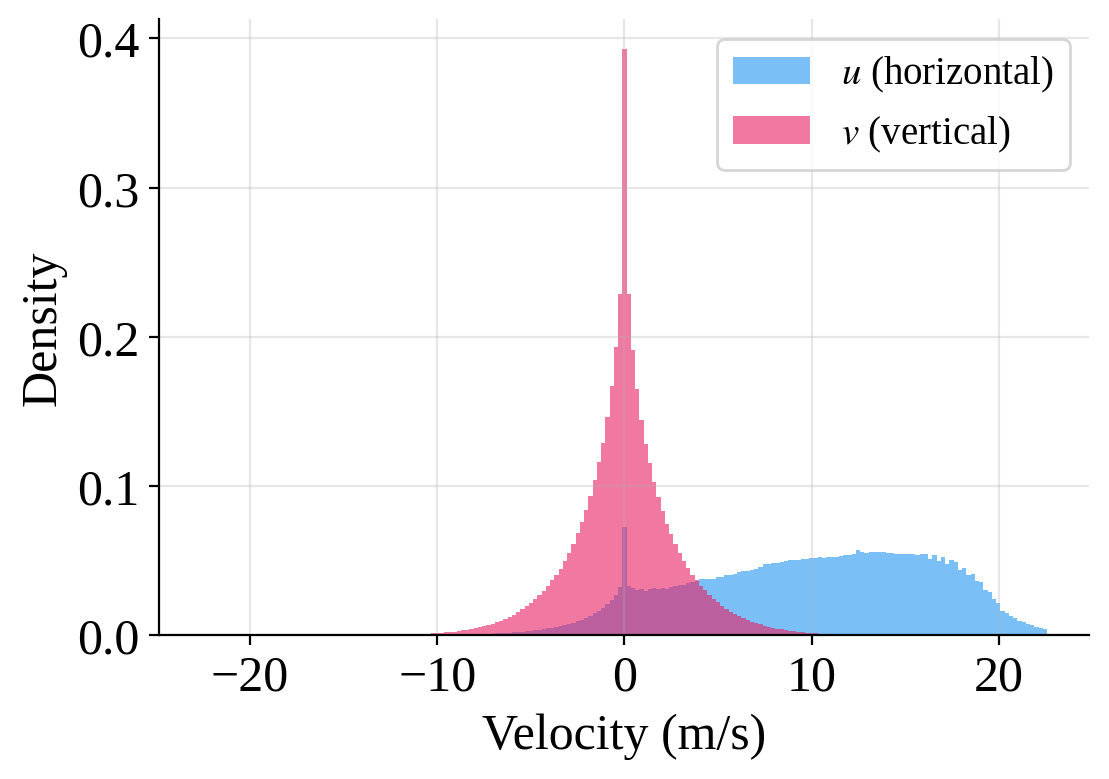}
    \caption{Marginal distributions of horizontal ($u$) and vertical ($v$) velocity components across the training set.}
    \label{fig:speed_distribution}
\end{figure}

Figure~\ref{fig:speed_distribution} shows the marginal speed distributions for both velocity components across the full training set. The horizontal component $u$ is approximately uniform between 0 and 20\,m/s, which is expected since the inlet flow is always aligned with the $u$-axis. The vertical component $v$ is roughly normally distributed around zero, with most values confined to $\pm10$\,m/s. Both distributions exhibit a sharp peak at zero, likely due to the no-slip boundary conditions at the building walls.

\section{Baseline Model Details}
\label{sec:appendix_models}

This section describes the six neural surrogate models evaluated for predicting
urban wind velocity fields.
Each model takes an initial condition (building footprint, wind speed, spatial
scale) and predicts the horizontal velocity components $(u, v)$ over a
$256\times256$ spatial grid across $T{=}112$ temporal steps.

All models are evaluated on a held-out test set of 2\,000 samples.
Autoregressive and one-shot models operate on ground-truth sequences downsampled by a factor of~4, producing 28-frame predictions that span the same physical time as $T{=}112$ full-resolution frames. This coarser stride was necessary to fit the computation into GPU memory.
For all-to-all models (Poseidon, U-Net), inference proceeds
autoregressively with a fixed time step matching the dataset's temporal
resolution.

\subsection{Training Setup}
\label{sec:appendix_training}

All models were trained with the same computational budget of 24 hours on a
single node with $4\times$NVIDIA H200 GPUs.
The models follow three distinct training paradigms:

\begin{itemize}
    \item \textbf{All-to-all (frame-to-frame).}
    Models are trained on random frame pairs $(u(t_k), u(t_{k'}))$ with
    arbitrary time separation $\Delta t \in [0, T]$, conditioned on the
    normalized time delta. At inference, these models behave autoregressively, stepping forward one
    frame at a time with a fixed $\Delta t$ matching WinDiNet's temporal resolution.

    \item \textbf{Autoregressive (direct prediction).}
    Models are trained on 28-frame sequences (the full 112-frame rollout downsampled by stride~4), predicting the next frame given a history of previous frames. Teacher forcing is applied at a ratio of 0.5: in half of the training steps, the model conditions on ground-truth history. In the other half, it conditions on its own predictions.

    \item \textbf{One-shot (direct).}
    Models predict all 28 output frames simultaneously from the initial condition (same stride-4 downsampling as above).
\end{itemize}

\subsection{Poseidon}
\label{sec:appendix_poseidon}

\noindent\textit{Type: all-to-all (frame-to-frame with $\Delta t$ conditioning). Parameters: 629\,M.}

\medskip\noindent
Poseidon~\cite{herde2024poseidon} is a pre-trained foundation model for PDE operator learning.
Conditioning is implemented through lead-time conditioned layer norms, where
the normalization parameters are affine functions of $\Delta t$, $u_\mathrm{in}$, and $L$ for
continuous-in-time evaluation with simulation-specific parameters.

\subsection{U-Net with FiLM Conditioning}
\label{sec:appendix_unet}

\noindent\textit{Type: all-to-all (frame-to-frame with $\Delta t$ conditioning). Parameters: 3.4\,M.}

\medskip\noindent
A convolutional U-Net~\cite{ronneberger2015unet} encoder--decoder (3 stages,
48 base channels) with self-attention blocks in the bottleneck and Feature-wise Linear Modulation
(FiLM)~\cite{perez2018film} for time and scalar conditioning.
FiLM generates per-channel affine parameters from a conditioning vector
containing Fourier embeddings of $\Delta t$, wind speed, and spatial scale.

\subsection{OFormer}
\label{sec:appendix_oformer}

\noindent\textit{Type: autoregressive (28-frame sequences, teacher forcing
ratio 0.5). Parameters: 99\,M.}

\medskip\noindent
OFormer~\cite{li2023transformer} is an attention-based operator transformer for
PDE operator learning. We add causal temporal attention masks to make it autoregressive.
The inlet speed is encoded in the first frame (initial condition), and the domain size $L$ is provided through a normalized coordinate grid concatenated as additional input channels.

\subsection{RNO (Recurrent Neural Operator)}
\label{sec:appendix_rno}

\noindent\textit{Type: autoregressive (28-frame sequences, teacher forcing
ratio 0.5). Parameters: 1.2\,M.}

\medskip\noindent
The Recurrent Neural Operator~\cite{liuschiaffini2023tipping} extends the GRU to
function spaces by replacing linear weight matrices with
FNO~\cite{li2021fourier} spectral convolution layers.
Our implementation uses a 2D FNO spatial encoder (Fourier modes $(16, 16)$, 64
hidden channels) that processes each frame independently, followed by a GRU (1
layer, 64 hidden units) applied per spatial location for temporal recurrence.
A pointwise $1{\times}1$ convolution decodes to the output channels.
As with OFormer, the inlet speed is encoded in the first frame and the domain size $L$ is provided through a normalized coordinate grid.

\subsection{PhysicsNeMo AFNO}
\label{sec:appendix_afno}

\noindent\textit{Type: one-shot (direct prediction of 28 frames). Parameters: 8.7\,M.}

\medskip\noindent
The Adaptive Fourier Neural Operator~\cite{guibas2022efficient} is a Vision Transformer variant that replaces self-attention with a Fourier-domain token mixer.
We use the NVIDIA PhysicsNeMo~\cite{physicsnemo} implementation.
The model predicts all 28 frames simultaneously from the initial condition in a single forward pass.
The inlet speed is encoded in the first frame and the domain size $L$ is provided through a normalized coordinate grid.

\subsection{PhysicsNeMo FNO}
\label{sec:appendix_fno}

\noindent\textit{Type: one-shot (direct prediction of 28 frames). Parameters: 67\,M.}

\medskip\noindent
The Fourier Neural Operator~\cite{li2021fourier} learns operator mappings through spectral convolutions.
We use the NVIDIA PhysicsNeMo~\cite{physicsnemo} 3D FNO operating jointly in $(T, H, W)$ space.
The model predicts all 28 frames simultaneously from the initial condition in a single forward pass.
The inlet speed is encoded in the first frame and the domain size $L$ is provided through a normalized coordinate grid.

\section{Physics-Informed Loss}
\label{sec:appendix_physics_loss}

The physics-informed training objective is a sum of three terms:
\begin{equation}
  \mathcal{L} = \mathcal{L}_\text{data}
    + \lambda_\text{div}\,\mathcal{L}_\text{div}
    + \lambda_\text{wall}\,\mathcal{L}_\text{wall}\,,
\end{equation}
with $\lambda_\text{div} = \lambda_\text{wall} = 10$ by default.
Let $B \in \{0,1\}^{H \times W}$ denote the building footprint,
$F = 1 - B$ the fluid mask, and $p$ a spatial pixel index on the
$H{\times}W$ grid.
All losses average over $T$ frames. The divergence and no-penetration terms are activated after a warmup phase
of 10 timesteps in which only $\mathcal{L}_\text{data}$ is optimized.

\subsection{Data Term}
\label{sec:loss_data}

The default data term is a distance-weighted MSE that emphasizes fluid
regions near building walls.
Let $d(p)$ be the Euclidean distance from fluid pixel $p$ to the nearest
building boundary.
The per-pixel weight is
\begin{equation}
  \omega(p) = F(p)\left[
    1 + \alpha\exp\!\left(-\frac{d(p)^2}{2\sigma^2}\right)
  \right],
  \qquad \alpha = 2,\;\sigma = 20\,,
\end{equation}
so that pixels adjacent to buildings receive approximately $3\times$ the
weight of those far from boundaries, with building pixels zeroed out.
The loss is the weighted MSE over the two velocity components
$\mathbf{w} = (u, v)$:
\begin{equation}
  \mathcal{L}_\text{data}
    = \frac{1}{T \sum_{p} \omega_p}
      \sum_{t=1}^{T}\sum_{p} \omega_p\,
      \lVert \hat{\mathbf{w}}_{t,p} - \mathbf{w}_{t,p} \rVert^2\,.
\end{equation}

\subsection{Divergence Loss}
\label{sec:loss_div}

This term penalizes violations of the incompressibility constraint
$\nabla \cdot \mathbf{w} = 0$.
At each pixel $p = (x,y)$, the discrete divergence is approximated by
first-order finite differences:
\begin{equation}
  D_t(p)
    = \bigl[\hat{u}_t(x{+}1,y) - \hat{u}_t(x,y)\bigr]
    + \bigl[\hat{v}_t(x,y{+}1) - \hat{v}_t(x,y)\bigr]\,.
\end{equation}
A stencil validity mask $V(p)$ equals 1 only if all four corners of the
finite-difference stencil lie in fluid, which avoids spurious penalties at
building boundaries.
Let $\mathcal{V} = \{p : V(p) = 1\}$.
The loss is
\begin{equation}
  \mathcal{L}_\text{div}
    = \frac{1}{T\,\lvert\mathcal{V}\rvert}
      \sum_{t=10}^{T}\sum_{p \in \mathcal{V}} D_t(p)^2\,.
\end{equation}

\subsection{Wall No-Penetration Loss}
\label{sec:loss_wall}

This term enforces zero normal velocity at building walls.
The outward wall normal $\mathbf{n}(p)$ is obtained from the spatial gradient
of the building footprint $B$, normalized to unit length.
Let $\mathcal{W}$ denote the set of pixels on building boundaries,
dilated by 1 pixel into the fluid.
The loss penalizes the normal velocity component $\hat{\mathbf{w}}_t(p) \cdot \mathbf{n}(p)$ at each wall pixel:
\begin{equation}
  \mathcal{L}_\text{wall}
    = \frac{1}{T\,\lvert\mathcal{W}\rvert}
      \sum_{t=10}^{T}\sum_{p \in \mathcal{W}} \bigl(\hat{\mathbf{w}}_t(p) \cdot \mathbf{n}(p)\bigr)^2\,.
\end{equation}

\section{Evaluation Metrics}
\label{sec:appendix_metrics}

Besides the standard pointwise metrics VRMSE, MAE, and MRE, we include two metrics designed to capture temporal and distributional fidelity, which matter for dynamic systems where a model could achieve low pointwise error by producing temporally smoothed or statistically implausible predictions.

The \textbf{Spectral Divergence} treats each fluid pixel as an independent temporal sensor and compares the log power spectra of the predicted and ground-truth velocity signals. It is sensitive to whether the model preserves temporal frequency content (fast oscillations vs.\ slow trends) while being invariant to phase shifts.

The \textbf{Wasserstein-1 distance} ($W_1$) also operates per pixel but compares the marginal speed distribution rather than temporal structure. It measures whether the model produces the correct statistical distribution of wind speeds at each location, regardless of temporal ordering. A model that reproduces the speed histogram exactly but with different timing scores zero on $W_1$ (but may score poorly on the spectral metric).

\section{Inverse Optimization: Method and Multi-Inlet Results}
\label{app:inverse_opt}

\subsection{Problem Formulation}

We formulate urban layout optimization as an inverse design problem.
Let $\mathbf{c} = \{c_i\}_{i=1}^{N_\mathrm{train}} \in \mathbb{R}^{N_\mathrm{train} \times 2}$ denote the trainable building centers ($N_\mathrm{train}=15$ out of 52 buildings on a $1100\,\text{m} \times 1100\,\text{m}$ domain).
The optimization objective is
\begin{equation}
  \label{eq:inv_objective}
  \min_{\mathbf{c}} \;
    \mathcal{L}\!\bigl(\mathcal{S}(\mathcal{G}(\mathbf{c}))\bigr)
    + \lambda_\mathrm{move}\,R_\mathrm{move}(\mathbf{c})
    + \lambda_\mathrm{coh}\,R_\mathrm{coh}(\mathbf{c}),
\end{equation}
where $\mathcal{G}$ is a differentiable rasterizer (Sec.~\ref{sec:diff_rast}), $\mathcal{S}$ is the frozen WinDiNet surrogate, $\mathcal{L}$ is the pedestrian wind comfort loss (Sec.~\ref{sec:pwc_loss}), and $R_\mathrm{move}$, $R_\mathrm{coh}$ are regularization terms detailed in Sec.~\ref{sec:regularization}.

We use Adam~\cite{kingma2015adam} ($\beta_1=0.9$, $\beta_2=0.99$) with learning rate 1.0. Single-inlet runs use 200 steps; multi-inlet runs use 400.

\subsection{Differentiable Rasterization}
\label{sec:diff_rast}

Following the soft rasterization principle of Liu et al.~\cite{liu2019softrasterizer}, each building~$i$ with center $c_i = (c_{x}^i, c_{y}^i)$ and fixed half-extents $(a_i/2, b_i/2)$ is rasterized onto an $H \times W$ grid via a product of sigmoids:
\begin{equation}
  o_i(x,y) =
    \sigma\!\Bigl(\tfrac{x - (c_{x}^i - a_i/2)}{\tau}\Bigr)
    \cdot \sigma\!\Bigl(\tfrac{(c_{x}^i + a_i/2) - x}{\tau}\Bigr)
    \cdot \sigma\!\Bigl(\tfrac{y - (c_{y}^i - b_i/2)}{\tau}\Bigr)
    \cdot \sigma\!\Bigl(\tfrac{(c_{y}^i + b_i/2) - y}{\tau}\Bigr),
\end{equation}
with temperature $\tau = 2.0$ controlling edge softness.
The differentiable occupancy field $B \in [0,1]^{H \times W}$ is a soft union across all $N$ buildings:
\begin{equation}
  B(x,y) = 1 - \prod_{i=1}^{N}\bigl(1 - o_i(x,y)\bigr).
\end{equation}
When a discrete mask is needed for conditioning the surrogate, we apply a straight-through estimator so that the forward pass produces a binary mask while gradients flow through the soft occupancy.

\subsection{Building Parameterization}

\paragraph{Rigid mode.}
Each trainable building is parameterized by its center~$c_i \in \mathbb{R}^2$; building dimensions are fixed.

\paragraph{Morph mode.}
Each trainable building is subdivided into $S \times S = 2 \times 2$ independent sub-blocks, each with its own trainable center and size $(a_i/S, \, b_i/S)$.
This allows buildings to deform, split, or rearrange internally while keeping the total footprint area constant.

\subsection{Pedestrian Wind Comfort Loss}
\label{sec:pwc_loss}

Let $\lVert\hat{\mathbf{w}}_{t,x,y}\rVert = \sqrt{\hat{u}^2 + \hat{v}^2}$ be the predicted speed at pixel $(x,y)$ and frame $t$, and let $\Omega \in \{0,1\}^{H \times W}$ denote the objective region mask.
We define three sigmoid-smoothed exceedance fractions:
\begin{align}
  e_\mathrm{danger}  &= \frac{1}{|\Omega|}\sum_{t,x,y} \Omega_{x,y}\;\sigma\!\bigl(\lVert\hat{\mathbf{w}}_{t,x,y}\rVert - \theta_\mathrm{d}\bigr), & \theta_\mathrm{d} &= 15.0\;\text{m/s}, \label{eq:e_danger}\\[4pt]
  e_\mathrm{comfort} &= \frac{1}{|\Omega|}\sum_{t,x,y} \Omega_{x,y}\;\sigma\!\bigl(\lVert\hat{\mathbf{w}}_{t,x,y}\rVert - \theta_\mathrm{c}\bigr),  & \theta_\mathrm{c} &= 5.0\;\text{m/s},\label{eq:e_comfort}\\[4pt]
  e_\mathrm{stag}    &= \frac{1}{|\Omega|}\sum_{t,x,y} \Omega_{x,y}\;\sigma\!\bigl(\theta_\mathrm{s} - \lVert\hat{\mathbf{w}}_{t,x,y}\rVert\bigr),  & \theta_\mathrm{s} &= 1.0\;\text{m/s},\label{eq:e_stag}
\end{align}
where $|\Omega|$ is the total count of (mask, time, pixel) entries with $\Omega_{x,y}=1$.
The first term penalizes dangerous wind speeds, the second penalizes speeds above the comfort threshold~$\theta_\mathrm{c}$, and the third penalizes stagnation below~$\theta_\mathrm{s}$.
The total comfort loss is
\begin{equation}
  \label{eq:pwc_loss}
  \mathcal{L} = 10\,e_\mathrm{danger} + e_\mathrm{comfort} + e_\mathrm{stag}.
\end{equation}
The high weight on danger reflects a safety-first priority.

\subsection{Regularization}
\label{sec:regularization}

\paragraph{Movement penalty.}
The $N_\mathrm{train}{=}15$ trainable buildings should not drift further than necessary from their initial positions $c_i^0$. We add a mean squared displacement term:
\begin{equation}
  R_\mathrm{move} = \frac{1}{N_\mathrm{train}} \sum_{i=1}^{N_\mathrm{train}} \|c_i - c_i^0\|^2, \qquad \lambda_\mathrm{move} = 10^{-4}.
\end{equation}

\paragraph{Cohesion penalty (morph mode only).}
In morph mode each building is split into $2{\times}2$ sub-blocks that can move independently. To prevent them from scattering, we penalize any sub-block whose displacement $\delta_j$ deviates from the mean displacement $\bar{\delta}$ of its parent building by more than a hinge of 5\,m:
\begin{equation}
  R_\mathrm{coh} = \frac{1}{N_\mathrm{train}} \sum_{i=1}^{N_\mathrm{train}} \frac{1}{4} \sum_{j \in i} \max\!\bigl(\|\delta_j - \bar{\delta}_i\| - 5,\; 0\bigr)^2, \qquad \lambda_\mathrm{coh} = 0.1.
\end{equation}
The hinge allows the full building to translate freely; only the relative spread of its sub-blocks is penalized.

\subsection{Multi-Inlet Aggregation}

For $K$ inlet wind directions, the total flow loss is a uniform average:
\begin{equation}
  \label{eq:multi_inlet}
  \mathcal{L} = \frac{1}{K}\sum_{k=1}^{K} \mathcal{L}^{(k)}.
\end{equation}
Each direction may use different thresholds ($\theta_\mathrm{d}^{(k)}, \theta_\mathrm{c}^{(k)}, \theta_\mathrm{s}^{(k)}$) to reflect different requirements under different wind directions.

\subsection{Experimental Configurations}

\begin{table}[H]
\centering

\caption{%
  Inverse optimization configurations. All runs share the same initial layout (52 buildings, 15 trainable). For two-inlet runs, thresholds are listed as [left, top].
}
\begin{tabular}{lccccc}
\toprule
Run & Inlets & Mode & Steps & $\theta_\mathrm{c}$ (m/s) & $\theta_\mathrm{s}$ (m/s) \\
\midrule
1-inlet rigid & $(15, 0)$              & Rigid & 200 & 5.0        & 1.0        \\
1-inlet morph & $(15, 0)$              & Morph & 200 & 5.0        & 1.0        \\
2-inlet rigid & $(15, 0),\;(0, 15)$    & Rigid & 400 & $[3.0,\,5.0]$ & $[1.0,\,3.0]$ \\
2-inlet morph & $(15, 0),\;(0, 15)$    & Morph & 400 & $[3.0,\,5.0]$ & $[1.0,\,3.0]$ \\
\bottomrule
\end{tabular}
\label{tab:inv_configs}
\end{table}

\subsection{Multi-Inlet Results}

\begin{table}[t]
\centering
\footnotesize
\caption{Ground-truth wind speed distribution (\%) within the objective region, before and after layout optimization with WinDiNet.}
\label{tab:wind_zones_extended}
\begin{tabular}{lllrrrrr}
\toprule
Experiment & Inlet & Stage & \enspace$<1$\,m/s & \enspace$1$--$3$\,m/s & \enspace$3$--$5$\,m/s & \enspace$5$--$15$\,m/s & \enspace$>15$\,m/s \\
\midrule
\multirow{2}{*}{1-inlet rigid}
 & \multirow{2}{*}{$(15,0)$} & Initial   &  6.8 & 25.1 & 18.5 & 47.1 & 2.6 \\
 &                            & Optimized & 23.7 & 45.1 & 18.4 & 12.6 & 0.2 \\
\midrule
\multirow{2}{*}{1-inlet morph}
 & \multirow{2}{*}{$(15,0)$} & Initial   &  6.8 & 25.1 & 18.5 & 47.1 & 2.6 \\
 &                            & Optimized & 19.3 & 39.1 & 22.9 & 18.3 & 0.4 \\
\midrule
\multirow{4}{*}{2-inlet rigid}
 & \multirow{2}{*}{$(15,0)$} & Initial   &  6.8 & 25.1 & 18.5 & 47.1 & 2.6 \\
 &                            & Optimized & 23.7 & 39.3 & 19.2 & 17.4 & 0.4 \\
\cmidrule{2-8}
 & \multirow{2}{*}{$(0,15)$} & Initial   &  7.1 & 23.2 & 20.3 & 47.3 & 2.0 \\
 &                            & Optimized & 14.2 & 33.8 & 27.0 & 24.7 & 0.2 \\
\midrule
\multirow{4}{*}{2-inlet morph}
 & \multirow{2}{*}{$(15,0)$} & Initial   &  6.8 & 25.1 & 18.5 & 47.1 & 2.6 \\
 &                            & Optimized & 22.6 & 44.0 & 17.8 & 15.4 & 0.2 \\
\cmidrule{2-8}
 & \multirow{2}{*}{$(0,15)$} & Initial   &  7.1 & 23.2 & 20.3 & 47.3 & 2.0 \\
 &                            & Optimized & 15.4 & 35.0 & 25.7 & 23.8 & 0.1 \\
\bottomrule
\end{tabular}
\end{table}

\begin{figure}[H]
  \centering
  \raisebox{-0.03\textwidth}{\rotatebox{90}{\small\textit{Rigid}}}%
  \hfill
  \begin{subfigure}[b]{0.28\textwidth}
    \centering
    \includegraphics[width=\textwidth]{figures/inverse_opt/2inlet_rigid/gt_snapshot_initial_left.png}
    \captionsetup{margin={0pt,10pt}}
    \caption{Left inlet: initial}
  \end{subfigure}\hfill
  \begin{subfigure}[b]{0.28\textwidth}
    \centering
    \includegraphics[width=\textwidth]{figures/inverse_opt/2inlet_rigid/gt_snapshot_final_left.png}
    \captionsetup{margin={0pt,10pt}}
    \caption{Left inlet: optimized}
  \end{subfigure}\hfill
  \begin{subfigure}[b]{0.34\textwidth}
    \centering
    \includegraphics[width=\textwidth]{figures/inverse_opt/2inlet_rigid/gt_speed_distribution_left.png}
    \caption{Left inlet: distribution}
  \end{subfigure}
  \\[0.4em]
  \phantom{\rotatebox{90}{\small\textit{Rigid}}}%
  \hfill
  \begin{subfigure}[b]{0.28\textwidth}
    \centering
    \includegraphics[width=\textwidth]{figures/inverse_opt/2inlet_rigid/gt_snapshot_initial_top.png}
    \captionsetup{margin={0pt,10pt}}
    \caption{Top inlet: initial}
  \end{subfigure}\hfill
  \begin{subfigure}[b]{0.28\textwidth}
    \centering
    \includegraphics[width=\textwidth]{figures/inverse_opt/2inlet_rigid/gt_snapshot_final_top.png}
    \captionsetup{margin={0pt,10pt}}
    \caption{Top inlet: optimized}
  \end{subfigure}\hfill
  \begin{subfigure}[b]{0.34\textwidth}
    \centering
    \includegraphics[width=\textwidth]{figures/inverse_opt/2inlet_rigid/gt_speed_distribution_top.png}
    \caption{Top inlet: distribution}
  \end{subfigure}
  \\[1.8em]
  \raisebox{-0.03\textwidth}{\rotatebox{90}{\small\textit{Morph}}}%
  \hfill
  \begin{subfigure}[b]{0.28\textwidth}
    \centering
    \includegraphics[width=\textwidth]{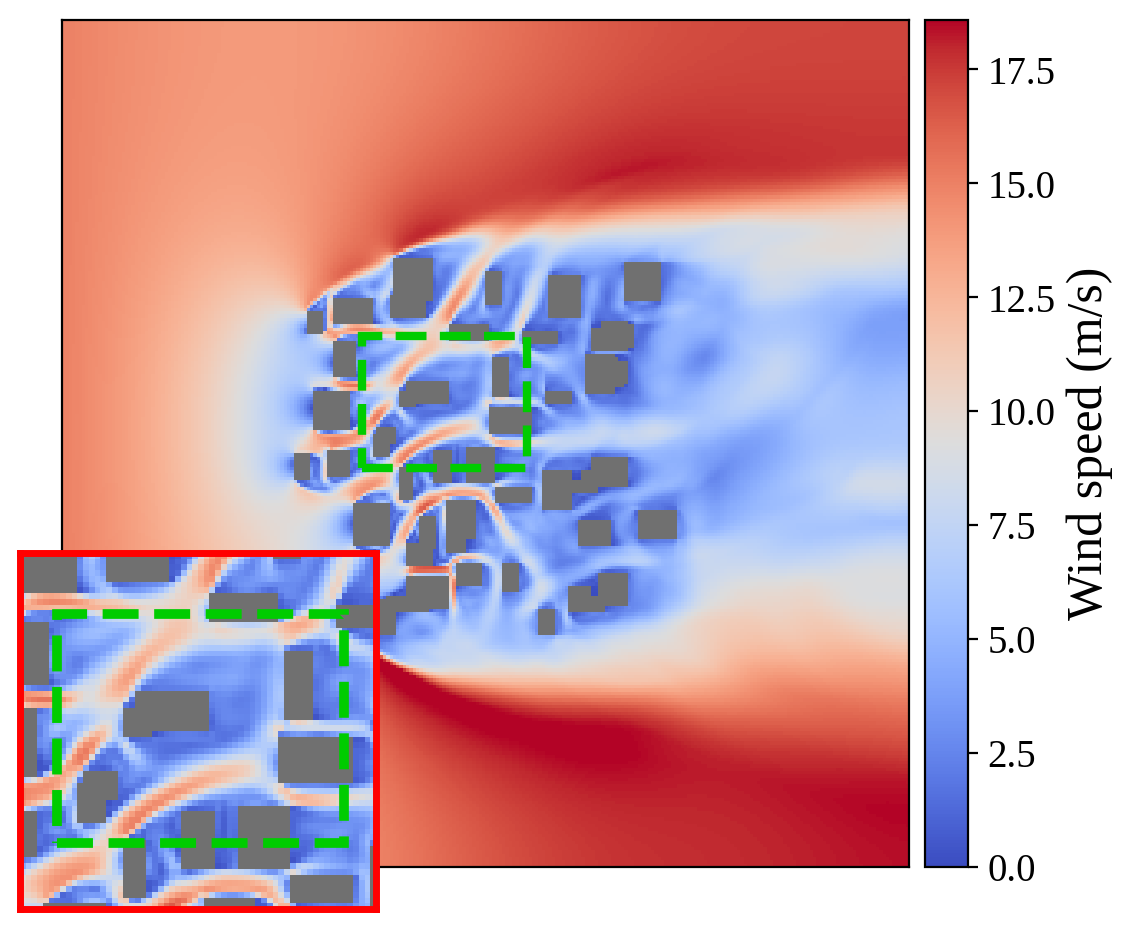}
    \captionsetup{margin={0pt,10pt}}
    \caption{Left inlet: initial}
  \end{subfigure}\hfill
  \begin{subfigure}[b]{0.28\textwidth}
    \centering
    \includegraphics[width=\textwidth]{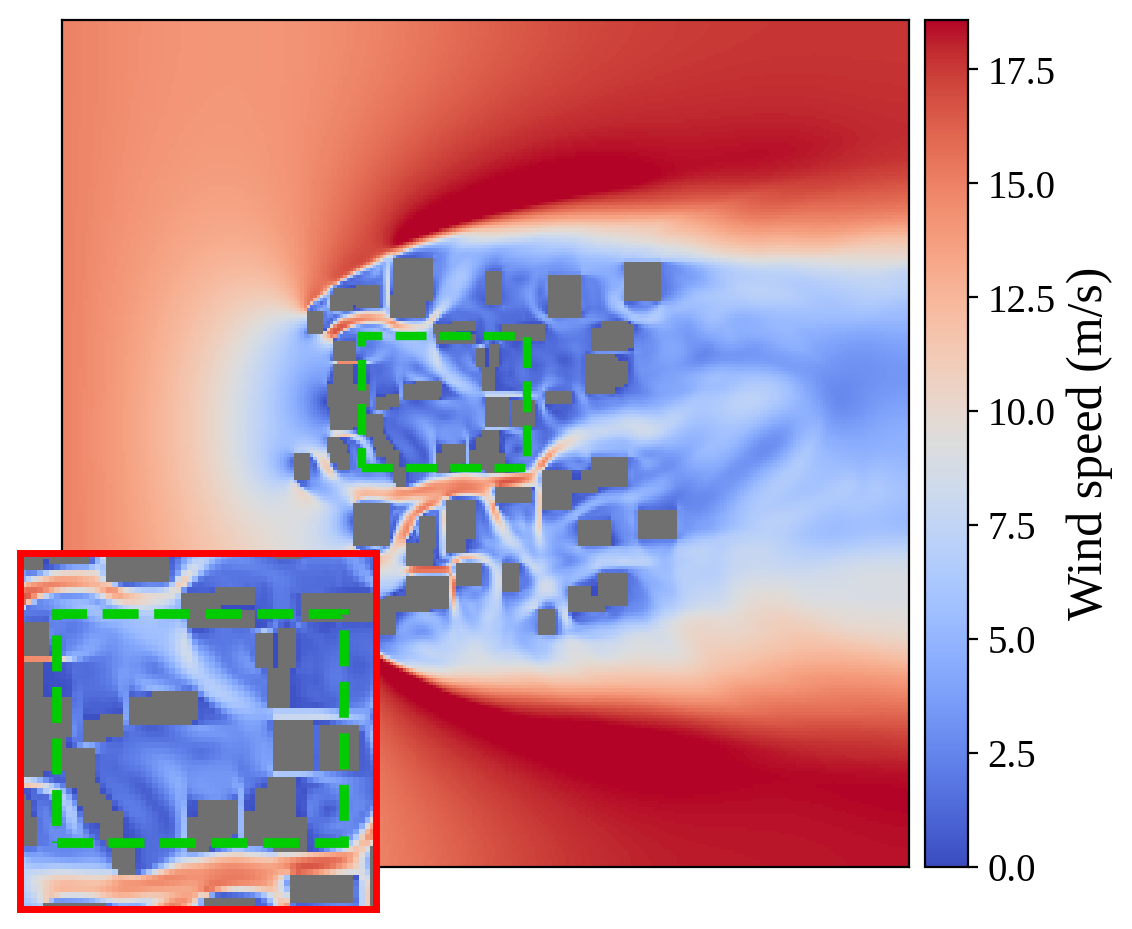}
    \captionsetup{margin={0pt,10pt}}
    \caption{Left inlet: optimized}
  \end{subfigure}\hfill
  \begin{subfigure}[b]{0.34\textwidth}
    \centering
    \includegraphics[width=\textwidth]{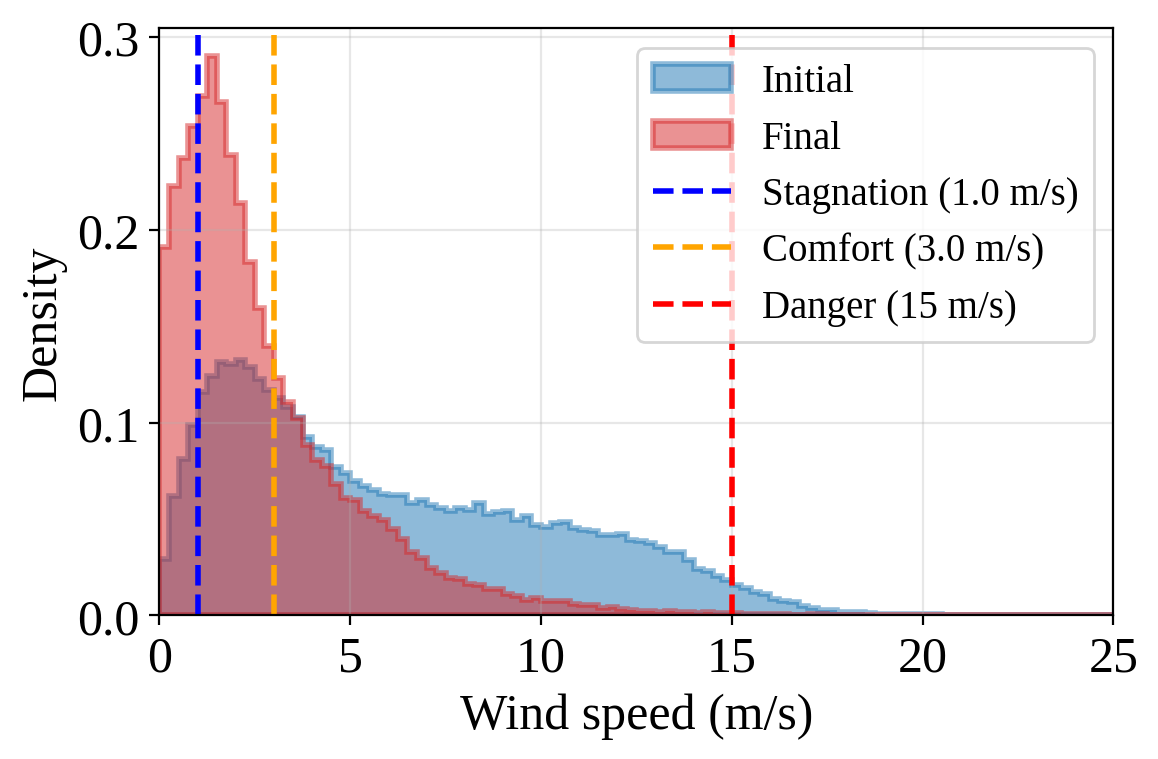}
    \caption{Left inlet: distribution}
  \end{subfigure}
  \\[0.4em]
  \phantom{\rotatebox{90}{\small\textit{Morph}}}%
  \hfill
  \begin{subfigure}[b]{0.28\textwidth}
    \centering
    \includegraphics[width=\textwidth]{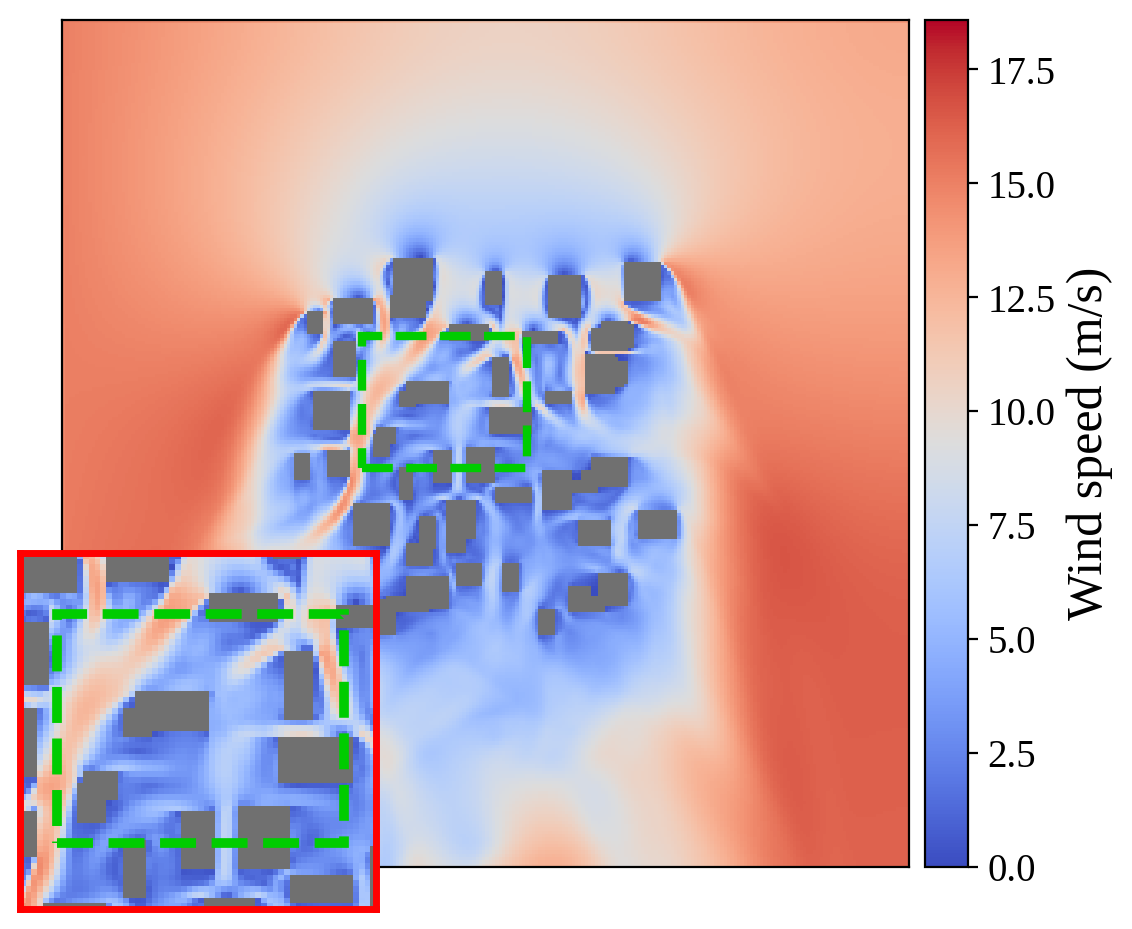}
    \captionsetup{margin={0pt,10pt}}
    \caption{Top inlet: initial}
  \end{subfigure}\hfill
  \begin{subfigure}[b]{0.28\textwidth}
    \centering
    \includegraphics[width=\textwidth]{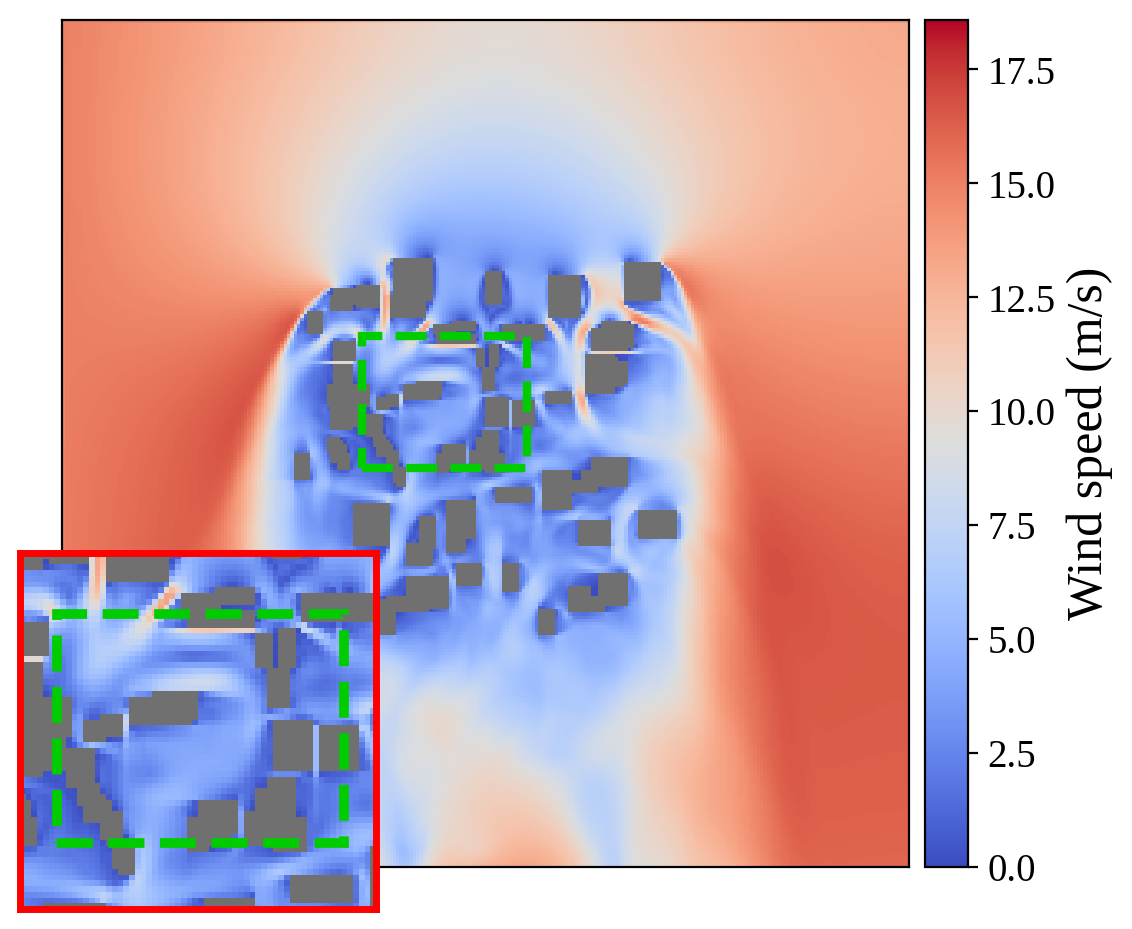}
    \captionsetup{margin={0pt,10pt}}
    \caption{Top inlet: optimized}
  \end{subfigure}\hfill
  \begin{subfigure}[b]{0.34\textwidth}
    \centering
    \includegraphics[width=\textwidth]{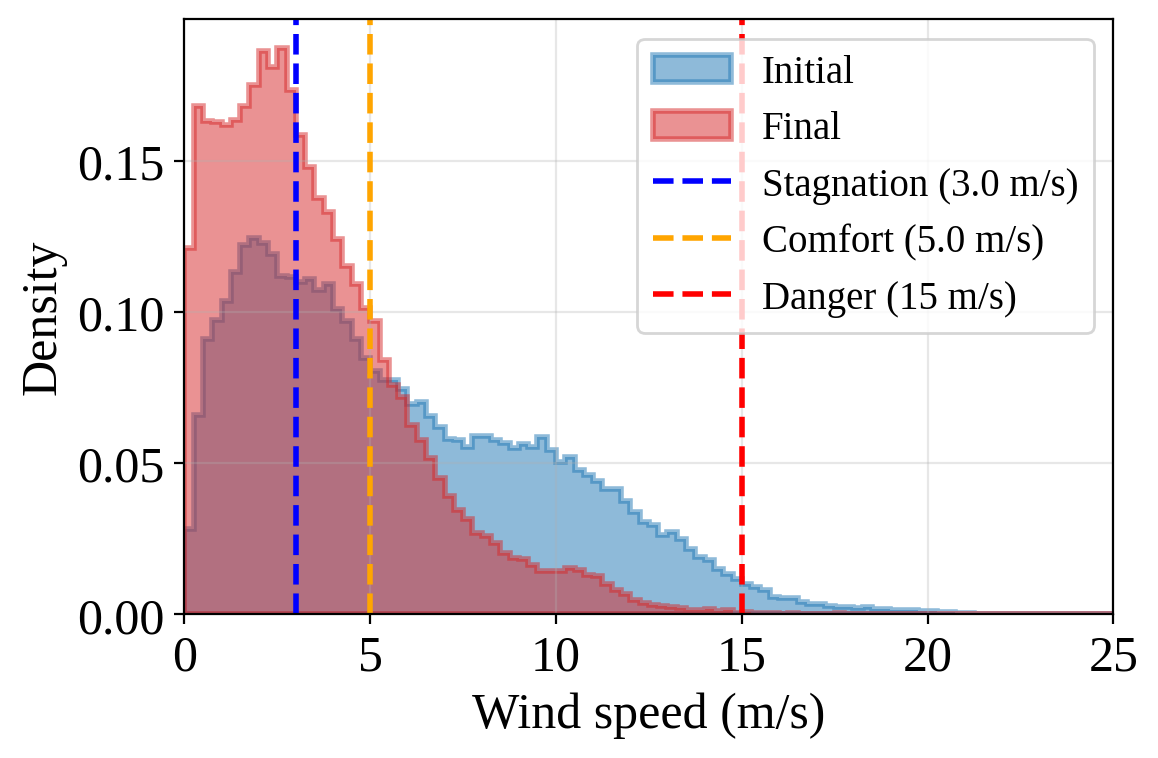}
    \caption{Top inlet: distribution}
  \end{subfigure}
  \caption{%
    Multi-inlet layout optimization: left $(15,0)$\,m/s (comfort band 1--3\,m/s) and top $(0,15)$\,m/s (comfort band 3--5\,m/s). Rigid mode (a--f) translates buildings, morph mode (g--l) additionally deforms them.
  }
  \label{fig:multi_inlet_results}
\end{figure}

Figure~\ref{fig:multi_inlet_results} shows results for two-inlet optimization in both rigid and morph modes.
A single optimized layout must balance conflicting objectives: the left inlet $(15, 0)$\,m/s targets a comfort band of 1--3\,m/s ($\theta_\mathrm{c}=3.0$, $\theta_\mathrm{s}=1.0$), while the top inlet $(0, 15)$\,m/s targets 3--5\,m/s ($\theta_\mathrm{c}=5.0$, $\theta_\mathrm{s}=3.0$).
This becomes possible because WinDiNet produces time-resolved velocity fields rather than aggregated statistics, which allows the optimizer to reason about the full flow dynamics under each direction independently. A practical scenario is a city where wind arrives predominantly from the west in winter and from the north in summer: planners may want to shelter pedestrians from cold winter gusts while preserving airflow for natural ventilation in summer.

The rigid rows of Figure~\ref{fig:multi_inlet_results} show the solution under both wind directions. The optimizer finds a compromise geometry that reduces comfort violations for both winds, even though the speed targets and flow patterns differ substantially.
For left wind, the initial distribution is broad with mass well above the 3\,m/s comfort threshold; final speeds shift toward the 1--3\,m/s band.
For top wind, initial speeds show stagnation below 3\,m/s; final speeds concentrate in the 3--5\,m/s target range.

The morph rows show the corresponding solution.
Sub-block deformation provides finer geometric control, allowing some buildings to partially reshape and redirect flow more precisely.
The speed distributions show that morph mode achieves a tighter concentration within the respective comfort bands for both wind directions.

The compromise emerges naturally from the multi-inlet aggregation~(Eq.~\ref{eq:multi_inlet}): the gradient $\nabla_{\mathbf{c}} \mathcal{L}_\mathrm{flow}$ averages contributions from each direction, so each wind exerts equal pressure on the layout.
A building repositioned to shelter the objective region from left wind may simultaneously alter the top-wind flow pattern, and vice versa.
The optimizer accepts suboptimal performance in either direction alone in exchange for balanced multi-directional comfort.
The multi-inlet results suggest that the surrogate and rasterizer are smooth enough to support gradient-based optimization in non-convex, multi-objective settings.

\subsection{Surrogate Comparison: WinDiNet vs.\ OFormer}

To assess how surrogate quality affects inverse optimization, we repeat the single-inlet rigid experiment using OFormer---a competitive baseline in our forward-prediction benchmarks---as the differentiable surrogate.
Both runs use identical settings (200 Adam steps, same initial layout, same loss function) and are evaluated on the \emph{ground-truth} CFD fields corresponding to the optimized layouts, so the comparison reflects actual wind-comfort improvements rather than surrogate-specific artifacts.

\begin{figure}[H]
  \centering
  \raisebox{0.1\textwidth}{\rotatebox{90}{\small\textit{OFormer}}}%
  \hfill
  \begin{subfigure}[b]{0.275\textwidth}
    \centering
    \includegraphics[width=\textwidth]{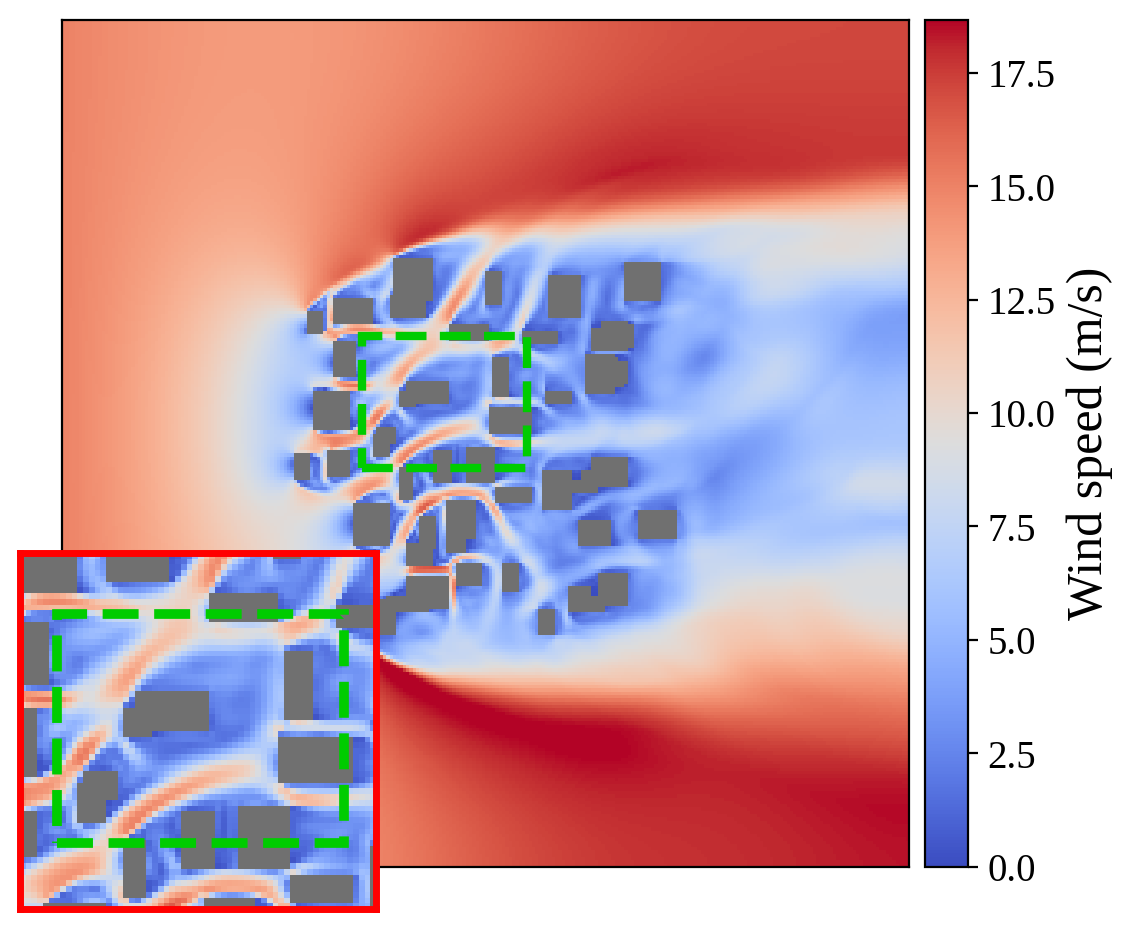}
    \captionsetup{margin={0pt,10pt}}
    \caption{Initial layout}
  \end{subfigure}
  \hfill
  \begin{subfigure}[b]{0.275\textwidth}
    \centering
    \includegraphics[width=\textwidth]{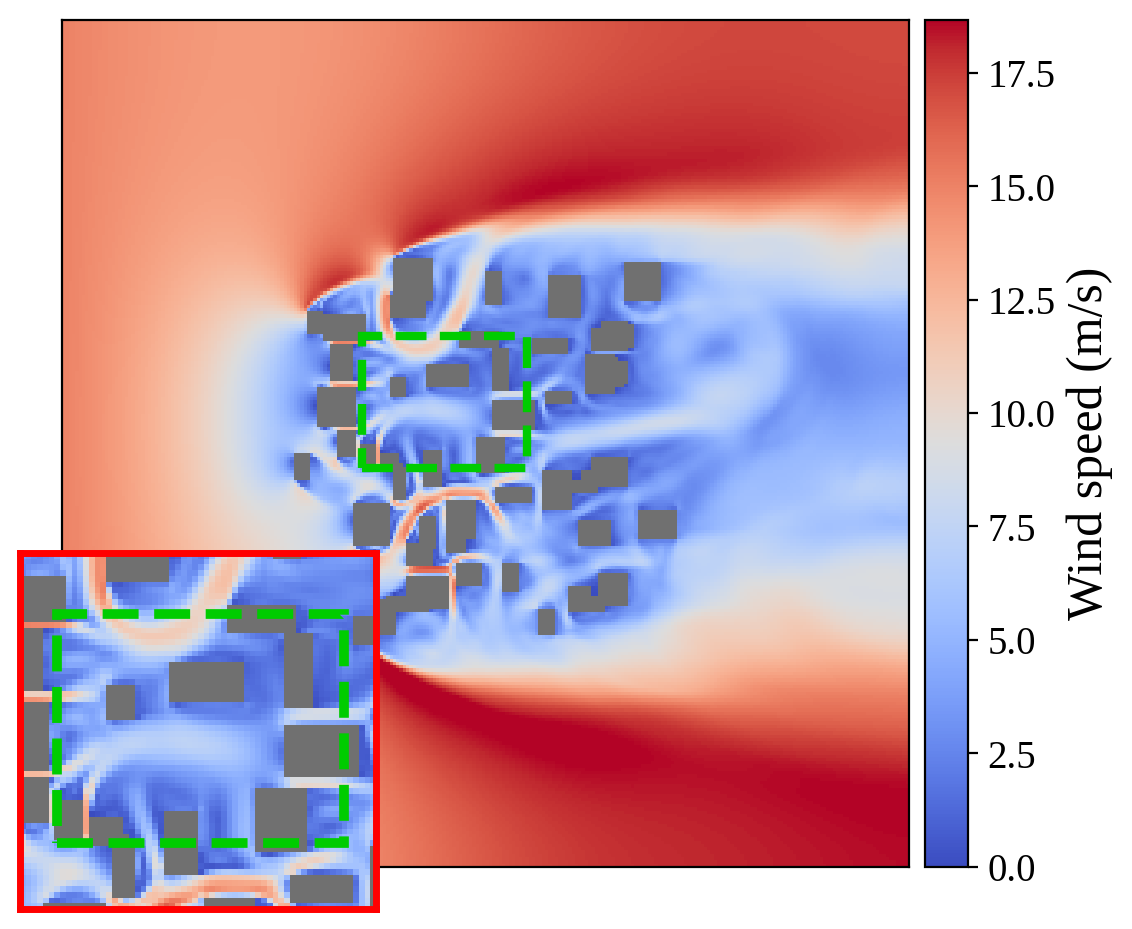}
    \captionsetup{margin={0pt,10pt}}
    \caption{Optimized layout}
  \end{subfigure}
  \hfill
  \begin{subfigure}[b]{0.355\textwidth}
    \centering
    \includegraphics[width=\textwidth]{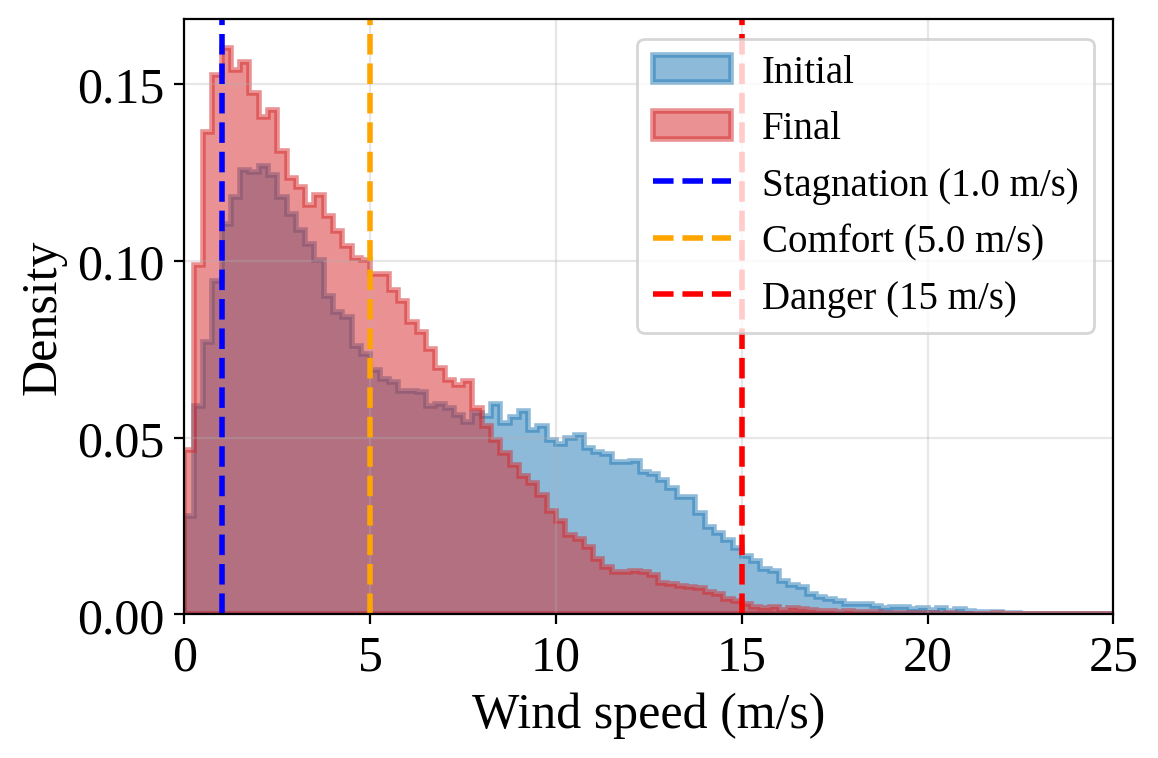}
    \caption{Distribution}
  \end{subfigure}
  \caption{%
    Inverse optimization using OFormer as surrogate (single inlet, rigid mode, Fig.~\ref{fig:inverse_opt_rigid}).
  }
  \label{fig:oformer_comparison}
\end{figure}

Figure~\ref{fig:oformer_comparison} shows the OFormer-optimized layout alongside the corresponding speed distribution.
The optimizer does shift buildings to reduce wind speeds in the objective region, but the effect is visibly weaker than under the WinDiNet surrogate (Figure~\ref{fig:inverse_opt_rigid}).
Table~\ref{tab:surrogate_comparison} quantifies this gap.
WinDiNet reduces the discomfort exceedance fraction from 49.67\% to 12.80\%, while OFormer only reaches 36.63\%.
The danger fraction drops to 0.22\% under WinDiNet versus 0.37\% under OFormer, and the 95th-percentile speed falls to 7.24\,m/s versus 10.43\,m/s.
In short, both surrogates enable gradient-based layout optimization, but the higher fidelity of the WinDiNet surrogate translates into substantially better comfort outcomes.
The stagnation fraction increases more under WinDiNet (23.69\% vs.\ 11.39\%), which is expected: more aggressive sheltering naturally produces more low-speed zones.
OFormer is also roughly $3.7\times$ slower per optimization step (8.4\,s vs.\ 2.3\,s), owing to its autoregressive rollout requirement.

\begin{table}[ht]
\centering
\caption{GT-verified comparison of inverse optimization using WinDiNet vs.\ OFormer as differentiable surrogate (single inlet, rigid mode). All metrics are evaluated on ground-truth CFD fields for the respective optimized layouts.}
\label{tab:surrogate_comparison}
\begin{tabular}{lccc}
\toprule
Metric & Initial & WinDiNet & OFormer Final \\
\midrule
Mean speed (m/s)          & 6.03  & 2.66  & 4.40  \\
Std speed (m/s)           & 4.26  & 2.39  & 3.17  \\
P95 speed (m/s)           & 13.90 & 7.24  & 10.43 \\
Danger ($>$15\,m/s)       & 2.57\%  & 0.22\%  & 0.37\%  \\
Discomfort ($>$5\,m/s)    & 49.67\% & 12.80\% & 36.63\% \\
Stagnation ($<$1\,m/s)    & 6.78\%  & 23.69\% & 11.39\% \\
Time per step             & ---     & 2.3\,s  & 8.4\,s  \\
\bottomrule
\end{tabular}
\end{table}

\subsection{Surrogate--Ground-Truth Loss Agreement}
\label{app:loss_agreement}

\begin{figure}[H]
  \centering
  \begin{subfigure}[b]{0.45\linewidth}
    \centering
    \includegraphics[width=\linewidth]{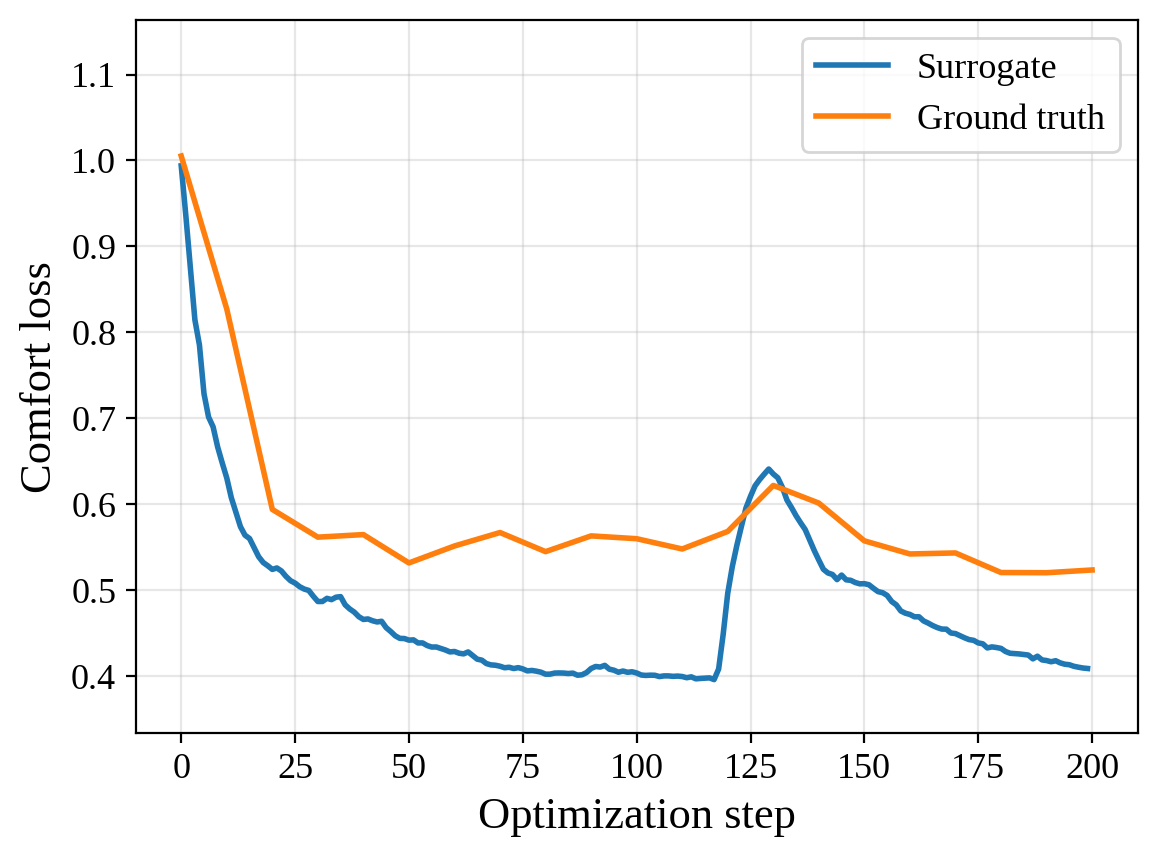}
    \caption{Single-inlet rigid}
  \end{subfigure}
  \begin{subfigure}[b]{0.45\linewidth}
    \centering
    \includegraphics[width=\linewidth]{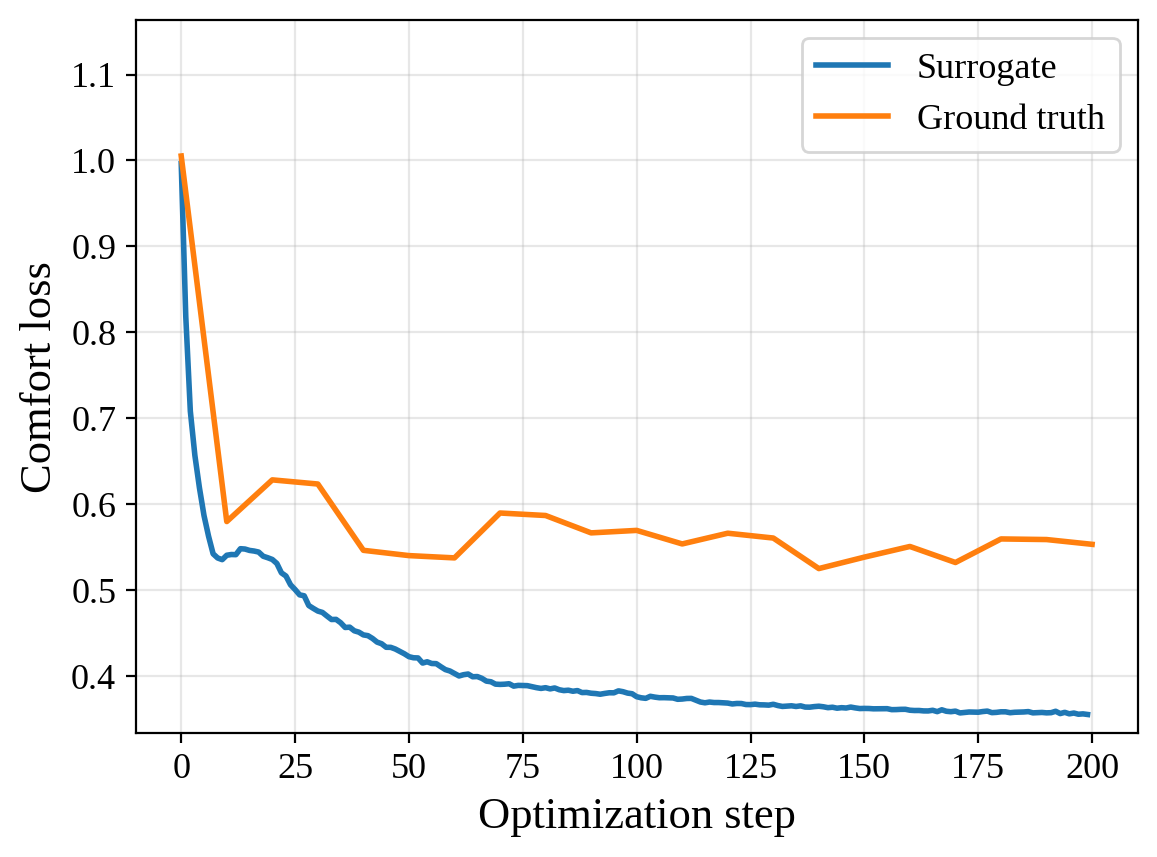}
    \caption{Single-inlet morph}
  \end{subfigure}\\[0.8em]
  \begin{subfigure}[b]{0.45\linewidth}
    \centering
    \includegraphics[width=\linewidth]{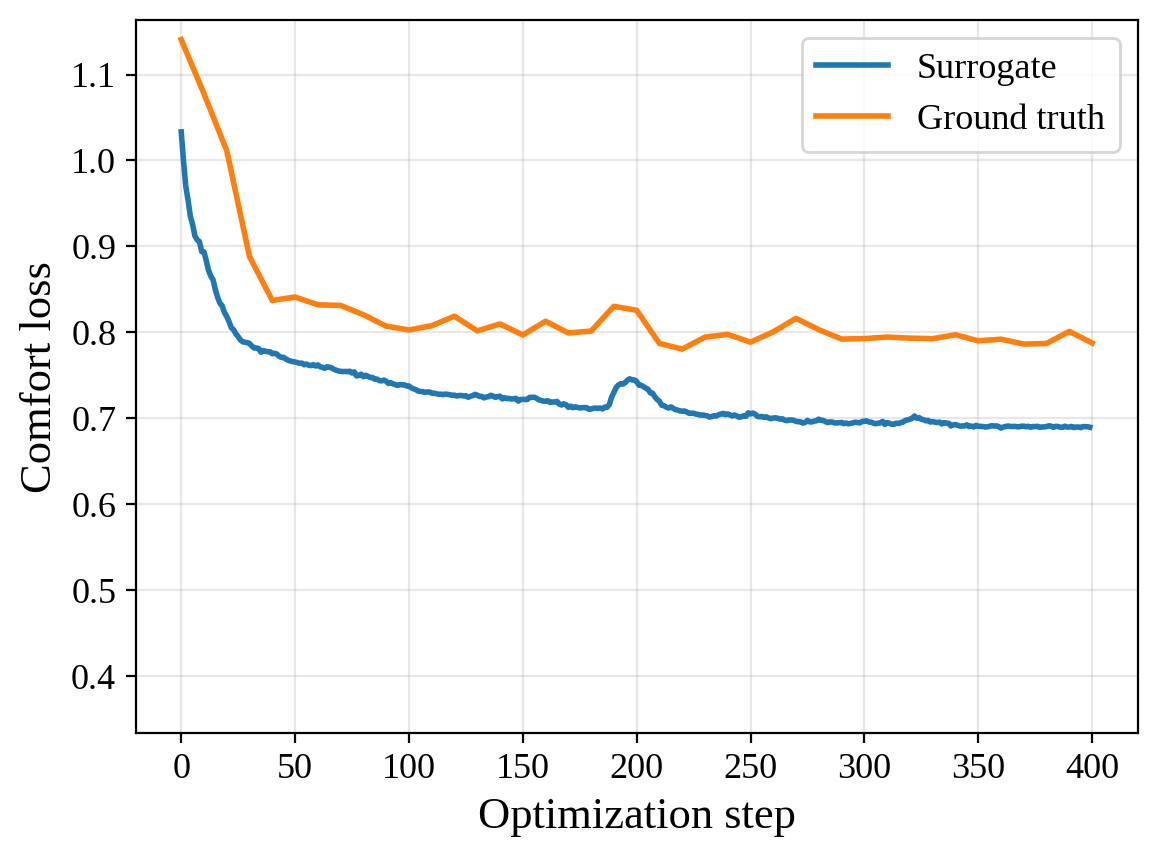}
    \caption{Multi-inlet rigid}
  \end{subfigure}
  \begin{subfigure}[b]{0.45\linewidth}
    \centering
    \includegraphics[width=\linewidth]{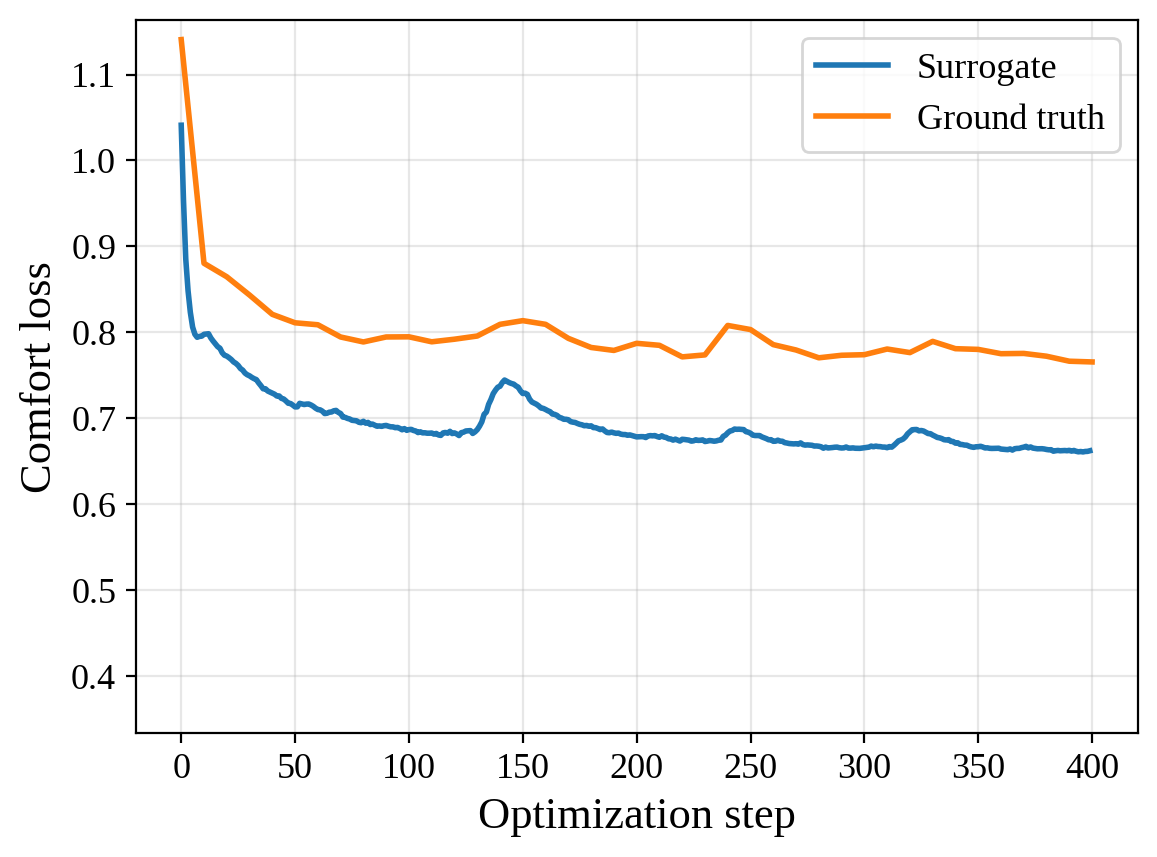}
    \caption{Multi-inlet morph}
  \end{subfigure}\\[0.8em]
  \begin{subfigure}[b]{0.45\linewidth}
    \centering
    \includegraphics[width=\linewidth]{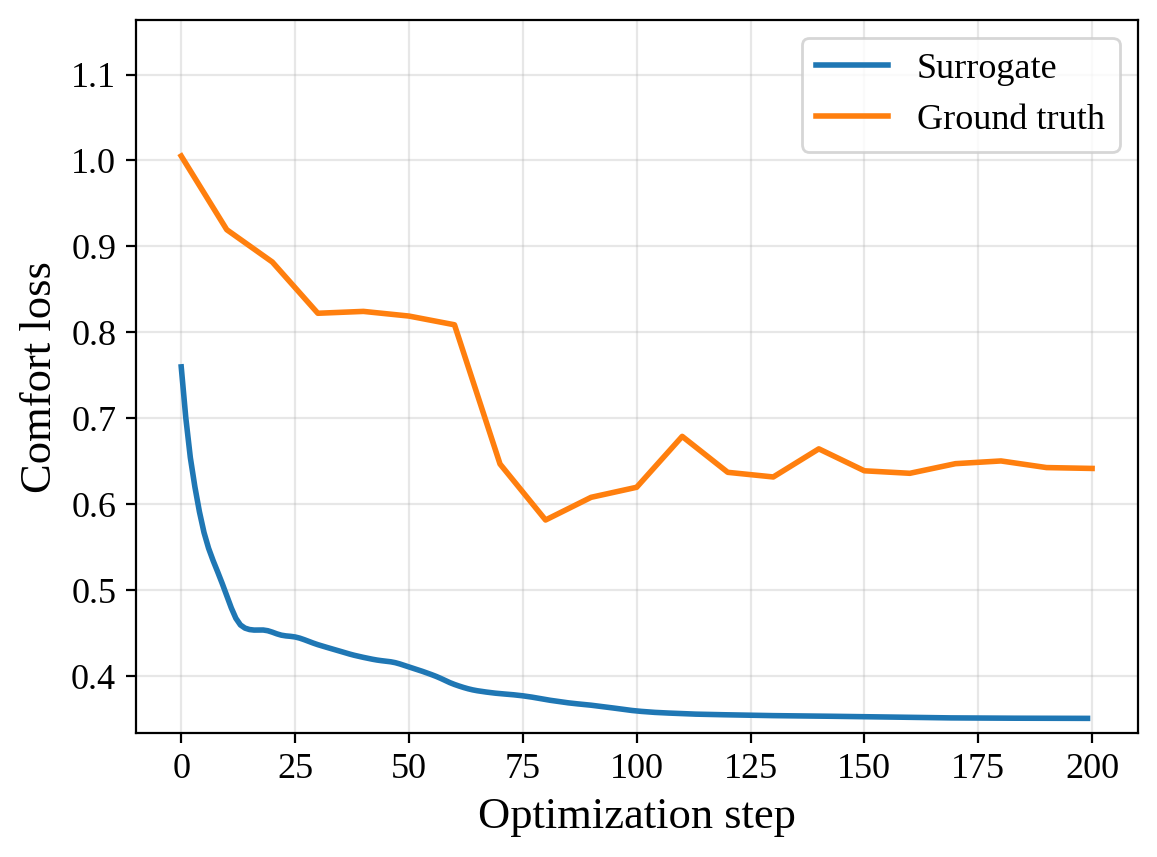}
    \caption{Single-inlet rigid (OFormer surrogate)}
  \end{subfigure}
  \caption{%
    Surrogate vs.\ ground-truth comfort loss over optimization steps for all five inverse design experiments. At each step, the CFD solver is run on the current layout and the comfort loss is evaluated on both the surrogate prediction and the true flow field.
  }
  \label{fig:loss_agreement}
\end{figure}

At every optimization step, the comfort loss is computed both on the surrogate prediction and on a ground-truth CFD solution for the current layout. Figure~\ref{fig:loss_agreement} overlays the two curves for all five experiments. The surrogate loss tracks the ground-truth loss closely throughout optimization, confirming that the gradients provided by the frozen WinDiNet surrogate guide the optimizer in a direction that genuinely improves wind comfort on the true flow field.

Even small transient increases or decreases in the ground-truth loss are reflected in the surrogate curve. Although WinDiNet consistently predicts a slightly lower comfort loss than the CFD solver, the two curves converge to nearby values without diverging. OFormer, by contrast, continues to decrease its predicted loss in later iterations while the ground-truth loss levels off or rises, suggesting that its gradients become less reliable as the layout moves away from the training distribution.

\subsection{Generality of the Objective}

The pedestrian wind comfort loss~(Eq.~\ref{eq:pwc_loss}) is one specific instantiation.
The framework extends to ventilation-focused objectives that penalize stagnation more heavily, multi-region optimization across neighborhoods with different comfort targets, wind-energy harvesting that favors consistent moderate speeds, drag minimization for structural design, and hybrid objectives combining several of these.
The sole constraint is differentiability: any objective that is continuous in the building parameters and amenable to backpropagation through the surrogate can replace or augment the comfort loss used here.

\section{Ablations and Generalization}
\label{app:generalization}

\subsection{Real Urban Configurations}
\label{sec:generalization_real}
\begin{figure}[H]
    \centering
    \includegraphics[width=\linewidth]{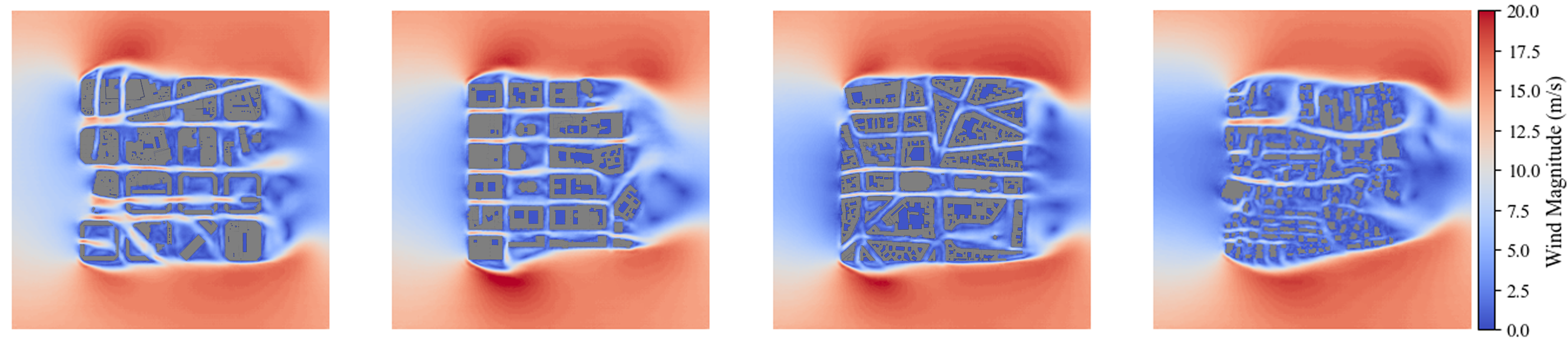}
    \caption{
    Predicted velocity fields at frame~56 for four real-world urban configurations (left to right): Barcelona, Berlin, Paris, and Zurich. The model was trained exclusively on synthetic layouts.
    }
    \label{fig:evaluation_cities}
\end{figure}

As a preliminary test of out-of-distribution generalization, we apply our best model to real-world building footprints extracted from four European cities: Barcelona, Berlin, Paris, and Zürich, selected to span a range of urban morphologies, from regular block grids to organic street networks. Figure~\ref{fig:evaluation_cities} shows the predicted velocity field at frame~56 for each city. Despite training exclusively on procedurally generated layouts, the model produces physically consistent flow structures, capturing acceleration through street canyons and recirculation in the lee of building clusters. 


\begin{table}[ht]
  \centering
  \caption{Prediction accuracy on real urban footprints at 15\,m/s inlet speed (4 cities $\times$ 8 augmentations = 32 cases). Per-city rows show WinDiNet averages over 8 augmentations. Baseline models are evaluated on the same 32 cases.}
  \label{tab:city_metrics}
  \vspace{0.5ex}
  \small
  \begin{tabular*}{\tablewidth}{@{\extracolsep{\fill}}lccccc}
    \toprule
    & \textbf{VRMSE$\downarrow$} & \textbf{MAE$\downarrow$} & \textbf{MRE$\downarrow$} & \textbf{Spectral$\downarrow$} & $\mathbf{W_1}\downarrow$ \\
    \midrule
    \multicolumn{6}{l}{\textit{WinDiNet (per city)}} \\
    \quad Barcelona & 0.687 & 2.380 & 7.88\% & 2.897 & 2.816 \\
    \quad Berlin    & 0.671 & 2.376 & 7.79\% & 2.701 & 2.776 \\
    \quad Paris     & 0.674 & 2.523 & 8.31\% & 2.960 & 3.057 \\
    \quad Zürich    & 0.698 & 2.179 & 7.27\% & 2.888 & 2.439 \\
    \midrule
    \textbf{WinDiNet} & \textbf{0.683} & \textbf{2.364} & \textbf{7.81\%} & 2.862 & 2.772 \\
    \midrule
    RNO      & 0.803 & 2.725 & 8.75\%  & \textbf{2.701} & \textbf{2.736} \\
    OFormer  & 0.904 & 2.975 & 9.44\%  & 2.708 & 2.947 \\
    Poseidon & 0.909 & 3.630 & 11.90\% & 2.938 & 4.724 \\
    AFNO     & 0.940 & 2.871 & 9.11\%  & 2.703 & 2.511 \\
    FNO      & 1.059 & 3.343 & 10.58\% & 2.703 & 3.016 \\
    U-Net     & 1.215 & 5.374 & 17.54\% & 3.086 & 8.185 \\
    \bottomrule
  \end{tabular*}
\end{table}

\subsection{Rollout Length Extrapolation}
\label{app:time_extrapolation}

\begin{figure}[H]
  \centering
  \newcommand{\snapshotw}{0.215\textwidth}
  \hspace{0.04\textwidth}%
  \makebox[\snapshotw][c]{\small $t=56$}%
  \hfill
  \makebox[\snapshotw][c]{\small $t=112$}%
  \hfill
  \makebox[\snapshotw][c]{\small $t=224$}%
  \hfill
  \makebox[\snapshotw][c]{\small $t=448$}%
  \\[0.2em]
  \raisebox{0.1\textwidth}{\rotatebox{90}{\small\textit{112\,fr}}}%
  \hfill
  \begin{subfigure}[b]{\snapshotw}
    \centering
    \includegraphics[width=\textwidth]{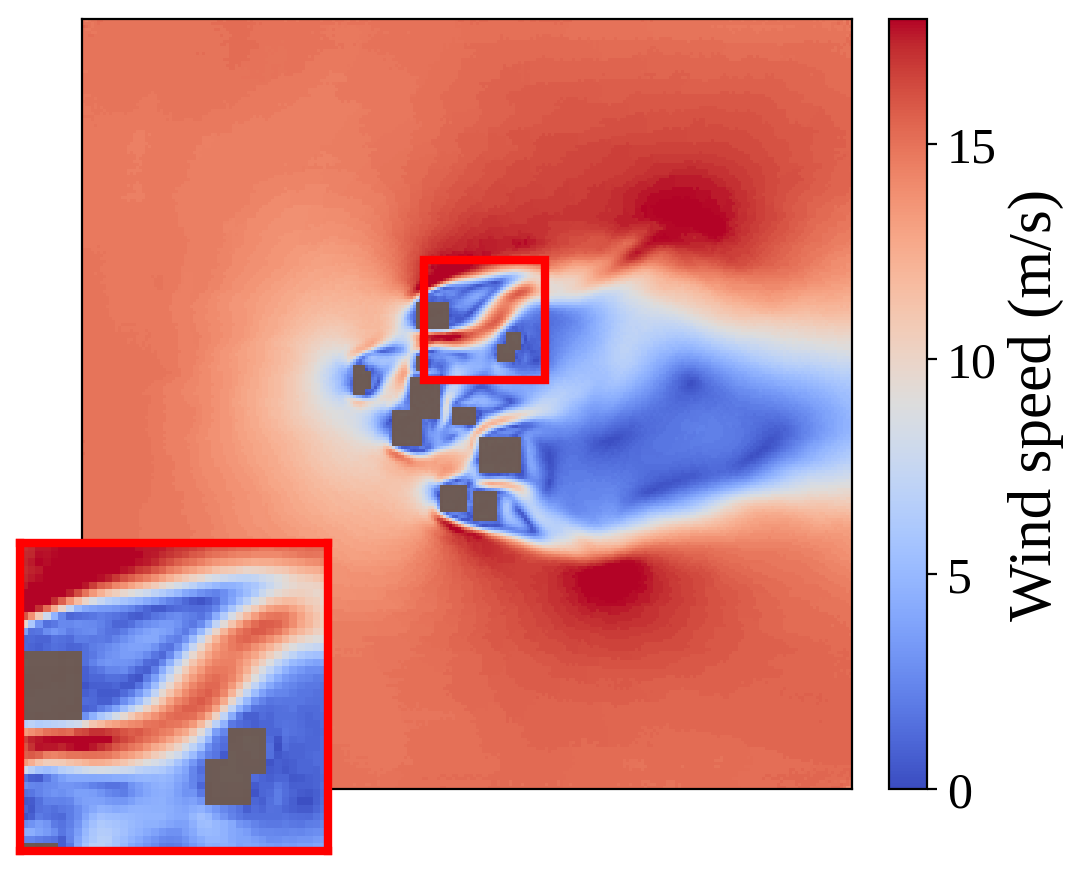}
    \captionsetup{margin={0pt,10pt}}
    \caption{}
  \end{subfigure}\hfill
  \begin{subfigure}[b]{\snapshotw}
    \centering
    \includegraphics[width=\textwidth]{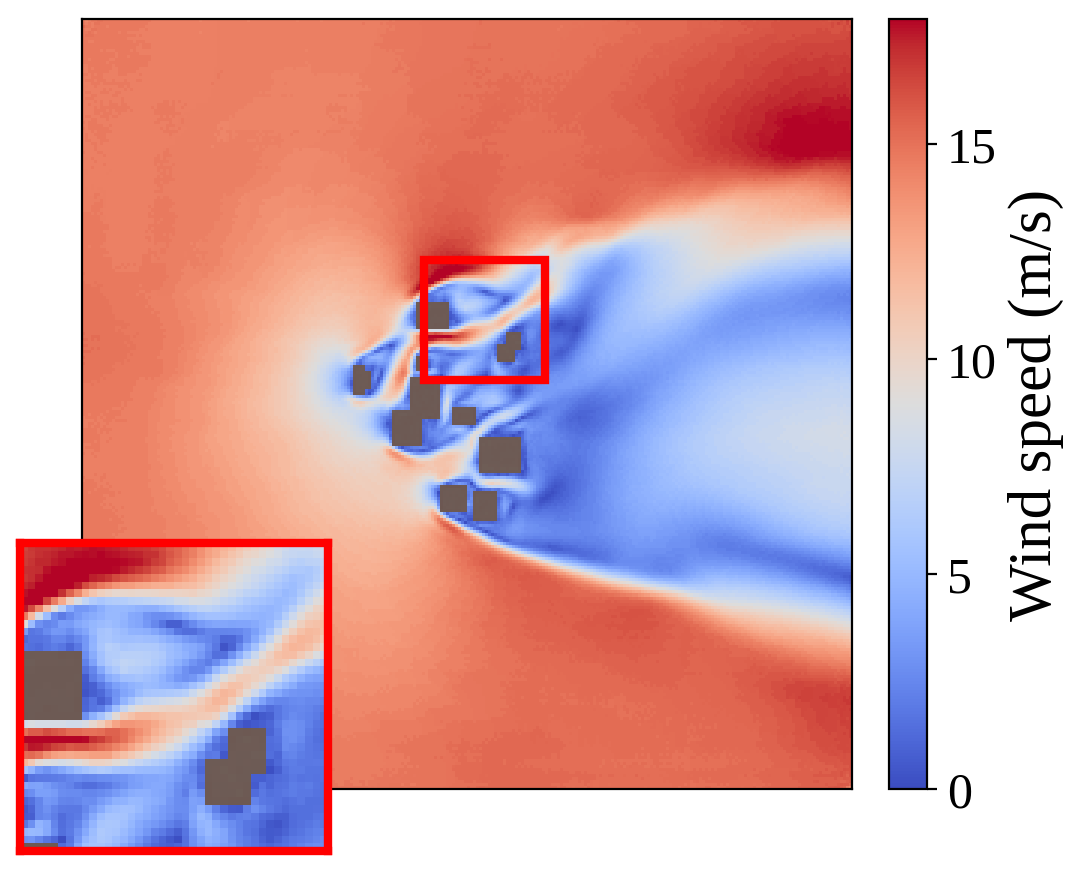}
    \captionsetup{margin={0pt,10pt}}
    \caption{}
  \end{subfigure}\hfill
  \begin{subfigure}[b]{\snapshotw}
    \centering
    \phantom{\includegraphics[width=\textwidth]{figures/ablations/time/diff_112_t0112.png}}
  \end{subfigure}\hfill
  \begin{subfigure}[b]{\snapshotw}
    \centering
    \phantom{\includegraphics[width=\textwidth]{figures/ablations/time/diff_112_t0112.png}}
  \end{subfigure}
  \\[0.3em]
  \raisebox{0.1\textwidth}{\rotatebox{90}{\small\textit{224\,fr}}}%
  \hfill
  \begin{subfigure}[b]{\snapshotw}
    \centering
    \includegraphics[width=\textwidth]{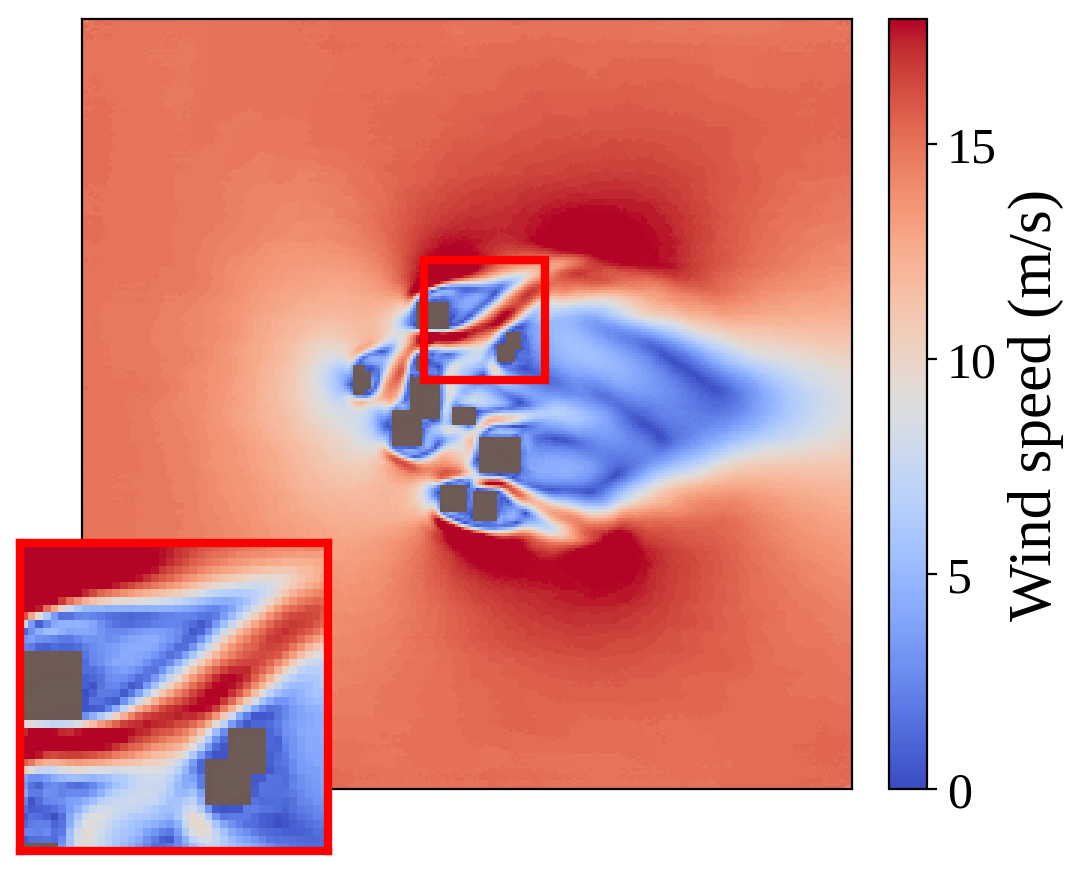}
    \captionsetup{margin={0pt,10pt}}
    \caption{}
  \end{subfigure}\hfill
  \begin{subfigure}[b]{\snapshotw}
    \centering
    \includegraphics[width=\textwidth]{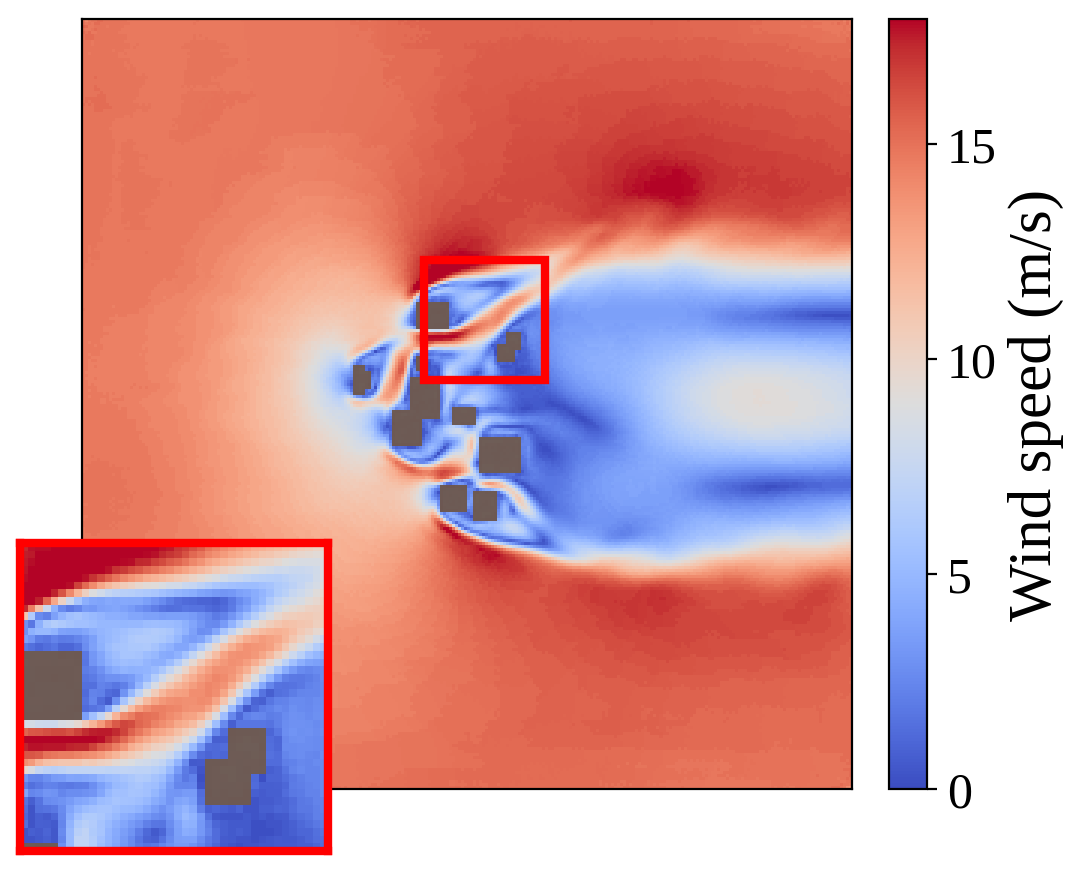}
    \captionsetup{margin={0pt,10pt}}
    \caption{}
  \end{subfigure}\hfill
  \begin{subfigure}[b]{\snapshotw}
    \centering
    \includegraphics[width=\textwidth]{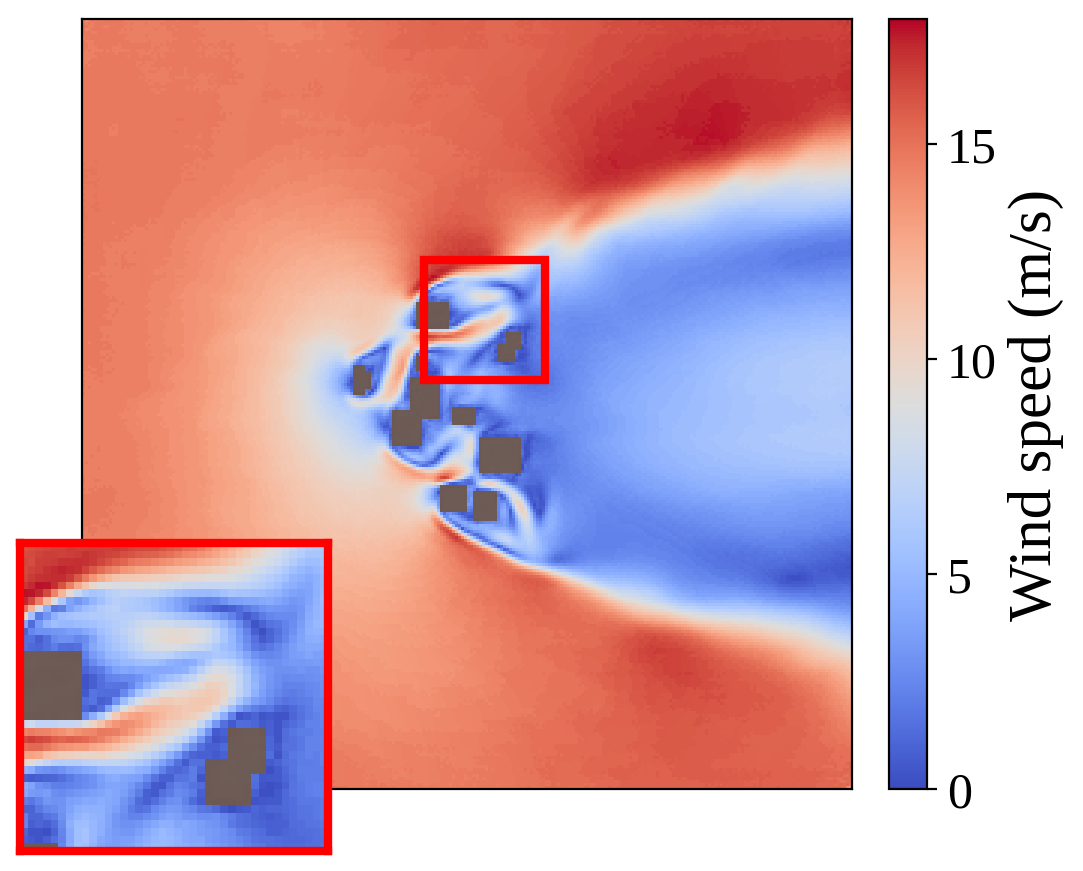}
    \captionsetup{margin={0pt,10pt}}
    \caption{}
  \end{subfigure}\hfill
  \begin{subfigure}[b]{\snapshotw}
    \centering
    \phantom{\includegraphics[width=\textwidth]{figures/ablations/time/diff_224_t0224.png}}
  \end{subfigure}
  \\[0.3em]
  \raisebox{0.1\textwidth}{\rotatebox{90}{\small\textit{448\,fr}}}%
  \hfill
  \begin{subfigure}[b]{\snapshotw}
    \centering
    \includegraphics[width=\textwidth]{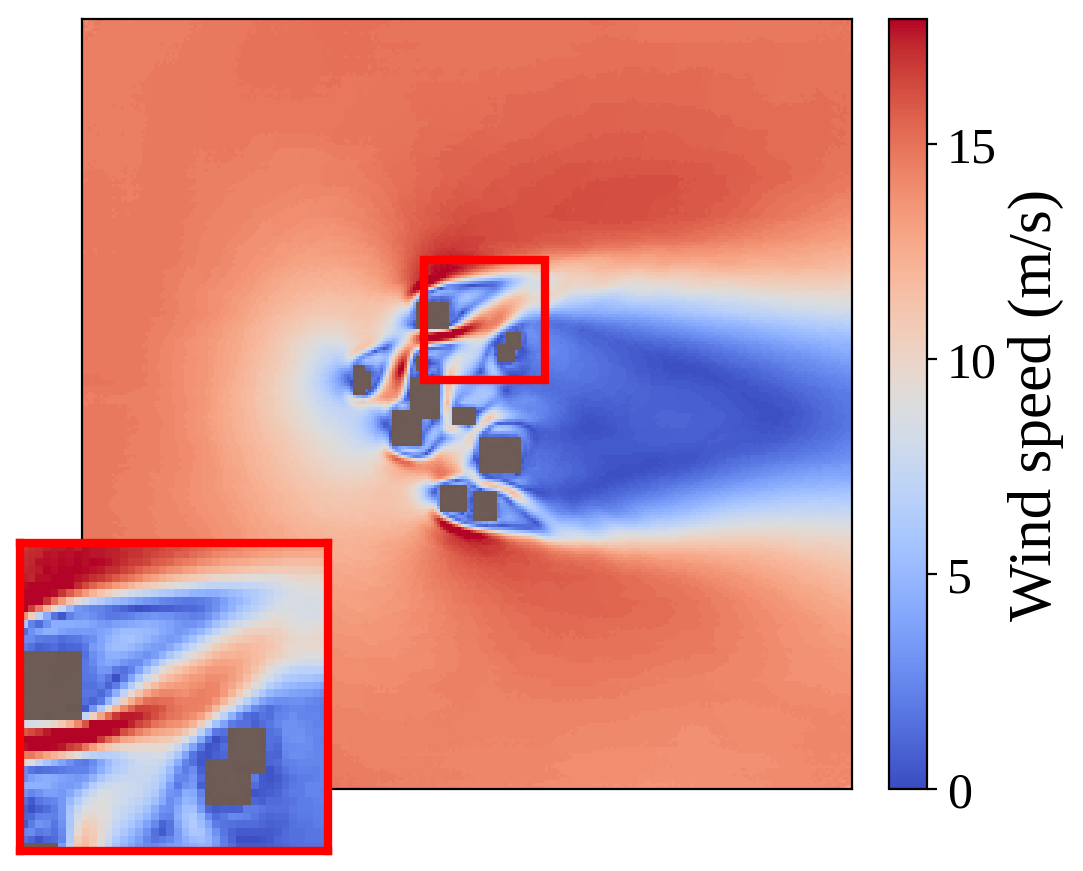}
    \captionsetup{margin={0pt,10pt}}
    \caption{}
  \end{subfigure}\hfill
  \begin{subfigure}[b]{\snapshotw}
    \centering
    \includegraphics[width=\textwidth]{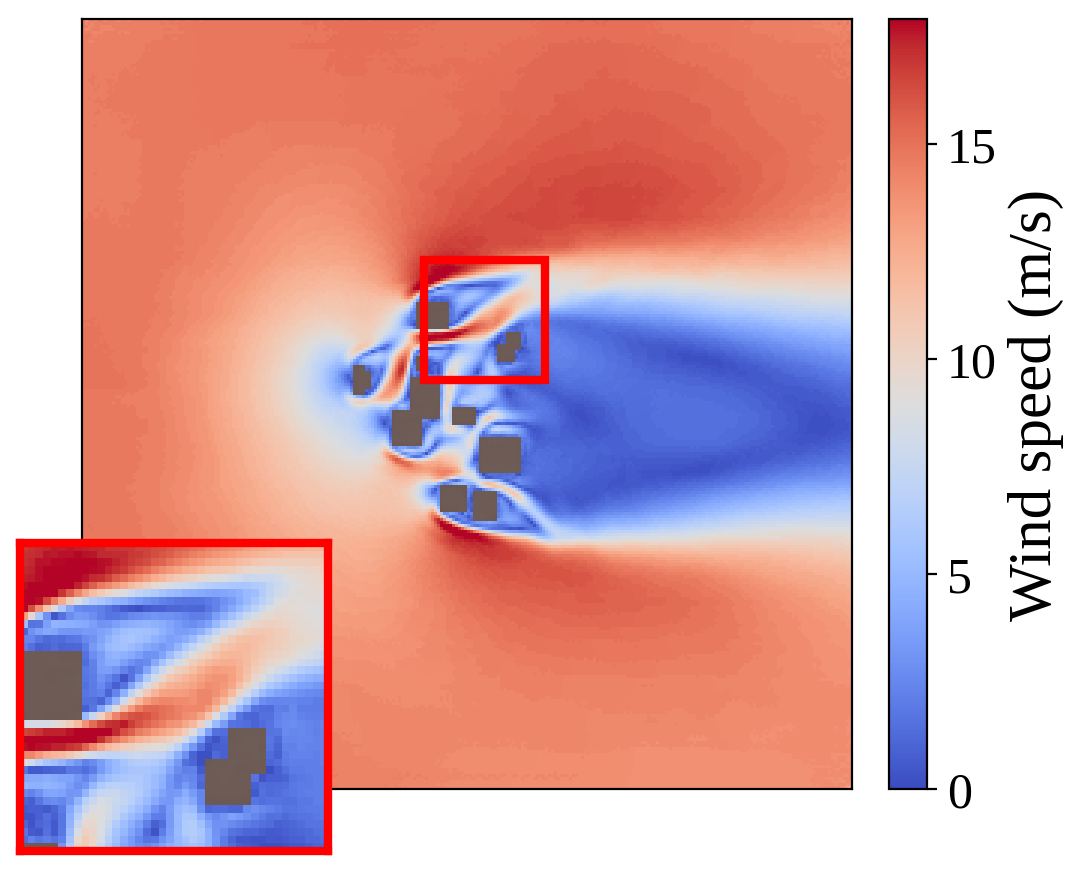}
    \captionsetup{margin={0pt,10pt}}
    \caption{}
  \end{subfigure}\hfill
  \begin{subfigure}[b]{\snapshotw}
    \centering
    \includegraphics[width=\textwidth]{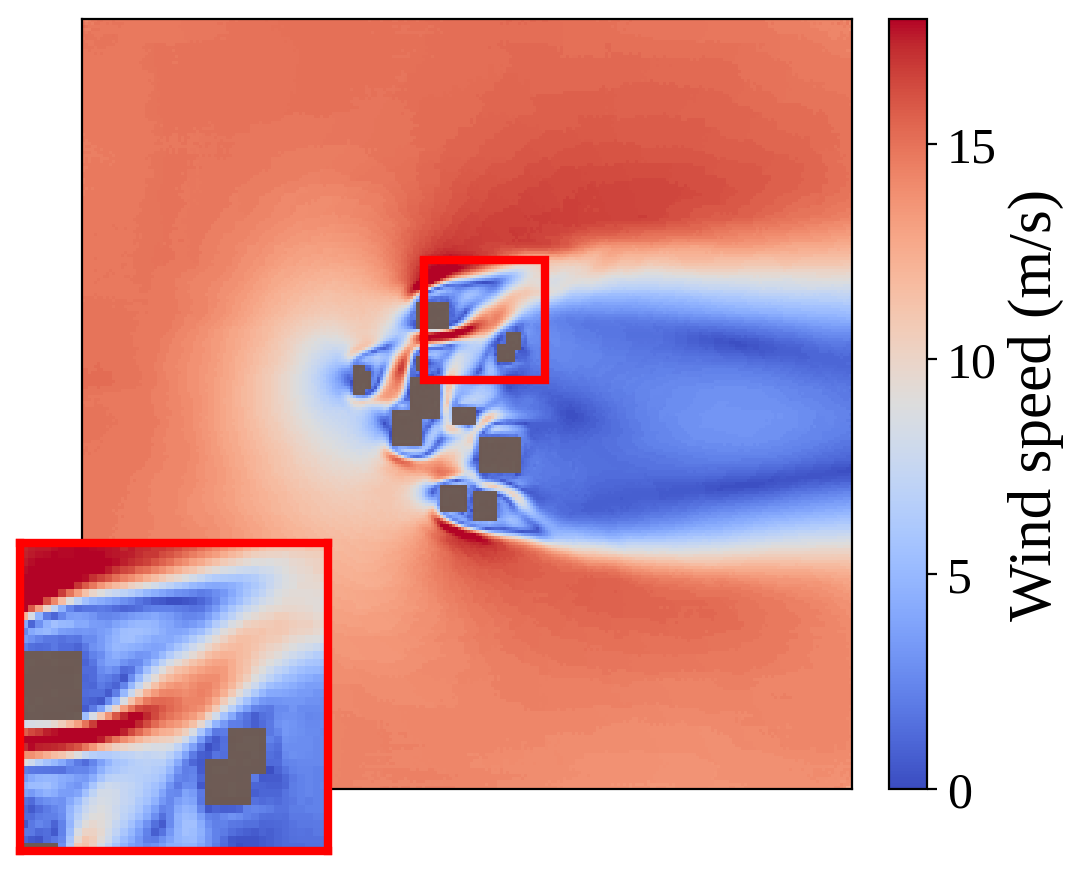}
    \captionsetup{margin={0pt,10pt}}
    \caption{}
  \end{subfigure}\hfill
  \begin{subfigure}[b]{\snapshotw}
    \centering
    \includegraphics[width=\textwidth]{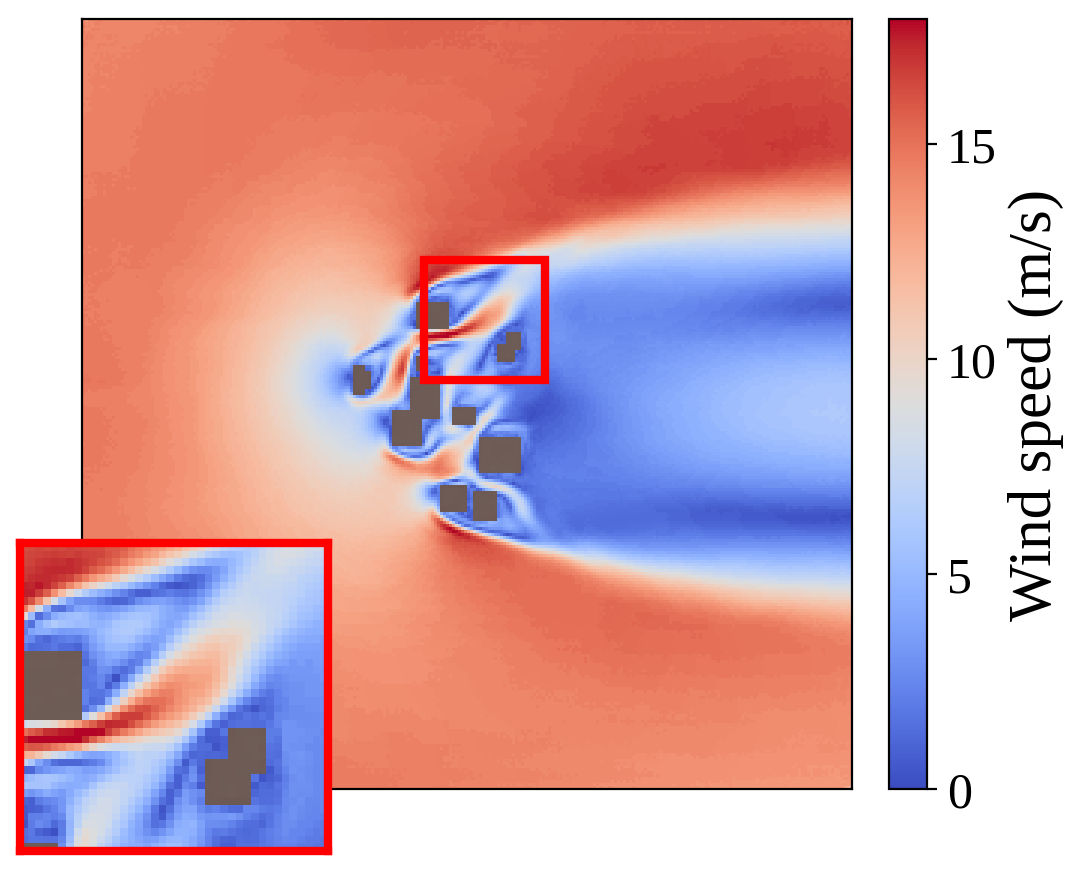}
    \captionsetup{margin={0pt,10pt}}
    \caption{}
  \end{subfigure}
  \\[0.3em]
  \raisebox{0.1\textwidth}{\rotatebox{90}{\small\textit{GT}}}%
  \hfill
  \begin{subfigure}[b]{\snapshotw}
    \centering
    \includegraphics[width=\textwidth]{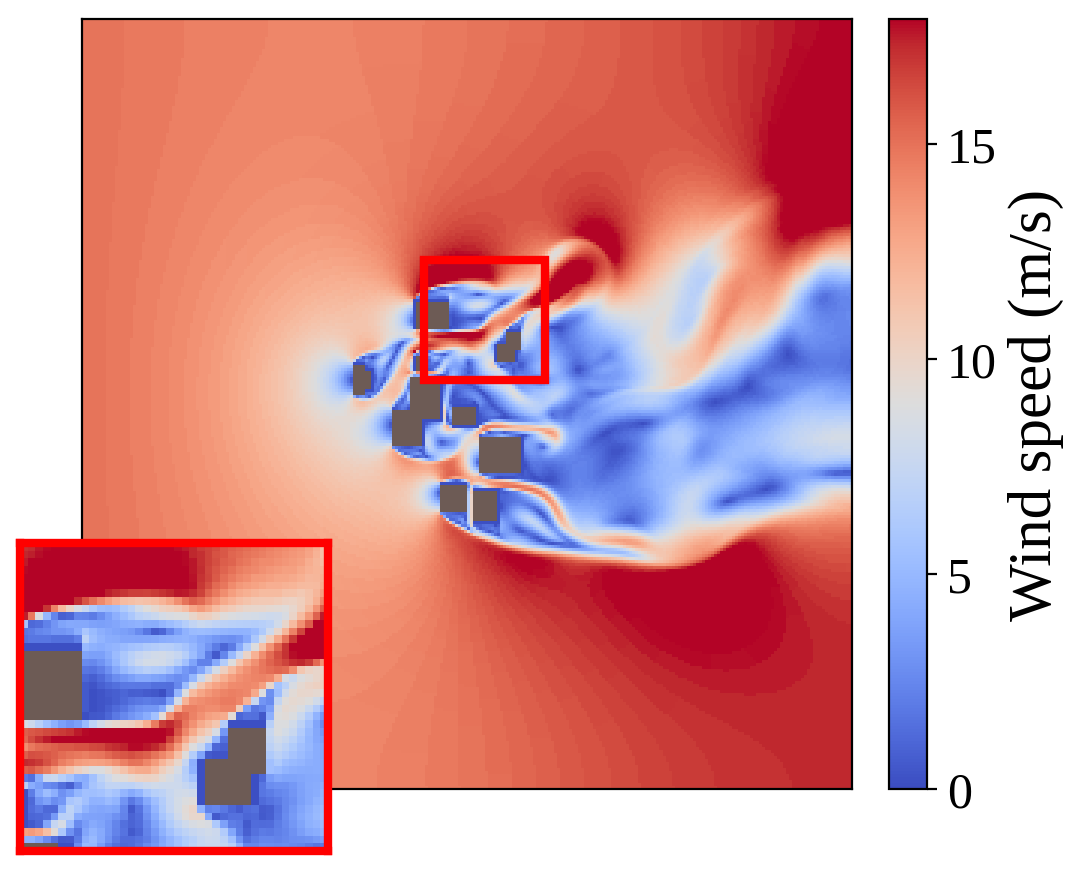}
    \captionsetup{margin={0pt,10pt}}
    \caption{}
  \end{subfigure}\hfill
  \begin{subfigure}[b]{\snapshotw}
    \centering
    \includegraphics[width=\textwidth]{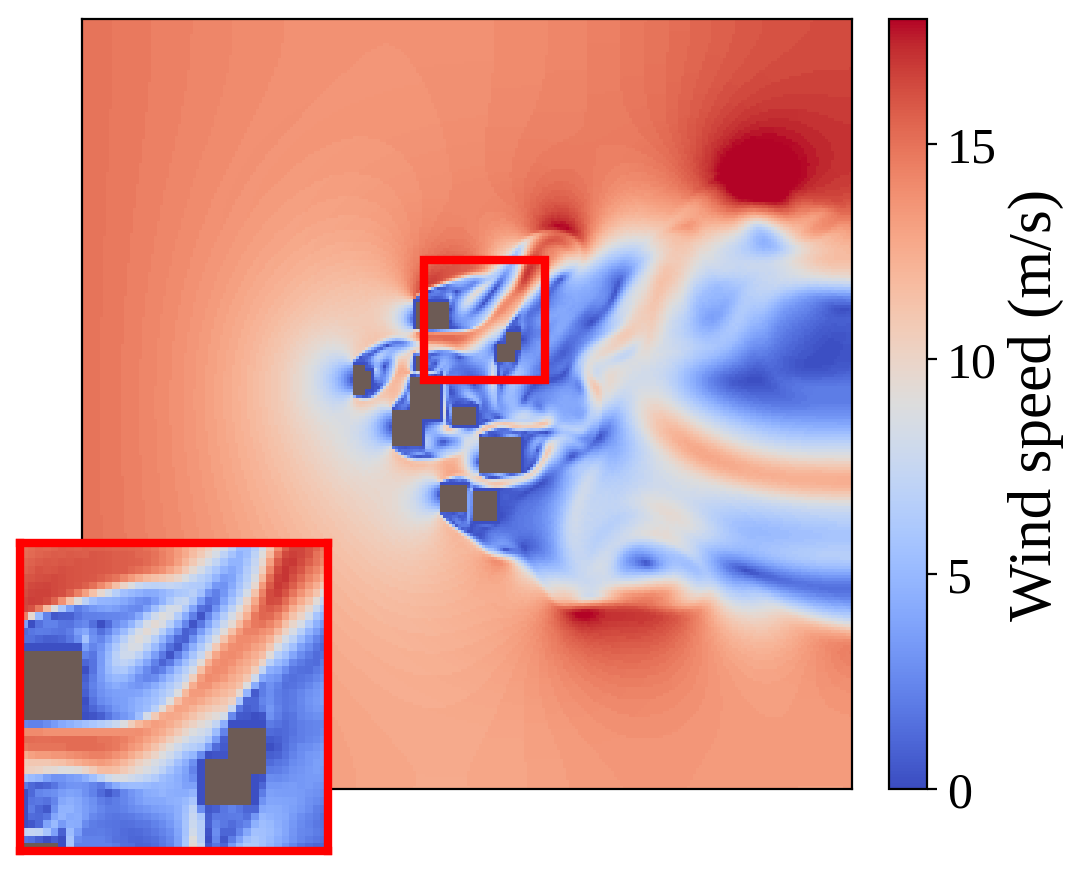}
    \captionsetup{margin={0pt,10pt}}
    \caption{}
  \end{subfigure}\hfill
  \begin{subfigure}[b]{\snapshotw}
    \centering
    \includegraphics[width=\textwidth]{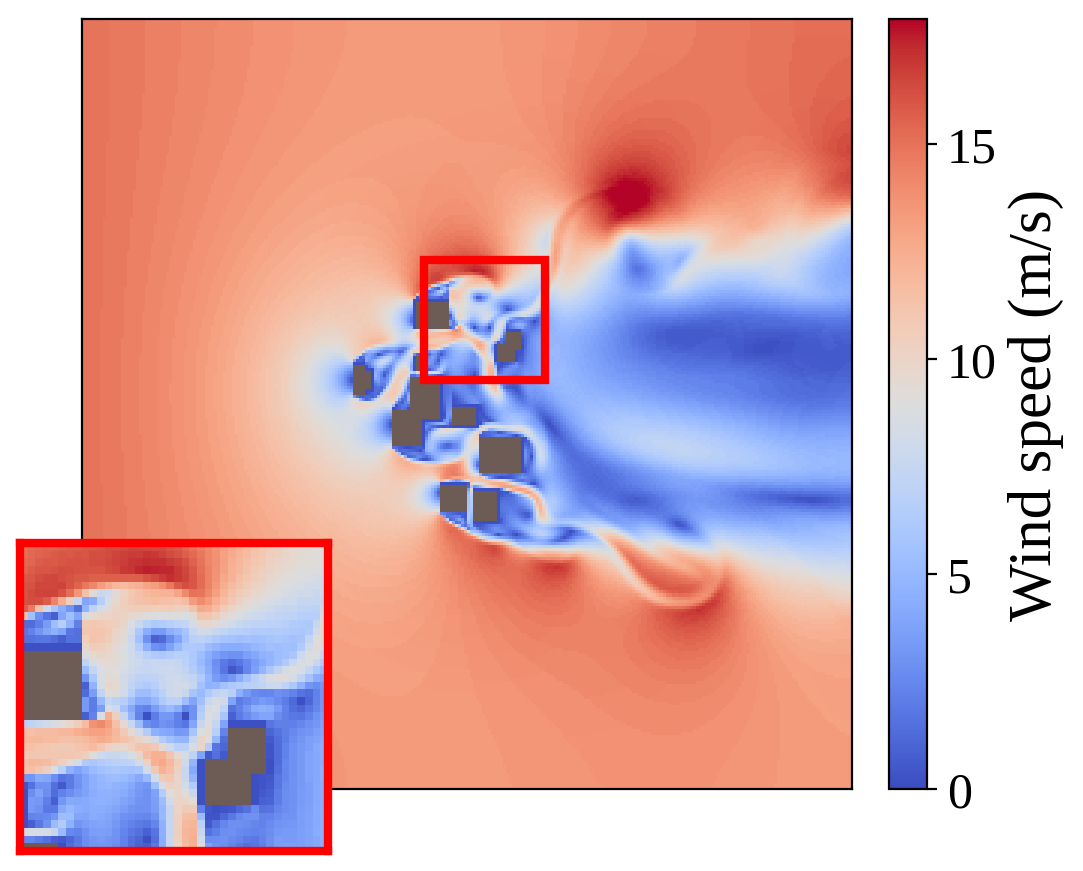}
    \captionsetup{margin={0pt,10pt}}
    \caption{}
  \end{subfigure}\hfill
  \begin{subfigure}[b]{\snapshotw}
    \centering
    \includegraphics[width=\textwidth]{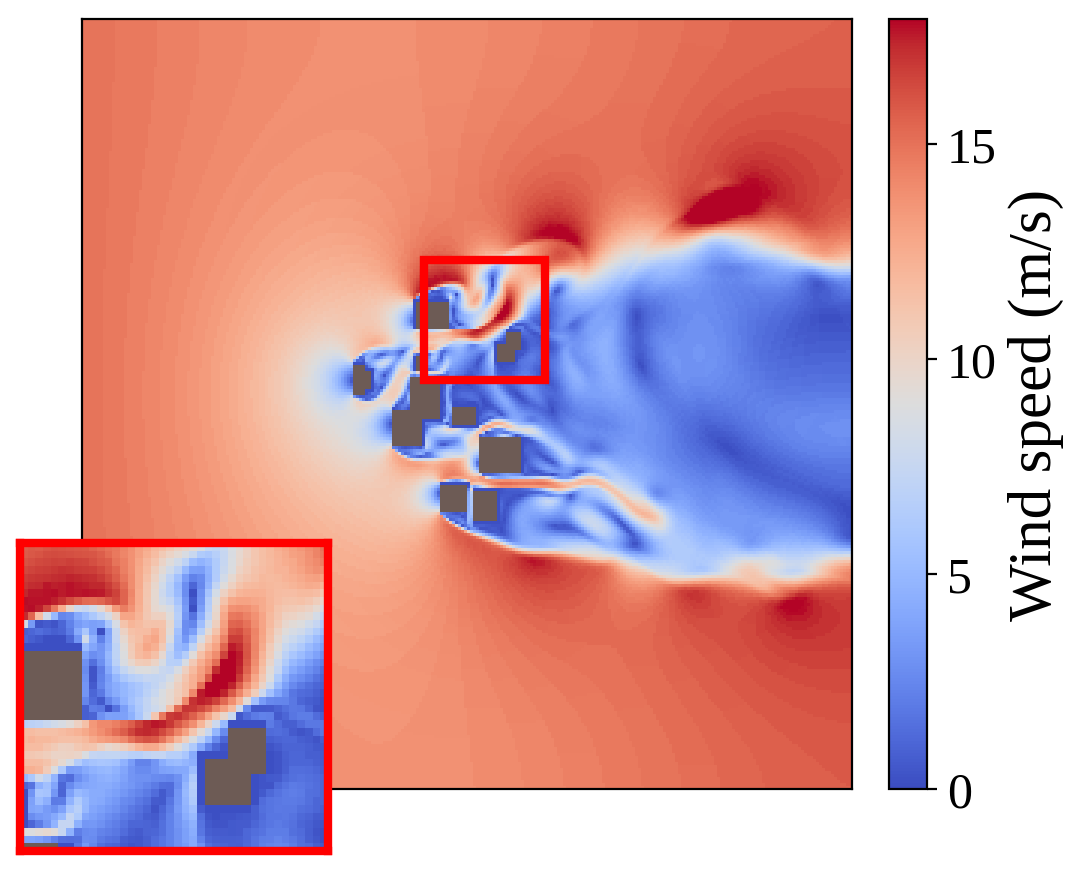}
    \captionsetup{margin={0pt,10pt}}
    \caption{}
  \end{subfigure}
  \caption{%
    Temporal extrapolation beyond the $T{=}112$-frame training horizon. Rows show predictions at rollout lengths $T$, $2T$, and $4T$. The bottom row is ground-truth CFD. All predictions use a single forward pass with no autoregressive chaining.
  }
  \label{fig:time_ablation_snapshots}
\end{figure}

\begin{figure}[H]
  \centering
  \includegraphics[width=0.6\linewidth]{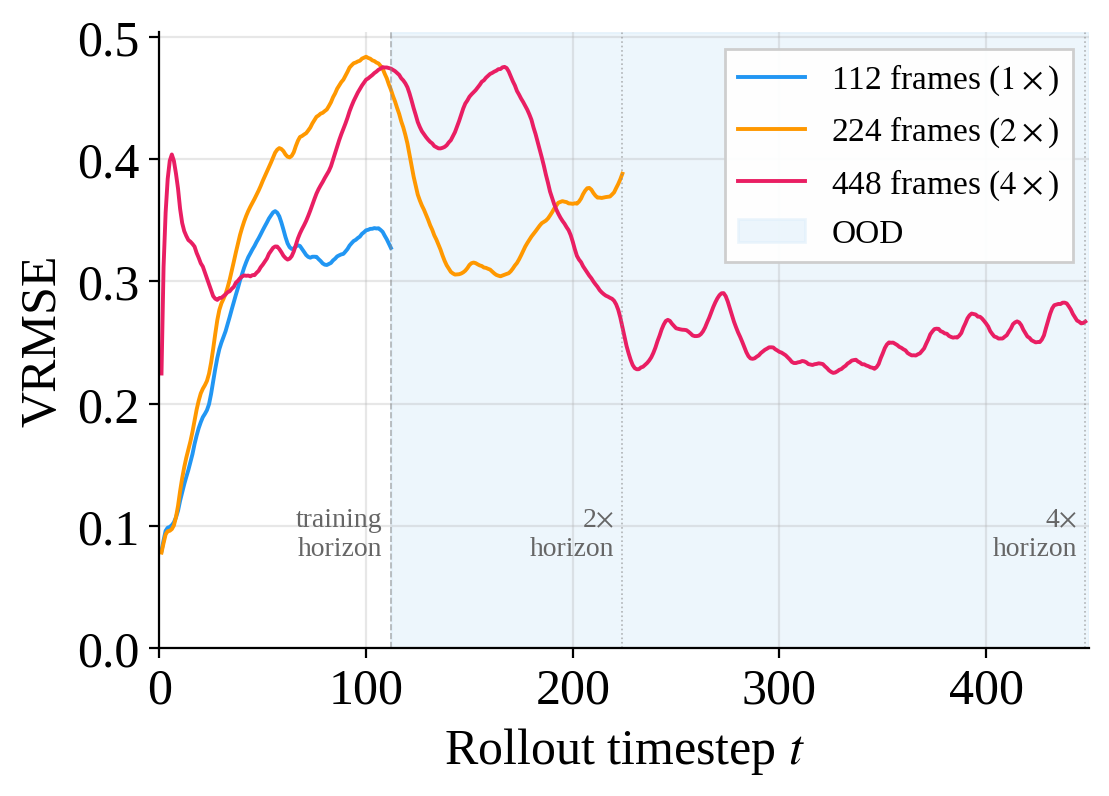}
  \caption{%
    Per-timestep VRMSE for rollout lengths $1\times$ ($T{=}112$), $2\times$ ($2T{=}224$), and $4\times$ ($4T{=}448$) the training horizon. Shaded regions indicate frames beyond each model's generation length.
  }
  \label{fig:time_ablation_vrmse}
\end{figure}

The model is trained on $T{=}112$-frame sequences. LTX-Video uses rotary position embeddings (RoPE) for temporal attention, which in principle allows variable-length generation at inference without architectural changes. We test this by generating rollouts of $T{=}112$, $2T{=}224$, and $4T{=}448$ frames from the same initial condition (Figs.~\ref{fig:time_ablation_snapshots} and~\ref{fig:time_ablation_vrmse}).

The $2T$ rollout still looks plausible, but longer sequences progressively lose fine-grained detail and produce increasingly averaged predictions. The per-timestep VRMSE for the $4T$ rollout rises initially but plateaus around ${\sim}0.25$ beyond $t \approx 250$. This coincides with the transition to quasi-steady-state flow, where the wake structure becomes locally periodic. In this regime an averaged prediction is close to the time-mean flow, which explains why the error stops growing rather than diverging.

\subsection{Inlet Speed Generalization}
\label{app:inlet_speed}

\begin{figure}[H]
  \centering
  \includegraphics[width=0.6\linewidth]{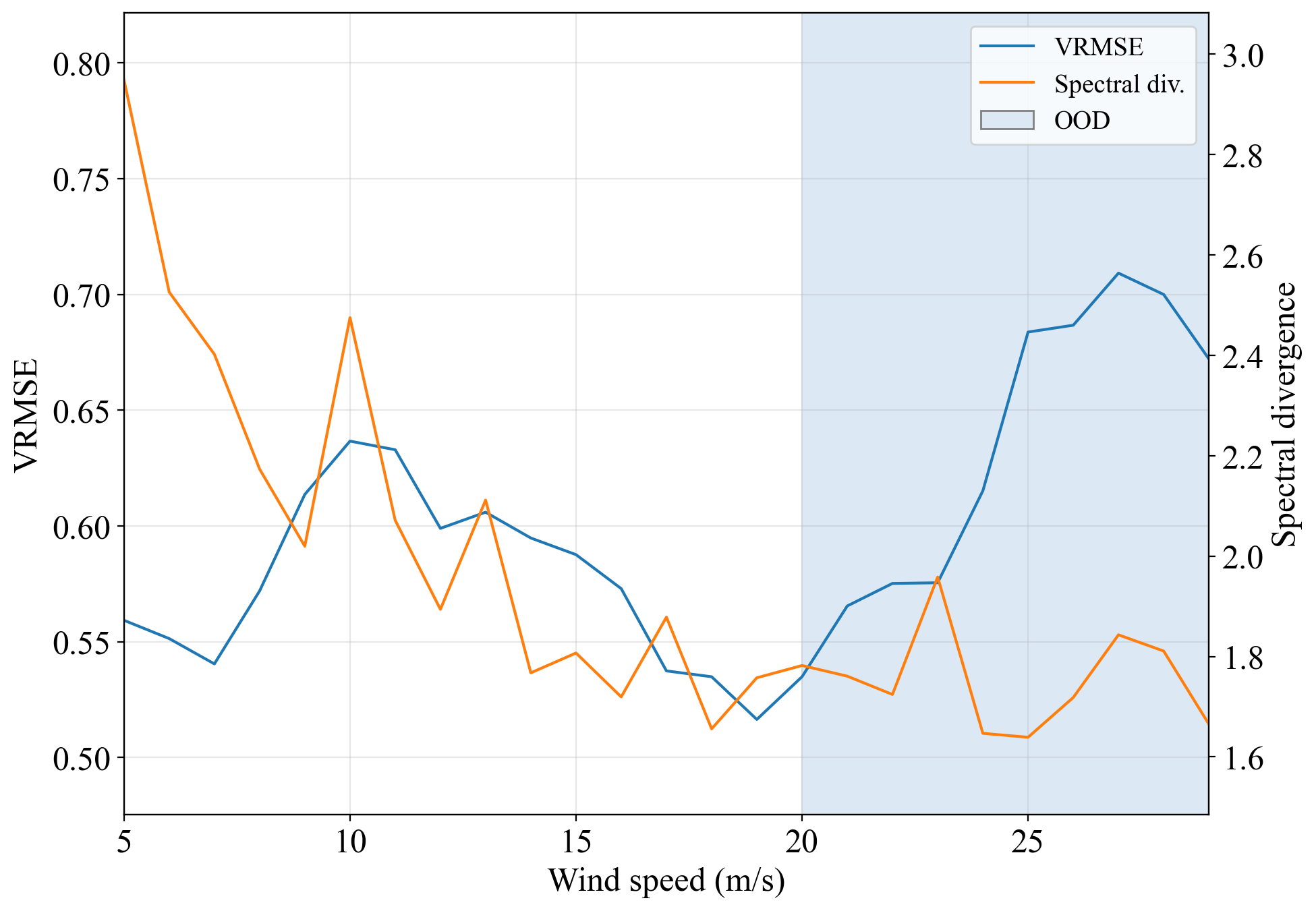}
  \caption{%
    VRMSE and spectral divergence as a function of $u_\mathrm{in}$ ($L = 1100$\,m).
    The shaded blue region indicates out-of-distribution conditioning values beyond the training range.
  }
  \label{fig:conditioning_speed}
\end{figure}

The training set covers $u_\mathrm{in} \in [0.1, 20]$\,m/s. We evaluate our best performing model across $u_\mathrm{in}$ from 5 to 29\,m/s at 1\,m/s resolution, keeping $L$ fixed at 1100\,m (mid-training range). All experiments use the same building configuration to isolate the effect of wind speed. Figure~\ref{fig:conditioning_speed} shows VRMSE and spectral divergence as a function of $u_\mathrm{in}$. Within the training range, VRMSE remains between 0.52 and 0.64, reaching its minimum of 0.52 at 19\,m/s. Beyond the training boundary at 20\,m/s, VRMSE increases gradually from 0.57 at 21\,m/s to 0.71 at 27\,m/s, indicating graceful degradation rather than catastrophic failure. Spectral divergence remains relatively stable across the entire range, suggesting that the model preserves the spatial frequency structure of the flow even when extrapolating in speed.

\begin{figure}[H]
  \centering
  \includegraphics[width=\linewidth]{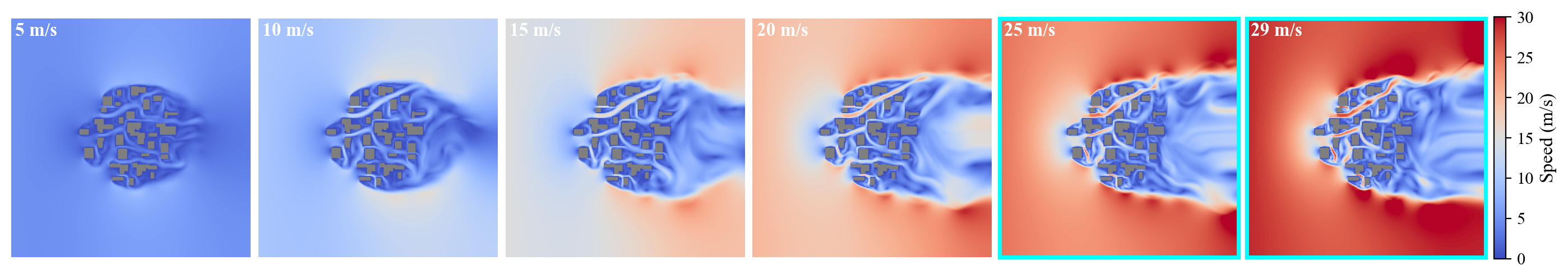}\\[4pt]
  \includegraphics[width=\linewidth]{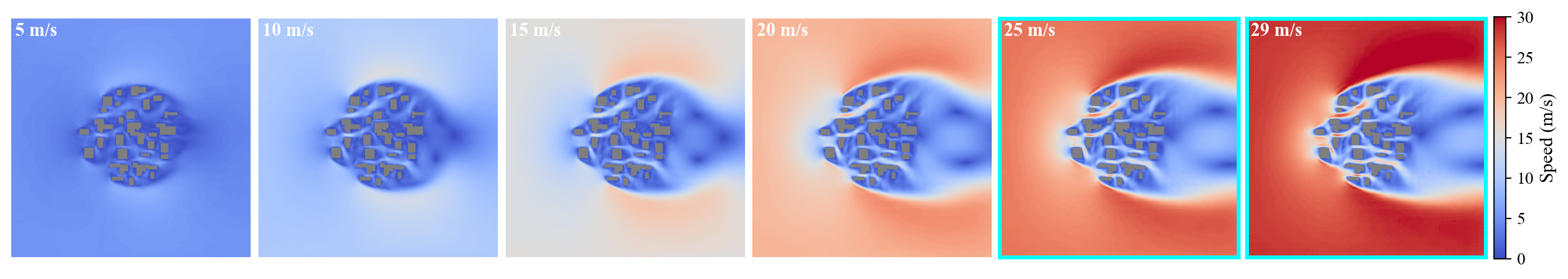}
  \caption{%
    Wind speed magnitude at $t = 24$ for inlet speeds from 5 to 29\,m/s (left to right).
    Top row: ground truth; bottom row: model prediction. Out-of-distribution inlet speeds ($u_\mathrm{in} > 20$\,m/s) are highlighted with a cyan border.
  }
  \label{fig:speed_sensitivity_frames}
\end{figure}

\subsection{Domain Size Generalization}
\label{app:domain_size}

\begin{figure}[H]
  \centering
  \includegraphics[width=0.6\linewidth]{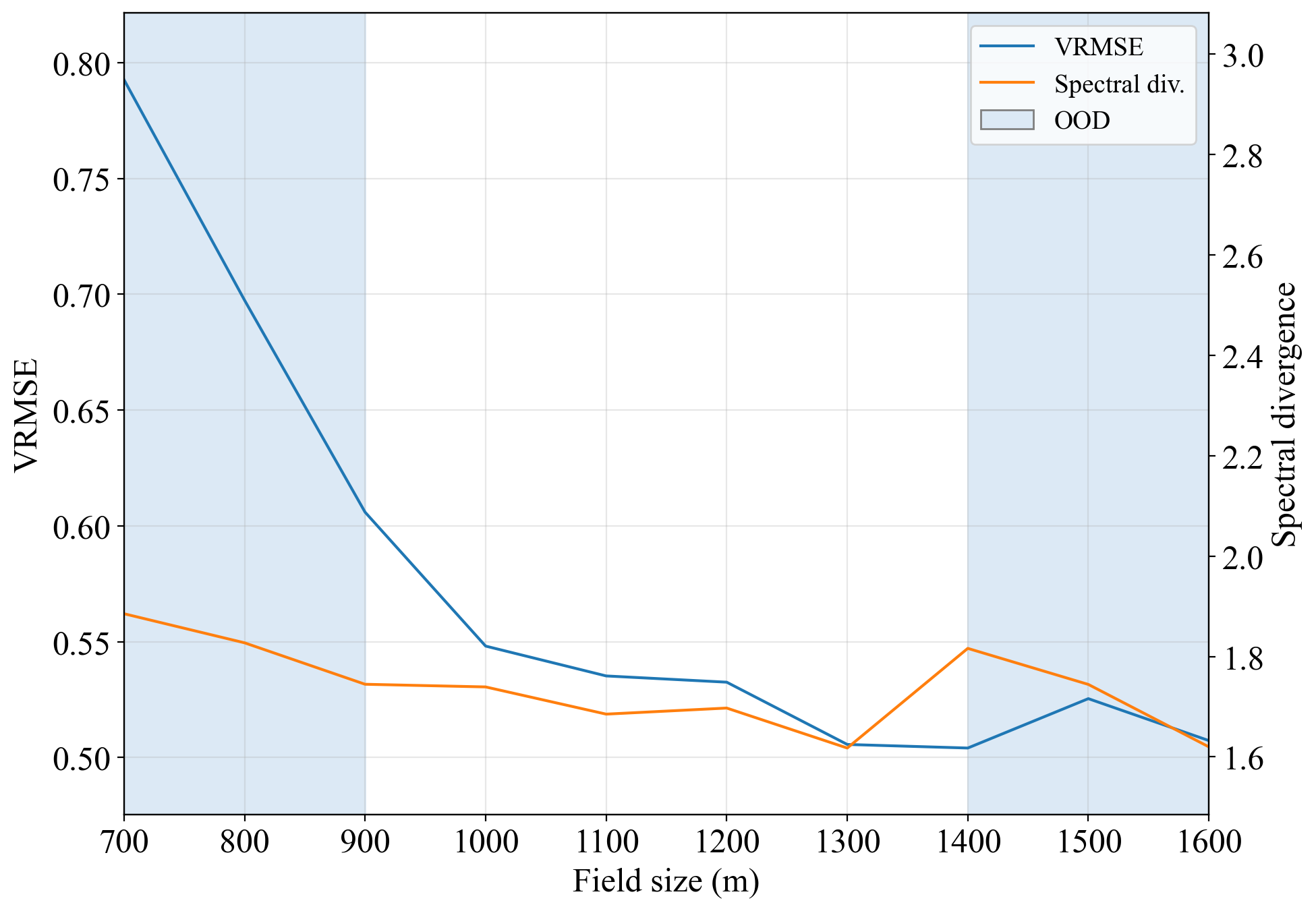}
  \caption{%
    VRMSE and spectral divergence as a function of $L$ ($u_\mathrm{in} = 18$\,m/s).
    Shaded blue regions indicate out-of-distribution conditioning values beyond the training range.
  }
  \label{fig:conditioning_field_size}
\end{figure}

The training set covers $L \in [900, 1400]$\,m. We vary $L$ from 700 to 1600\,m in 100\,m increments, with $u_\mathrm{in}$ fixed at 18\,m/s (the speed with lowest spectral divergence within the training range). The same building configuration is used throughout; as $L$ changes, buildings scale proportionally with the domain, preserving the layout geometry while varying the effective resolution relative to the $256{\times}256$ grid. Figure~\ref{fig:conditioning_field_size} shows the results. Within the training range, VRMSE decreases from 0.61 at $L{=}900$\,m to 0.50 at $L{=}1400$\,m, with spectral divergence stable around 1.7. The model generalizes well to larger domains: at 1500 and 1600\,m, VRMSE remains at 0.53 and 0.51, comparable to in-distribution performance. This is expected, as larger domains correspond to sparser building layouts that are easier to resolve. Generalization to smaller domains is less robust: at $L{=}800$\,m VRMSE rises to 0.70, and at $L{=}700$\,m it reaches 0.79. Smaller values of $L$ pack the same urban geometry into a tighter domain, producing denser configurations that the model has not seen during training. The asymmetric degradation reflects the inherent difficulty of resolving densely packed layouts where flow interactions between buildings are stronger.

\subsection{Channel Assignment Ablation}
\label{sec:appendix_channel_ablation}

Table~\ref{tab:channel_permutation} evaluates all six permutations of the channel-to-RGB assignment using the default (frozen) LTX-Video VAE. The four permutations that keep the building footprint off the green channel (ranks 1--4) all achieve similar performance, with VRMSE between 0.237 and 0.252 and MAE between 0.41 and 0.43\,m/s. Assigning the building footprint to the green channel (ranks 5--6) degrades VRMSE by roughly 35\%. In all experiments we use the natural ordering $R{=}u$, $G{=}v$, $B{=}b$.

\begin{table}[ht]
  \centering
  \caption{Default LTX-Video VAE reconstruction quality across all channel permutations, evaluated on 200 test simulations.}
  \label{tab:channel_permutation}
  \vspace{0.5ex}
  \small
  \begin{tabular*}{\linewidth}{@{\extracolsep{\fill}}clccccc}
    \toprule
    Rank & Mapping & VRMSE\,$\downarrow$ &  MAE\,$\downarrow$ & MRE\,$\downarrow$ & Spectral\,$\downarrow$ & $W_1$\,$\downarrow$ \\
    \midrule
    1 & $R{=}v,\; G{=}u,\; B{=}b$ & 0.237 &  0.425 & 2.49\% & 0.794 & 0.406 \\
    2 & $R{=}b,\; G{=}u,\; B{=}v$ & 0.242 &  0.405 & 2.44\% & 0.808 & 0.397 \\
    3 & $R{=}u,\; G{=}v,\; B{=}b$ & 0.246 & 0.418 & 2.67\% & 0.792 & 0.425 \\
    4 & $R{=}b,\; G{=}v,\; B{=}u$ & 0.252 & 0.421 & 2.26\% & 0.805 & 0.422 \\
    5 & $R{=}v,\; G{=}b,\; B{=}u$ & 0.323 & 0.516 & 3.17\% & 0.847 & 0.516 \\
    6 & $R{=}u,\; G{=}b,\; B{=}v$ & 0.333 & 0.499 & 3.16\% & 0.845 & 0.518 \\
    \bottomrule
  \end{tabular*}
\end{table}

\end{document}